\RequirePackage{fix-cm}

\documentclass[12pt,letterpaper]{article}
\usepackage[top=0.6in,left=1.5in,footskip=0.75in,marginparwidth=2in]{geometry}
\date{}

\usepackage[pdftex]{graphicx}
\usepackage{subfigure}

\usepackage{algorithmic}
\usepackage{algorithm}
\usepackage{amssymb,amsmath}
\usepackage{stmaryrd}
\usepackage{xcolor}
\usepackage{multicol}
\usepackage{multirow}
\usepackage[T1]{fontenc}
\usepackage{xintexpr}

\usepackage{rotating}
\usepackage{multirow}
\usepackage{longtable}
\usepackage{colortbl}
\usepackage{dsfont}

\usepackage[export]{adjustbox}%

\usepackage{tikz}
\usepackage{centernot}%
\usetikzlibrary{shapes.geometric}

\tikzstyle{extremely densely dashed} = [dash pattern=on 2pt off 1pt]
\tikzstyle{pattern}=[draw,circle,black,bottom color=white, top color= white, text=black,minimum width=10pt]
\tikzstyle{peers}=[draw,circle,violet,bottom color=green!60, top color= white, text=violet,minimum width=10pt, scale = 0.7]
\definecolor{burntorange}{cmyk}{0,0.52,1,0}
\def\oran{orange!30}
\tikzstyle{cliquepeers}=[draw, red, fill= red, text=black,minimum width=1pt, scale=0.4]
\tikzstyle{myFivePoly} =  [regular polygon,regular polygon sides=5,minimum width=1pt, scale=0.9]
\tikzstyle{mytriangle} =  [fill=blue!20, regular polygon, regular polygon sides=3,minimum width=1pt, scale=0.4]
\tikzstyle{extremely densely dashed} = [dash pattern=on 2pt off 1pt]
\tikzstyle{finedashpath}=[violet, extremely densely dashed,thick]
\tikzstyle{superpeers}=[draw,circle,thick,burntorange, left color=\oran,text=black, minimum width=20pt]
\tikzstyle{pattern}=[draw,circle,black,bottom color=white, top color= white, text=black,minimum width=10pt]
\newcommand{\ue}{%
	\begin{tikzpicture}%
	\draw[fill = black] (.25ex,.25ex) circle (.3ex);
	\draw[thick] (.55ex,.25ex) -- (1.55ex,.25ex);%
	\draw[fill = black] (1.85ex, .25ex) circle (.3ex);%
	\end{tikzpicture}%
}

\newcommand{\sampleObjectGraphA}{%
	\begin{tikzpicture}
	\node[scale=0.5] (box_super1) at ([xshift=-6mm,yshift=-4mm] 0,0) {\includegraphics[scale=0.1]{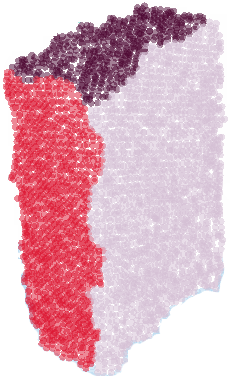}};
	\foreach \place/\name/\counter/\colorvertex/\shapevertex in {{([xshift=0,yshift=7] box_super1)/grapha/1/green/mytriangle}, {([xshift=2,yshift=-1] box_super1)/graphb/2/cyan/myFivePoly}, {([xshift=-3,yshift=-3] box_super1)/graphc/3/blue/rectangle}}
	\node[cliquepeers,\shapevertex, \colorvertex,scale = 0.7] (\name) at\place {};%
	\foreach \source/\dest in {grapha/graphc, graphb/graphc}
	\path (\source) edge (\dest);
	\end{tikzpicture}%
}

\newcommand{\sampleWordCliqueCompleteA}{%
	\begin{tikzpicture}
	\node[superpeers, minimum width=20pt] (sampleCliqueVertex) at (0,0) {};
	
	\foreach \shapeI/\colorI/\shapeII/\colorII\shapeIII/\colorIII in {{rectangle/blue/myFivePoly/cyan/mytriangle/green}}{
		\node[] (subgraph_n1) at ([xshift=-1mm,yshift=-1mm] sampleCliqueVertex){};
		\node[] (subgraph_n2) at ([xshift=1mm,yshift=1mm] sampleCliqueVertex){};
		\node[] (subgraph_n3) at ([xshift=-1mm,yshift=1mm] sampleCliqueVertex){};
		
		\path[thin,gray, shorten <=-2.0pt,shorten >=-2.0pt] (subgraph_n1)  edge (subgraph_n2);
		\path[thin,gray, shorten <=-2.0pt,shorten >=-2.0pt] (subgraph_n1) edge (subgraph_n3);
		
		\node[cliquepeers,\shapeI,\colorI,scale=1.3] (subgraph_n1_img) at (subgraph_n1){};
		\node[cliquepeers,\shapeII,\colorII,scale=1.3] (subgraph_n2_img) at (subgraph_n2){};
		\node[cliquepeers,\shapeIII,\colorIII,scale=1.3] (subgraph_n3_img) at (subgraph_n3){};
	}
	\end{tikzpicture}%
}

\newcommand{\sampleWordA}{%
	\begin{tikzpicture}%
	\node[cliquepeers,myFivePoly,cyan,scale=1.5] (samplWordA) at (0,0) {};
	\end{tikzpicture}%
}
\newcommand{\sampleWordB}{%
	\begin{tikzpicture}%
	\node[cliquepeers,rectangle,blue,scale=1.75] (samplWordB) at (0,0) {};
	\end{tikzpicture}%
}
\newcommand{\sampleWordC}{%
	\begin{tikzpicture}%
	\node[cliquepeers,mytriangle,green,scale=1.5] (samplWordC) at (0,0) {};
	\end{tikzpicture}%
}

\newcommand{\sampleWordD}{%
	\begin{tikzpicture}%
	\node[cliquepeers,circle,green,scale=1.5] (samplWordD) at (0,0) {};
	\end{tikzpicture}%
}

\newcommand{\sampleWordE}{%
	\begin{tikzpicture}%
	\node[cliquepeers,myFivePoly,green,scale=1.5] (samplWordE) at (0,0) {};
	\end{tikzpicture}%
}

\newcommand{\sampleWordCliqueA}{%
	\begin{tikzpicture}%
	\node[] (a) at (0.25,0) {};
	\node[] (a_subgraph_n1) at ([xshift=-1mm,yshift=-1mm] a){};
	\node[] (a_subgraph_n2) at ([xshift=1mm,yshift=0.05mm] a){};
	\path[thin,gray, shorten <=-1.75pt,shorten >=-1.75pt] (a_subgraph_n1) edge (a_subgraph_n2);
	\node[cliquepeers,rectangle,blue] (a_subgraph_n1_img) at (a_subgraph_n1){};
	\node[cliquepeers,myFivePoly,cyan] (a_subgraph_n2_img) at (a_subgraph_n2){};
	\end{tikzpicture}%
}

\newcommand{\samplePattern}{%
	\begin{tikzpicture}%
	\node[pattern, inner sep=0pt, bottom color = gray, scale=0.5] (samplePattern) at (0,0) {};
	\end{tikzpicture}%
}

\newcommand{\sampleWordCliqueB}{%
	\begin{tikzpicture}%
	\node[] (b) at (0,0) {};
	\node[] (b_subgraph_n1) at ([xshift=-1mm,yshift=-1mm] b){};
	\node[] (b_subgraph_n2) at ([xshift=-3mm,yshift=-1mm] b){};
	\path[thin,gray, shorten <=-1.75pt,shorten >=-1.75pt] (b_subgraph_n1) edge (b_subgraph_n2);
	\node[cliquepeers,rectangle,blue] (b_subgraph_n1_img) at (b_subgraph_n1){};
	\node[cliquepeers,mytriangle,green] (b_subgraph_n2_img) at (b_subgraph_n2){};
	\end{tikzpicture}%
}

\newcommand{\sampleWordCliqueC}{%
	\begin{tikzpicture}%
	\node[] (d) at (0,0) {};
	\node[] (d_subgraph_n1) at ([xshift=-1mm,yshift=-1mm] d){};
	\node[] (d_subgraph_n2) at ([xshift=1mm,yshift=0.05mm] d){};
	\node[] (d_subgraph_n3) at ([xshift=-3mm,yshift=-1mm] d){};    
	\path[thin,gray, shorten <=-1.75pt,shorten >=-1.75pt] (d_subgraph_n1)  edge (d_subgraph_n2);
	\path[thin,gray, shorten <=-1.75pt,shorten >=-1.75pt] (d_subgraph_n1) edge (d_subgraph_n3);
	\node[cliquepeers,rectangle,blue] (d_subgraph_n1_img) at (d_subgraph_n1){};
	\node[cliquepeers,myFivePoly,cyan] (d_subgraph_n2_img) at (d_subgraph_n2){};
	\node[cliquepeers,mytriangle,green] (d_subgraph_n3_img) at (d_subgraph_n3){};;
	\end{tikzpicture}%
}

\newcommand{\inlineimage}[1]{$\vcenter{\hbox{\protect\includegraphics[height=0.8\baselineskip,origin=c]{#1}}}$}
\newcommand{\inlineimagetable}[1]{$\vcenter{\hbox{\protect\includegraphics[height=0.8\normalbaselineskip,origin=c]{#1}}}$}

\newcommand{\usubref}[1]{\protect\subref{#1}}

\usepackage{url}
\DeclareMathOperator*{\argmax}{arg\,max}

\begin{document}
\title{\bf Visual Object Categorization \\ Based on Hierarchical Shape Motifs \\Learned From Noisy Point Cloud Decompositions}

\author{Christian A. Mueller \and Andreas Birk
\thanks{The authors are with the Robotics Group of the Computer Science \& Electrical Engineering Department, Jacobs University Bremen, Germany, \tt \{chr.mueller,a.birk\}@jacobs-university.de} 
}

\maketitle

\begin{abstract}
Object shape is a key cue that contributes to the semantic understanding of objects.
In this work we focus on the categorization of real-world object point clouds to particular shape types.
Therein surface description and representation of object shape structure have significant influence on shape categorization accuracy, when dealing with real-world scenes featuring noisy, partial and occluded object observations.
An unsupervised hierarchical learning procedure is utilized here to symbolically describe surface characteristics on multiple semantic levels. 
Furthermore, a constellation model is proposed that hierarchically decomposes objects.
The decompositions are described as constellations of symbols (shape motifs) in a gradual order, hence reflecting shape structure from local to global, i.e., from parts over groups of parts to entire objects. 

The combination of this multi-level description of surfaces and the hierarchical decomposition of shapes leads to a representation which allows to conceptualize shapes.
An object discrimination has been observed in experiments with seven categories featuring instances with sensor noise, occlusions as well as inter-category and intra-category similarities.
Experiments include the evaluation of the proposed description and shape decomposition approach, and comparisons to Fast Point Feature Histograms, a Vocabulary Tree and a neural network-based Deep Learning method.
Furthermore, experiments are conducted with alternative datasets which analyze the generalization capability of the proposed approach.
\end{abstract}

\section{Introduction}\label{sec:intro}
Everyday and everywhere we are surrounded by objects. 
Objects augment our environment and serve as entities in space which provide information about state. 
They are essential to interpret the state of environment and situation by their presence, pose, composition, etc.
One may say objects discretize our environment and form building blocks that \emph{describe} our environment.

Visual perception of objects is essential for a plethora of tasks in service and
industrial robotics: robots are supposed to assemble parts on assembly lines, unload shipping containers, maintain stock shelves in super markets, guiding visitors in museums, reason about objects in big data applications like cloud robotics, etc.~\cite{JonschkowskiEHM16,7553531,winkler16shopping,CloudRoboticsSurvey-IEEETASE15}.
Consequently, robots are constantly confronted with object-related tasks like detection, recognition or categorization of objects as well as estimating or tracking their pose.
In particular, robots are increasingly applied in manipulation, monitoring or surveying tasks where they face a significant amount of unknown objects making object perception tasks more and more challenging.
Therefore a perception system is required that is flexible and scalable to classify a large set of individual, even unknown, object instances into a significantly smaller set of groups in which instances within a group share commonalties such as in form of shape appearance; such group can be denoted as a category which underlies a shape concept.

Besides object detection, where potential object candidates are localized in a scene~\cite{richtsfeld2014learning,MuellerBirkIcra2016}, the goal of instance recognition is to learn a particular model w.r.t. a very specific known object instance (e.g. Heinz Tomato 57 Varieties Ketchup, 397gramm, plastic bottle), whereas in categorization the goal is to learn a generic model of a \emph{bottle}.
Therefore the challenging task in categorization, in contrary to recognition, is to learn a generic model from instance appearances that belong to the same \emph{type} or category based on similarities or commonalities~\cite{Sloutsky2010}.
In other words, in our example case of a \emph{bottle}, the goal is to learn the essence of what makes objects appear as a \emph{bottle}.
Consequently, information about individual instance specifics have generally a neglectable relevance in categorization.
One can say that \emph{recognition} is the generalization task of identifying \emph{known} objects from \emph{unknown} viewpoints whereas \emph{categorization} is the generalization task of classifying \emph{unknown} objects to \emph{known} categories which also entails the classification of \emph{unknown} objects from \emph{unknown} viewpoints~\cite{Palmeri2004a}.
Note that categorization of objects bears also the risk of uncertainty by the absence of an explicit object model in contrary to instance recognition where an explicit object model which is supposed to be recognized is given beforehand.
Therefore the major goal in the categorization task is robustness towards inter-category and intra-category variability, i.e. the extraction of category-specific characteristics while considering the diversity in instance appearances within each category.

Given raw 3D point cloud data, reasoning about object semantics such as a shape category requires an abstraction process: the data is processed towards an abstract level such that clusters of points are detected as a single entity that can be labeled to be part of a category (e.g., \emph{box}).  %
One can identify three subproblems:

\textbf{1) Robust extraction of regions} that abstract semantically meaningful segments in the scene from real-world (noisy and raw) sensor data. %
So-called over-segmentation techniques~\cite{local_Comaniciu2002,Papon13CVPR} are often applied to segment a scene into uniformed-sized (fine-grained) segments. %
A subsequent goal is to segment distinctive and homogeneous regions (e.g., Super Patch Segments~\cite{MuellerBirkIcra2016}), which ideally represent components or parts of objects and that can be used as building blocks for object reasoning purposes. 

\textbf{2) Detection and localization of object candidates} in scenes that reflect clusters or groups of regions.
The objective is to detect groups of neighboring regions, which comply with generic-object appearance characteristics or patterns~\cite{MuellerBirkIcra2016,Frintrop2010,richtsfeld2014learning}. 
Such appearances can be derived from theories of cognitive science, e.g. saliency analysis~\cite{Rahtu2010} or rule-based approaches such as the \emph{minima rule}~\cite{Hoffman1997}.
The latter infers objects based on the assumption that parts of objects do predominantly appear in a convex alignment.
By considering such geometric or texture features in the decision process, unknown primitive-shaped objects like boxes or cylinders can even be extracted in cluttered scenes~\cite{ContextSemanticLabelling-IJRR13,local_mueller2013a,richtsfeld2014learning}.
A robust detection of more complex-shaped objects requires the exploration of more sophisticated object appearance analyses (e.g. so-called \emph{objectness}~\cite{Alexe:2012:MOI:2377349.2377551} or analysis based on compositions of regions~\cite{MuellerBirkIcra2016}) in order to identify appearance patterns of shape part compositions and relationships among those which can represent complex-shaped objects.

\begin{figure*}[t]
	\centering
	\includegraphics[width=0.99\linewidth]{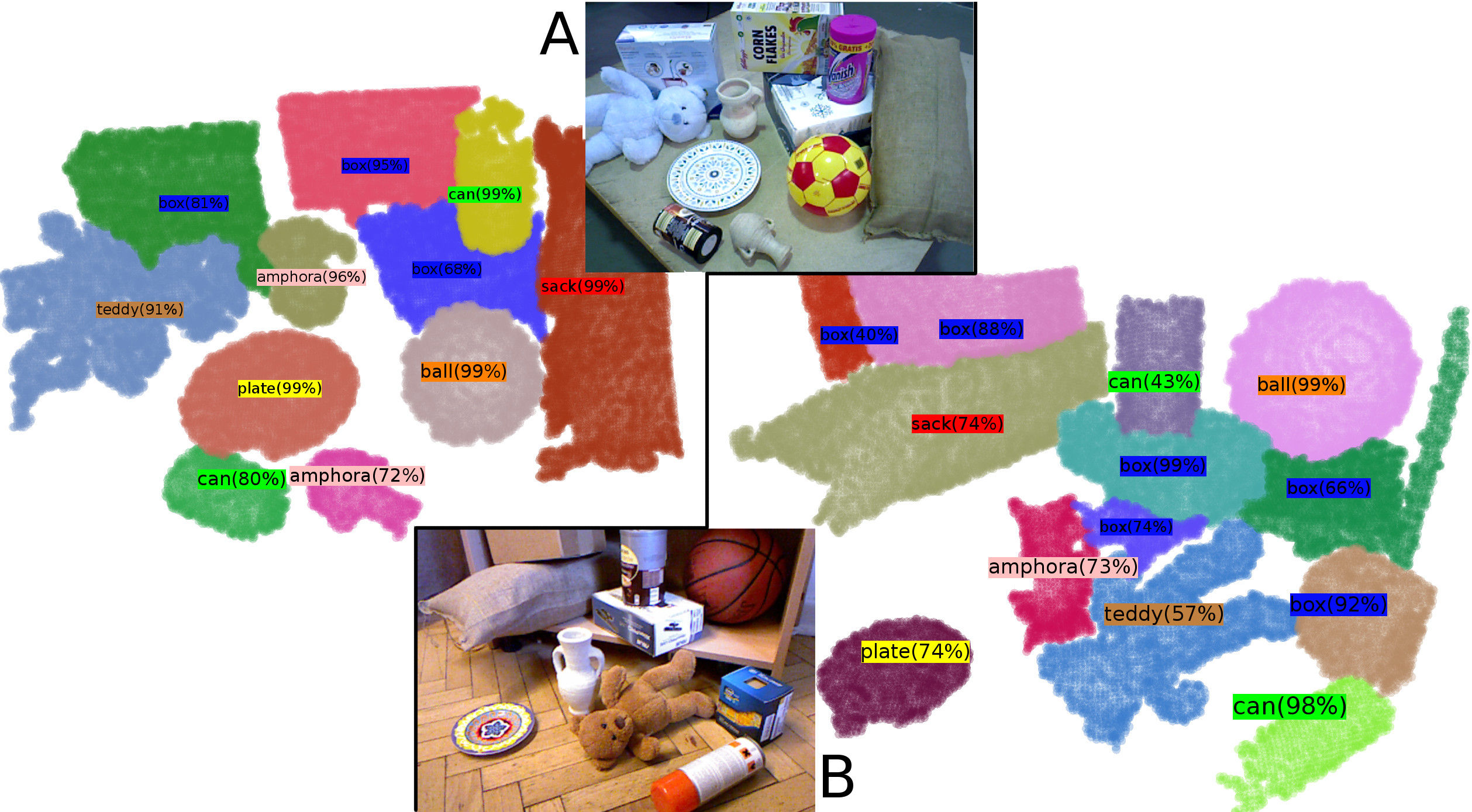}
	\caption{Example categorization of two unstructured scenes A and B. Object candidates (randomly colored) are segmented using our previous work~\cite{MuellerBirkIcra2016}. The shape category labels of the classification results according to the work presented here are colored as in Fig.~\ref{fig:vw_dict_distri}.
	}
	\label{fig:recognition_eval}
\end{figure*}

\textbf{3) Classification of object candidates} to specific object categories, which is the focus of this work (see Fig.~\ref{fig:recognition_eval}), with two main shape-related aspects as further motivated in Sec.~\ref{sec:rw}: i) the surface description and ii) the representation of shape structure.

\section{Motivation and Related Work}\label{sec:rw}
Robust reasoning about the shape of an object that is based on single-shot point cloud data has to deal with several challenges~\cite{JonschkowskiEHM16} like sensor noise, object occlusions, variations of object appearances from different viewpoints, quality of extracted object segmentations provided by an object candidate detector, etc.   
Under these challenging conditions, shape analysis with the goal of learning categories relies on a robust i) \emph{description} and ii) \emph{representation} of objects \cite{Biasotti:2008:DSG:1391729.1391731,dicarlo:tics_2007}.

Concerning i), an initial low-level abstraction of point cloud surfaces of objects can be accomplished by descriptors (e.g., Fast Point Feature Histograms (FPFH)~\cite{5152473}), which encode the point cloud characteristics in the form of \emph{description vectors}.
As shown by concepts like Bag-of-Words~\cite{DBLP:conf/iccv/JurieT05} an abstraction to a symbolic representation of according vectors is beneficial for perception tasks.
Therein, a crucial initial step is the quantization of the description space into a set of partitions, which are described as (visual) words of a dictionary or codebook. 
A word can be interpreted as an \emph{abstract symbol that represents a description vector}, which is set at an approximately optimal pose in description space -- identified by unsupervised 
classification techniques, like \emph{k}-means or more advanced techniques~\cite{Jain:1999:DCR:331499.331504}. %
The granularity of the dictionary is a major parameter that corresponds to the variation of the object appearance.
Inspired by~\cite{6942984}, the impact of this parameter can be alleviated with a hierarchical quantization which encodes surface characteristics in a coarse-to-fine manner -- see Sec.~\ref{sec:dict}. %

Concerning ii), the methodology of our work is inspired by cognitive and psychology theories on object recognition, which suggest a hierarchical~\cite{Riesenhuber:1999,Leonardis2011,DiCarlo2012} and component-based~\cite{ObjectRecognition_Biederman1987,CERELLA19801,Kirkpatrick-steger98effectsof} representation of object information. 
These hierarchical approaches combine the benefits of so-called \emph{local} and \emph{global-based approaches} as follows. 
Solely local-based classification approaches represent objects as a composition of features like segments or key points to encode object appearances.
The actual constellation of these features is analyzed to infer patterns, which are predominant for certain objects.
Local-based approaches are successfully used in object recognition and also categorization tasks.
Many variations exist, which analyze not only the occurrence of feature constellations but also texture and geometry relationships among features~\cite{6942984,ContextSemanticLabelling-IJRR13,Leibe04combinedobject,Prasad11a,7139358}.
Given a constellation model, the inference is often based on the analysis of local evidences, i.e., in a constrained spatial range w.r.t.~the object (using e.g. Markov Networks~\cite{citeulike:8742196,10.1109/TPAMI.1984.4767596}).
As a consequence, the global shape aspects of objects are insufficiently reflected in local methods. This leads to difficulties to distinguish complex structures, which are only apparent on a global scale. 
Global-based approaches hence represent and analyze objects as single entities, encoding their surface structural properties. 
Accordingly, global approaches (e.g., template-based approaches) can handle complex structures, but partial object observations, e.g., caused by occlusions, may lead to distortions regarding the encoding of the structural properties.
Sophisticated mechanisms~\cite{6751365} have to be applied to enhance the robustness under these circumstances.
As a consequence, partial absence of features due to occlusions or even deformations can be more effectively dealt with using local-based approaches.

Existing work of both approaches have in common that applied descriptors are typically handcrafted, domain-dependent, and fine-tuned -- with the goal to transform the point cloud space into a discrete lower-dimensional space, which often results in a loss of information.
Consequently, sophisticated classification procedures are required to compensate for the loss and to identify distinctive patterns.
Another concept -- artificial neural network-based Deep Learning -- fuses the description and classification process.
In a hierarchical fashion, layers learn local statistical evidences (e.g. from RGB images) which are composed to more abstract evidences on higher layers~\cite{zhang2015fgs-struct,7487310}.
Recently, so-called Geometric Deep Learning~\cite{DBLP:journals/spm/BronsteinBLSV17,DBLP:conf/cvpr/MontiBMRSB17} specifically focuses on learning geometry in non-Euclidean space from manifolds or graphs.
However these deep neural networks are successful and do not require handcrafted descriptors, but they suffer from the demand of large training data and expensive parameterization cost~\cite{7487310,DBLP:journals/spm/BronsteinBLSV17}.

Another related research field focuses on compositional hierarchies~\cite{DBLP:conf/nips/Utans93,FidlerChapter09,DBLP:conf/iccv/OzayAWL15} in which in general geometric entities like edges or contours are hierarchically composed to unions of those.

Inspired by Bag-of-Words, composition and template-based models, and hierarchical abstraction methodology, following \textbf{contributions} are presented in this article: \textbf{i)} A hierarchical quantization of the \emph{description space} (Sec.~\ref{sec:dict}) is introduced to alleviate quantization effects like over-fitting and under-fitting. 
Therein, visual words are learned, which describe surfaces in a coarse-to-fine manner -- from surface primitives to fine-grained individual object appearances. 
This hierarchical abstraction of raw sensor data facilitates the recognition of constellation patterns on a \emph{symbolic} level.
\textbf{ii)} W.r.t.~ the \emph{shape space}, a new shape constellation model is proposed (Sec.~\ref{sec:ch}) that hierarchically decomposes object shapes based on their surface characteristics.
Shape decompositions are \emph{symbolically} expressed in \textbf{i)} and gradually encoded in a local-to-global bottom-up manner -- from single part appearances over part compositions to objects represented as single entities. 
This representation captures facets which we denote as shape motifs, across multiple levels that are exploited as evidences for classification.  
\textbf{iii)} Our approach is data-driven; it is inherently capable of evolving continuously by integrating new shape information. 
The shape category inference drawn is non-invasive and based on single-shots captured from noisy 2.5D point clouds.

\section{Object Instance Representation}\label{sec:refine}
The \emph{correspondence problem} among detected parts and previously observed parts which are associated to a certain meaning such as a category is a major challenge when dealing with noisy data.
Therefore, a robust detection of parts is required, i.e. parts are repetitively and stably detectable so that they can be used as building blocks which constitute objects.
For instance, given a set of various \emph{cans} from different perspectives in 2.5D, the goal is to extract  similar parts from all \emph{cans}, e.g., one planar and one cylindrical part.
Due to the repetitive appearance of planar and cylindrical parts in certain constellations, patterns can be learned, recognized and associated to the shape category \emph{can}.
This inference has to be robust to noisy sensor data as well as partial and occluded object observations.
The object detection process (see subproblems \textbf{1)} and \textbf{2)} in Sec.~\ref{sec:intro}) which segments object candidates often does not require this repetitive detection of parts.  
However to contribute to a robust segmentation that alleviates the correspondence problem, the (over-)~segmented candidate can be refined in order to facilitate a confident shape reasoning for categorization purposes.

\begin{figure}[tb]
  \centering
 \subfigure[$i\mathrm{=}1$]{\label{fig:ref_can1}\includegraphics[width=0.12\linewidth]{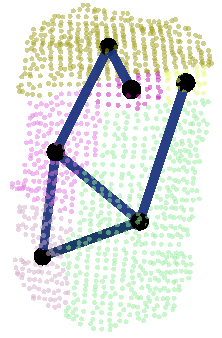}}
 \subfigure[$i\mathrm{=}2$]{\label{fig:ref_can2}\includegraphics[width=0.12\linewidth]{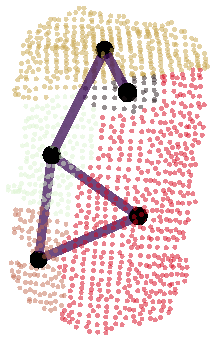}}
 \subfigure[$i\mathrm{=}3$]{\label{fig:ref_can3}\includegraphics[width=0.12\linewidth]{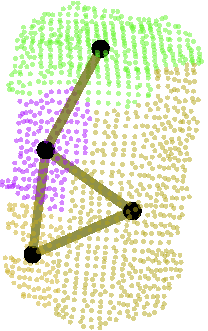}}
 \subfigure[$i\mathrm{=}4$]{\label{fig:ref_can4}\includegraphics[width=0.12\linewidth]{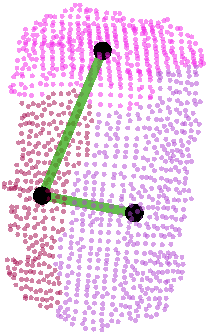}}
 \subfigure[$i\mathrm{=}5$]{\label{fig:ref_can5}\includegraphics[width=0.12\linewidth]{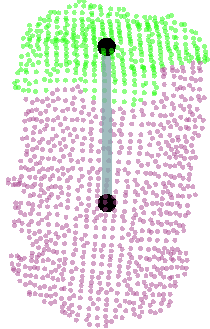}}
 \caption{Segment refinement over 5 iterations ($i$) of an over-segmented $can$.}
 \label{fig:refinement_samples}
\end{figure}

A graph representation is chosen for further analysis of the object topology and of its characteristics, i.e., a graph $g^{os}\mathrm{=}(V^{os},E^{os})$ represents the neighborhood of segments in object $o$ -- see Table~\ref{tab:nocl:inst_rep}.
\begin{table}[tb]
	\centering
	\small
	\caption{Nomenclature -- object instance representation}
	\label{tab:nocl:inst_rep}
	\begin{tabular}{|ll|}
		\hline 
		$P^o$	& Point cloud of object $o$ \\
		$S^o\mathrm{=}\{s_1,s_2, ...\}$ & Point cloud segments of $P^o$\\
		$g^{os}\mathrm{=}(V^{os},E^{os})$& Neighborhood graph of $S^o$ (see Fig.~\ref{fig:ref_can5}):  \\
		&vertices: $V^{os}\mathrm{=}\{v_1,v_2, ...\}$ \\ 
		&\quad\quad\quad\quad with $v_{i}\mathrm{=}\langle s_i\in S^o\rangle$, \\
		&edges: $E^{os}\mathrm{=}\{e_1,e_2, ...\}$ \\
		&\quad\quad\quad with $\text{if } v_i \ue v_j$ (is adjacent) : $e_{k}\mathrm{=}\{v_i,v_j\}$\\ \hline 
	\end{tabular}
\end{table}
Each vertex $v\in V^{os}$ represents an extracted segment. 
Spatially neighboring segments $v_i,v_j\in V^{os}$ are connected with an edge $e_k\in E^{os}$.
In Alg.~\ref{alg:seg_refinement}, the proposed refinement procedure is shown. %
\begin{algorithm}
  \small
  \caption{\small Segment Refinement}
  \label{alg:seg_refinement}   
  \begin{algorithmic}[1]
  \floatname{algorithm}{Procedure}
  \renewcommand{\algorithmicrequire}{\textbf{Input:}}
  \renewcommand{\algorithmicensure}{\textbf{Output:}}
  \REQUIRE Object instance graph $g^{os}=(V^{os},E^{os})$ (e.g. as given in Fig.~\ref{fig:ref_can1}) and merging threshold $\theta$~(e.g. $\theta=0.3$)
  \REPEAT
  \STATE  $is\_merged$ $\gets$ $false$
  \STATE Create list $\Gamma$ of sorted edges in $E^{os}$ w.r.t. segment size in ascending order. Note that, given both connected vertices of an edge, the vertex with the smaller segment size is selected as reference for sorting.
  \FORALL{$e_k \in \Gamma$ \AND $is\_merged = false$}
  \STATE Compute mean surface normal $\mu_i$, $\mu_j$ in border region of segment $v_i$ and $v_j$ connected to edge $e_k\in\Gamma$
  \STATE $\sigma$ $\gets$ compute bounded cosine similarity of $\mu_i$ and $\mu_j$
  \IF {$\sigma$ $<$ $\theta$}
  \STATE Merge segments and update $g^{os}$
  \STATE $is\_merged$ $\gets$ $true$
  \ENDIF
  \ENDFOR
  \UNTIL{$is\_merged = false$}  %
  \ENSURE Refined $g^{os}$ (e.g., result in Fig.~\ref{fig:ref_can5}, if instance in Fig.~\ref{fig:ref_can1} was given as input)
  \end{algorithmic}
  \end{algorithm}
Considering that the signal-to-noise ratio (between segments and noise) is low for small-sized segments, our objective is to minimize the number of small segments by merging them with neighboring larger segments while also considering topological aspects such as surface similarities based on surface normals.
In this merging process small segments are prioritized to facilitate an eventual extraction of segments (see Fig.~\ref{fig:refinement_samples}) which can be semantically meaningful and function as building blocks for further analyses.

\section{Hierarchical Description of Object Segments}\label{sec:dict}

Given a refined object segmentation and the corresponding graph, point cloud segments are projected to a symbol space.
Therein a segment represented as a vertex of an object instance graph is labeled with a symbol.
The process can also be interpreted as a quantization of point cloud data to a constrained set of symbols (a.k.a.~visual words) of a dictionary.
It facilitates reasoning about shape categories on the basis of the analysis of symbol patterns which can be associated to particular shape appearances that may form categories (see Sec.~\ref{sec:ch}).

For the dictionary generation process, surface-structural properties of segments are initially described on the basis of the local pose-invariant descriptor FPFH\footnote{We use the well-known FPFH descriptor as a baseline to illustrate the effectiveness of the proposed shape representation approach.}\cite{5152473}, which relies on the analysis of angle differences among surface normals of 3D points.
In order to receive a global description of an entire segment, a mean histogram is typically created over all local histograms of the point cloud~\cite{5152473}.
This mean histogram forms the \emph{description vector} of a segment point cloud.
The quantization procedure is crucial for the assignment of a visual word to a description vector.
Therein, a word is assigned to a vector with the shortest distance to it using $L^2$-\textit{norm}. %
The objective is to assign similar description vectors to the same visual word.
However, %
a too small set of words is not able to express information variations due to under-fitting of the description space, whereas a too large set of words over-fits the space and hence small description variations can lead to assignments to different words. 
In order to reduce these quantization effects, the feature space is decomposed here in a hierarchical top-down manner. %

\begin{figure}[tb]
  \centering
  \includegraphics[width=0.6\linewidth]{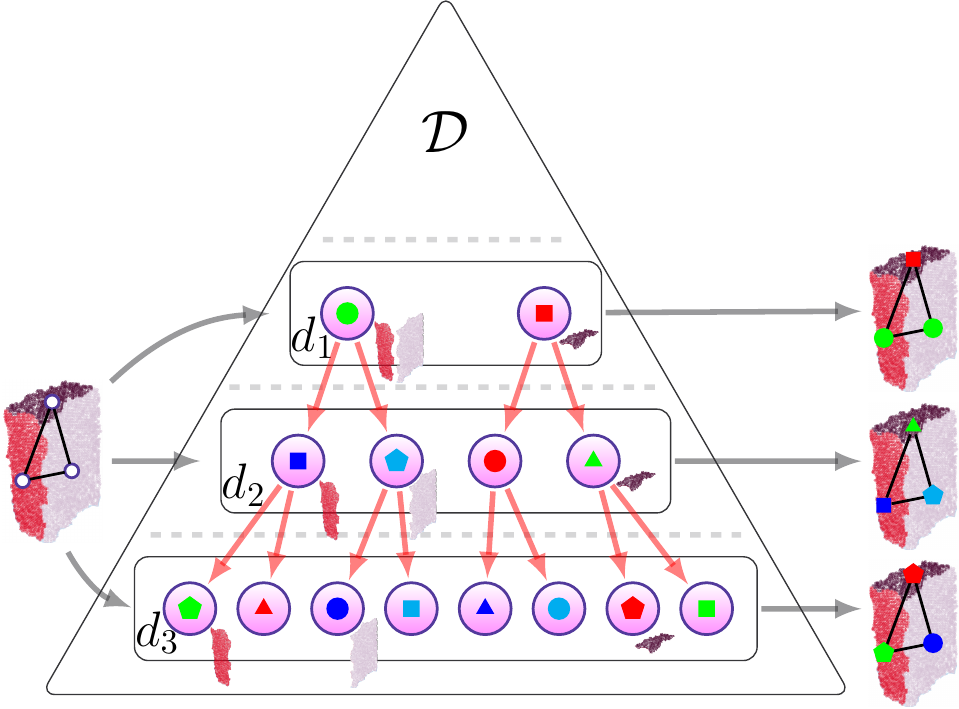}
  \caption{An example hierarchical dictionary $\mathcal{D}$ in which the first three \emph{description levels}~$\{d_1, d_2, d_3,...\}$ are depicted. For illustration, each visual word is depicted as circle containing a colored polygon.}
  \label{fig:dict_illustration}
\end{figure}

Similar to divisive clustering methods~\cite{Jain:1999:DCR:331499.331504}, a set of training description vectors is initially clustered into two child clusters by applying $k$-means, i.e. $k\mathrm{=}2$.
The description vectors that are assigned to the child clusters are further clustered 
in an iterative manner into two for the next level.
Each cluster center represents a visual word and it is assigned to the current level, which we denote as \emph{description level}.
As a result, a tree-like structure is obtained that
leads to a dictionary $\mathcal{D}\mathrm{=}\{d_1, d_2,...,d_n\}$ of $n$ description levels, in which each $d_f$ consists of a set of $2^f$ words ($f$ indicates the description level in $\mathcal{D}$) -- see Table \ref{tab:nocl:dict}; an illustration is shown in Fig.~\ref{fig:dict_illustration}.
\begin{table}[]
	\centering
	\small
	\caption{Nomenclature -- Hierarchical Dictionary}
	\label{tab:nocl:dict}
	\begin{tabular}{|ll|}
		\hline 
		$\mathcal{D}\mathrm{=}\{d_1,..., d_n\}$ & dictionary with $n$ description levels, $d_i\mathrm{=}\{w_1,..., w_{2^i}\}$,\\ 
		&see Fig.~\ref{fig:dict_illustration} \\
		$\kappa(v)\mathrm{=}p$ & point cloud description (FPFH) $p$ of vertex $v\in V^{os}$\\  
		$\omega_f(\kappa(v))\mathrm{=}w$ & visual word assignment $w \in d_f$ to\\
		&point cloud segment of $v\in V^{os}$\\
		\hline 
	\end{tabular}
\end{table}
Similar to our previous work~\cite{6942984}, this quantization of the description space allows to distinguish surface-structural properties in a coarse-to-fine manner. 
Words on lower levels reflect primitive surfaces such as planar and curved shapes, whereas words on higher levels gradually distinguish individual object facets.
\section{Shape Motif Hierarchy}\label{sec:ch}
This section deals with the analysis of object segment constellations for shape categorization.
The Shape Motif Hierarchy approach is proposed, which hierarchically decomposes shape into different granularity levels.
The constellations of decompositions are encoded in a bottom-up manner: from fine-grained segments to groups of segments, which can gradually cover object parts to entire objects.
The objective is to exploit these hierarchical decompositions of an object candidate to detect different topological aspects that function as evidences for certain categories.

\begin{figure}[tb]
  \centering
  \subfigure[Shape motif  hierarchy]{\label{fig:ch_illustration}\includegraphics[width=0.35\linewidth]{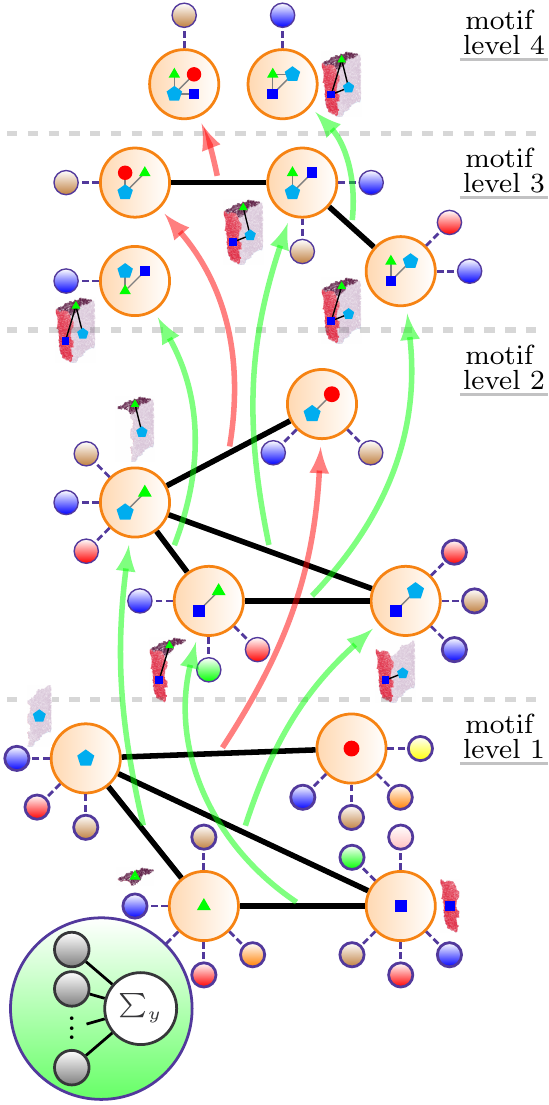}}
  \subfigure[Motif level components]{\label{fig:ch_legend}\includegraphics[width=0.33\linewidth]{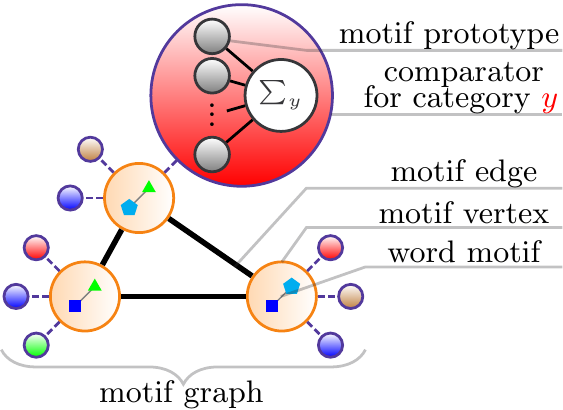}}
  
  \caption{A shape motif hierarchy example is shown in \usubref{fig:ch_illustration}, consisting of multiple \emph{motif levels}. Each node \inlineimage{sample_clique_vertex_a} represents a specific \emph{motif vertex}, whereas each smaller linked node represents a \emph{comparator} \protect \resizebox{.03\linewidth}{!}{\inlineimage{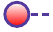}} of a specific shape category $y$ consisting of several \emph{motif prototypes} \protect \samplePattern). A sample propagation of a box \inlineimage{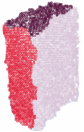} through (\protect \inlineimage{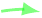}) the hierarchy is shown that consists of three segments (\protect\resizebox{.02\linewidth}{!}{\inlineimage{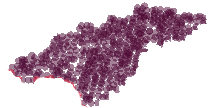}}, \inlineimage{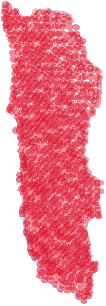}, \inlineimage{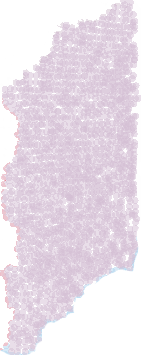}) with corresponding words (\protect\sampleWordC, \protect\sampleWordB, \protect\sampleWordA).
  Feasible propagations which have been previously encoded in the hierarchy during training phase but are not affected by the \emph{box} are depicted as \protect \inlineimage{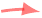}.
  Components of a \emph{motif level} are illustrated in \usubref{fig:ch_legend}.
  }
  \label{fig:dict_illustration_and_ch}
\end{figure}

Given an object instance $o$ represented as a graph $g^o$, each vertex in $g^o$ corresponds to a segment in $o$, subsequently the vertex is labeled with a word $w \in d_f$ ($d_f \in \mathcal{D}$, $f$ indicates the description level in $\mathcal{D}$) according to the corresponding appearance of the segment (see Sec. \ref{sec:dict}).
Our goal is to detect distinctive relations, ranging from relations among particular vertices to subgraphs in $g^o$, which allow inferring the respective shape category $y$ from a set of categories $\mathcal{Y}$.
The shape motif hierarchy allows to encode these relations and provides the capability to analyze the appearance in a hierarchical way. %
A shape motif hierarchy model $\mathcal{H}$, as illustrated in Fig.~\ref{fig:ch_illustration} (see Table~\ref{tab:nocl:ch}),  consists of multiple hierarchical levels in which the observed vertices from an object instance graph are propagated in a bottom-up fashion, beginning from particular segments over composition of segments until the object instance is represented as a single composition. 
\begin{table}[]
	\centering
	\small
	\caption{Nomenclature -- Shape Motif Hierarchy}
	\label{tab:nocl:ch}
	\begin{tabular}{|ll|}
		\hline 
		$g^o\mathrm{=}\{V^o, E^o\}$ & augmented object $g^{os}\mathrm{=}(V^{os},E^{os})$\\
		&$V^o\mathrm{=}\{v^o_1,v^o_2...\}$ where $v^o_i\mathrm{=}\langle s_i\in S^o,p^y_i\mathrm{=}\kappa(s_i),w_i\mathrm{=}\omega_f(p_i)\rangle$\\
		&(for description level $f$, given a label $y$),\\
		&$E^o\mathrm{=}\{e^o_1,e^o_2...\}$ where $\text{if } v^o_j \ue v^o_k$  (is adjacent): $e_{i}\mathrm{=}\{v^o_j,v^o_k\}$ \\
		$\tau^s(v^o_i)\mathrm{=}s_i$& returns $s_i$ of $v^o_i$\\
		$\tau^p(v^o_i)\mathrm{=}p^y_i$& returns $p^y_i$ of $v^o_i$ given a label $y$\\
		$\tau^w(v^o_i)\mathrm{=}w_i$& returns $w_i$ of $v^o_i$\\	
		$\mathcal{Y}\mathrm{=}\{y_1, y_2,...\}$ & set of category labels \\
		$\mathcal{P}\mathrm{=}\{\{\mathcal{P}_y\}_{\forall y \in \mathcal{Y}} \}$& set of labeled motif prototype descriptions, see Fig.~\ref{fig:ch_legend}\\
		$\mathcal{P}_y\mathrm{=}\{p_1^y,p_2^y,...\}$ & motif prototype descriptions of label $y$\\
		$\mathcal{H}\mathrm{=}\{h^1,h^2...\}$ & motif hierarchy, see Fig.~\ref{fig:ch_illustration} \\
		$h^l\mathrm{=}(V^{h^l},E^{h^l})$	&motif graph, $h^l\in \mathcal{H}$ at motif level $l$, see Fig.~\ref{fig:ch_legend}\\ 
		& $V^{h^l}\mathrm{=}\{v^{h^l}_1,v^{h^l}_2...\}$ \\
		& $E^{h^l}\mathrm{=}\{e^{h^l}_1,e^{h^l}_2...\}$  where $\text{if } v_m^{h^l} \ue v_n^{h^l}$: $e_{k}^{h^l}\mathrm{=}\{v_m^{h^l},v_n^{h^l}\}$ \\
		$v^{h^l}_j$& motif vertex $j$ of motif graph $h$ \\
		&at motif level $l$ (see Fig.~\ref{fig:dict_illustration_and_ch},  \inlineimagetable{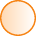})\\
		$v^{h^{l+1}}_j\mathrm{=}\pi(e^{h^l})$ & propagation step for an edge in $h^l$ where \\		
	    &$\pi(e^{h^l})\mathrm{=}\pi(v^{h^l}_m\ue v^{h^l}_n) $\\
	    &\quad\quad \ \ $= \langle \ s^{l+1}_j\mathrm{=}\langle s^{l}_m\cup s^{l}_n\rangle,$ \\
	    &\hspace{1.43cm}$p^{l+1}_j\mathrm{=}\kappa(s^{l+1}_j),$\\ &\hspace{1.35cm}$w^{l+1}_j\mathrm{=}\langle w^{l}_m\cup w^{l}_n \rangle \ \rangle$\\
		$\langle s^{l}_m\mathrm{\cup}s^{l}_n\rangle\mathrm{=}s^{l+1}_j$ & propagation step of segments. Merging of two \\
		& point cloud segments. \\
		$\langle w^{l}_m\mathrm{\cup}w^{l}_n\rangle\mathrm{=}w^{l+1}_j$ & propagation step of words. Union that considers the word\\
		& constellations in  $w^{l}_m$ and $w^{l}_n$.\\
		& E.g.,$\langle w^{l}_m\mathrm{=}$ \sampleWordCliqueB $\cup$\ $w^{l}_n\mathrm{=}$ \sampleWordCliqueA $\rangle \ = \ w^{l+1}_j\mathrm{=}$\sampleWordCliqueC\\
		&where $w$ can be interpreted as a word motif (see Fig.\ref{fig:ch_legend})\\
		& -- initially represented, as a single word, e.g. \protect\sampleWordC, \sampleWordA, \sampleWordB, etc.\\
		$\mathds{1}_{v^{h^l}_j}(g^o)$ & motif vertex Indicator function, where \\
		& $v^{h^l}_j\in V^{h^l},v^o_k \subseteq V^o$ \\
		& {\footnotesize $\begin{cases}
			1, & \text{\footnotesize if }(\tau^w(v^{h^l}_j) \mathrm{=}w_j)\mathrm{\cap}(\tau^w(v^o_k)\mathrm{=}w_k)\mathrm{\neq}\emptyset\\
			 &  \text{\footnotesize i.e. word motif } w_j \text{\footnotesize\ exists in object } g^o\text{\footnotesize.} \\
			 &\text{\footnotesize E.g., }w_j \text{\footnotesize\ in \sampleWordCliqueCompleteA and } w_k  \text{\footnotesize \ in \sampleObjectGraphA match}\\
			 &\text{\footnotesize as shown in see Fig.~\ref{fig:ch_illustration}, motif level 3.} \\
			0, & \text{\footnotesize otherwise}\\
		 \end{cases}$ }\\
		\hline
		\end{tabular}
\end{table}
These compositions are denoted as \emph{motifs}.
Motifs are described and propagated in a symbolic manner using the respective words, which are assigned to the object segments. %
As a result multiple evidences are extracted and analyzed over multiple motif levels, which contribute to a more confident prediction of a shape category compared to purely local-based or purely global-based approaches.
Ergo, objects are fine-grained and locally analyzed on lower motif levels and with increasing motif level, more global semantic properties affect the analysis until the analysis is based on a single entity representing the entire object.
Each motif vertex \inlineimage{sample_clique_vertex_a} in $\mathcal{H}$ represents a unique word constellation which we denote as word motif (e.g.  $w\mathrm{=}$\sampleWordCliqueC), see Fig.~\ref{fig:ch_legend}.
A word motif can be interpreted as a component that can function as a \emph{building block} of an (\emph{unknown}) observed object during the inference phase.
Basically, words can appear in various constellations.
Due to the fact that word constellations which are encoded in $\mathcal{H}$ are only observed from objects, these constellations can be inherently denoted as motifs, respectively, shape motifs in this context that describe objects.

\subsection{Training Phase}\label{sec:ch:tr}
The shape motif hierarchy $\mathcal{H}$ generation is data-driven, i.e., each object instance graph from a given labeled training set of categories $\mathcal{Y}\mathrm{=}\{y_1, y_2,...\}$ is propagated through motif levels in a bottom-up fashion as shown in Alg.~\ref{alg:clique_hiearchy_generation}. 
\begin{algorithm}
	\small
	\def \scalecol{0.7} 
	\caption{\small Shape Motif Hierarchy -- Training Phase
	\newline
		For explanatory purposes, illustrations are provided for a \emph{box} object sample with three segments that represents a graph of three vertices with their corresponding words:
	\newline
	\protect \resizebox{.5\textwidth}{!}{
		$\quad\quad\quad\quad\quad$
		 \inlineimage{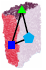}}
	}
	\label{alg:clique_hiearchy_generation}   
	\begin{algorithmic}[1]
		\floatname{algorithm}{Procedure}
		\renewcommand{\algorithmicrequire}{\textbf{Input:}}
		\renewcommand{\algorithmicensure}{\textbf{Output:}}
		\REQUIRE Set of object instance graphs $\mathcal{G}\mathrm{=}\{g^o_1 \dots g^o_n \}$ and empty $\mathcal{H}$ model
		\FORALL{$g^o \in \mathcal{G}$}
		\STATE $y\gets$ label assigned to $g^o$, given by supervision
		\STATE Level $l \gets 1 $
		\FORALL{$v \in V^o$ of $g^o$}
		\STATE \begin{minipage}[t]{\scalecol\linewidth}
			Introduce word $w\mathrm{=}\tau^w(v)$ to $\mathcal{H}$ by augmenting the corresponding motif vertex $v_m^{h^{l}} \in h^{l}$ with $s\mathrm{=}\tau^s(v)$, $p^y\mathrm{=}\tau^p(v)$ and $w\mathrm{=}\tau^w(v_m^{h^{l}})$, see Table~\ref{tab:nocl:ch}. For instance, in the illustration on the right, word \sampleWordA \ representing vertex of segment \inlineimage{box_super_part_b_single} is assigned to motif vertex \inlineimage{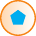}. 
			The prototype description $p^y$ is added to the corresponding comparator \resizebox{.05\linewidth}{!}{\inlineimage{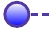}}, where $y$ represents the label \emph{box}.
		\end{minipage}
		$\quad\hspace{0.5cm}$
		\includegraphics[width=0.19\linewidth,valign=t]{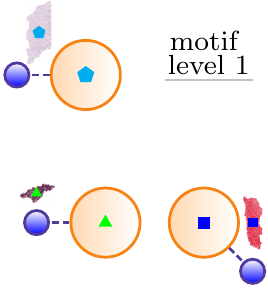}		
		
		\STATE \begin{minipage}[t]{.64\textwidth}
			Create motif edge (\inlineimage{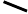}) $e^{h^{l}}$ between $v_m^{h^{l}}$ and a neighboring vertex  $v_n^{h^{l}}$, only if corresponding segments 
			of $v_m^{h^{l}}$ and $v_n^{h^{l}}$ are neighbors in $g^o$. Subsequently, in the illustration on the right, both segments (\resizebox{.045\linewidth}{!}{\inlineimage{box_super_part_a_single}} and \inlineimage{box_super_part_b_single}
			) are neighbors, thus an edge is created between \inlineimage{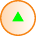} and \inlineimage{sample_clique_vertex_a_w_word2}
			\end{minipage}
			\quad\hspace{0.5cm}
			\includegraphics[width=0.15\textwidth,valign=t]{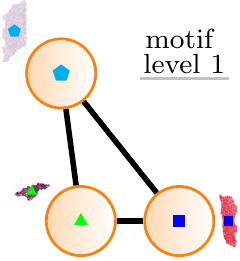}
		\ENDFOR
		\STATE $E^{h^l} \gets get\_edges(h^l)$
		\REPEAT
		\FORALL{$e^{h^{l}}\in E^{h^l}$}
		\STATE \begin{minipage}[t]{.6\textwidth}
	 Propagate \inlineimage{sample_ch_prop_object}  the edge $e$ according to $\pi(e^{h^{l}})$ to receive $v_m^{h^{l+1}}\in h^{l+1}$, see Table~\ref{tab:nocl:ch}. A propagation of the sample \emph{box} object is illustrated as  \inlineimage{sample_ch_prop_object} as well as shown in Fig.~\ref{fig:ch_illustration}.
	\end{minipage}
	\quad
	\includegraphics[width=0.22\textwidth,valign=t]{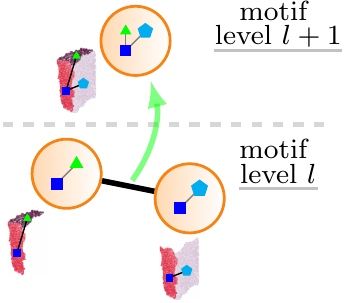}	
		\STATE 
		\begin{minipage}[t]{.6\textwidth}
		Create motif edge (\inlineimage{sample_clique_edge}) $e^{h^{l+1}}$ between $v_m^{h^{l+1}}$ and a neighboring vertex  $v_n^{h^{l+1}}$, only if $v_m^{h^{l+1}}$ and $v_n^{h^{l+1}}$ contain  word(s) representing the same segment(s) of $g^o$. E.g., in illustration on the right, a motif edge is created, since both motif vertices \inlineimage{sample_clique_vertex_a} contain \sampleWordCliqueA which represent the same segments \inlineimage{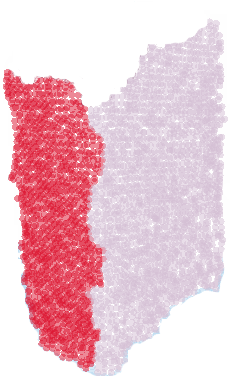} in \inlineimage{box_super}.
		\end{minipage}
		\hspace{0.35cm}
		\includegraphics[width=0.21\textwidth,valign=t]{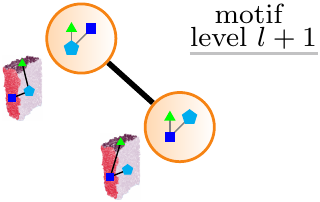}	
		\ENDFOR %
		\STATE $l\gets l+1$ 
		\STATE $E^{h^l} \gets get\_edges(h^l)$
		\UNTIL{$E^{h^l} \neq \emptyset$} %
		\ENDFOR 
		\ENSURE $\mathcal{H}$ model augmented with instances $\mathcal{G}$
	\end{algorithmic}
\end{algorithm}  
Note that, as Fig.~\ref{fig:ch_illustration} suggests, given $m$ motif levels, the motif vertices and edges form a separate motif graph $h^l$ in motif level $l$ of %
$\mathcal{H}\mathrm{=}\{h^1,h^2,...,h^m\}$.
Each motif vertex in $\mathcal{H}$ represents a word motif (see Fig.~\ref{fig:ch_legend}) which has been observed in instance graphs of the training set.
Edges among the motif vertices are created if two motif vertices contain words which correspond to the same segment of the propagated instance (general case: step 12 in Alg.~\ref{alg:clique_hiearchy_generation}).
Given the set of vertices and edges in $\mathcal{H}$ at level $l$, the edges are propagated as vertices to the next higher level $l+1$ (see \inlineimage{sample_ch_prop_object} in Fig.~\ref{fig:ch_illustration} and step 11 in Alg.~\ref{alg:clique_hiearchy_generation}) such that the word motifs of the connected vertices of a motif edge (see \inlineimage{sample_clique_edge} in Fig.~\ref{fig:ch_illustration}) are merged to generate a larger motif, which is used to represent the motif vertex of the higher $l+1$  motif level. 
As a result, the higher the motif level, the greater the word motif order within each motif vertex until on higher motif levels a single motif vertex is generated that can encode a word motif representing entire objects.
Only unique motif vertices are created and propagated, i.e., each vertex contains a unique word motif (see Fig.~\ref{fig:ch_legend}).

Consequently, $\mathcal{H}$ evolves and reflects different facets of the respective category with the propagation of multiple instance graphs.
The rationale behind encoding of object-related information in $\mathcal{H}$ is that (unknown) similar-shaped objects reflect similar propagations through $\mathcal{H}$ across motif levels.
Given an object $o$ represented as a graph $g^o$, two aspects of similarity are considered which will be utilized during the inference phase.

First we consider, the \emph{activation} of a motif vertex, i.e.
a subgraph of words in object $g^o$ matches with a word motif contained in a motif vertex $v$ in $\mathcal{H}$.
In other words, the word motif contained in $v$ is a subgraph in $g^o$. 
This matched word motif occurrence that is denoted as an \emph{activation} is represented with the Indicator function $\mathds{1}_{v}(g^o)$ which returns $1$ if a match is found, otherwise $0$ if no match is found, see Table \ref{tab:nocl:ch}.

Second, the spatial appearance similarity of the matched motif is considered where a motif vertex serves as a comparator, which returns a \emph{stimulus} that reflects the similarity.
Therein, descriptions $\mathcal{P}_y\mathrm{=}\{p_1^y,p_2^y,...\}$, which are observed during the propagation of training instances, are memorized with the corresponding category label $y\in \mathcal{Y}$ in a motif vertex $v$, i.e. $\{\{\mathcal{P}_y\}_{\forall y \in \mathcal{Y}} \}$ descriptions are associated to $v$.
Each description $p_i^y$ represents a description vector of the corresponding point cloud constituting of segments that are observed in instances during training phase and subsequently propagated through $\mathcal{H}$. 
Spatial variations of these constituted point clouds are naturally encoded by the description in the motif vertex.
Inspired by the Prototype Theory~\cite{Rosch1973}, these descriptions can be called as \emph{motif prototypes} of shape appearances associated to the corresponding motif vertex w.r.t.~$y$.
Subsequently, the concept of Probabilistic Neural Networks~\cite{Huang:2004:APN:1011980.1011984} applied as a comparator utilizes the prototype descriptions in $v$ to compute the stimulus -- see Fig.~\ref{fig:ch_illustration} and Fig.~\ref{fig:ch_legend}.
This template-based network does not abstract or generalize the descriptions, consequently avoiding the loss of hidden category-related properties at this stage.
Note that, the prototype descriptions describing the shape of a motif vertex are pose-invariant, which facilitates the correct classification of an object from new viewpoints and the reusability as building block to deal with new shape information of unknown objects.

Similarly to the concept of stimuli in visual-perceptual processes~\cite{Edelman1998-EDERIR,dicarlo:tics_2007,Kriegeskorte2013a}, 
the computed stimuli of an unknown object can be exploited for category inference purposes by the analysis of stimuli patterns of motif vertices across the motif levels (see Sec.~\ref{sec:inference_phase}), since similar object appearances can lead to similar stimuli patterns and vice versa.
\subsection{Inference Phase}
\label{sec:inference_phase}
Given a graph $g^o$ of an unknown query object $o$ which is augmented with the corresponding word for each segment of object $o$, $g^o$ is propagated through $\mathcal{H}$ as described in the training phase.
However, motif vertices and edges are not modified at this stage.
The inference is based on the evaluation of \emph{activations} of motif vertices and the corresponding \emph{stimuli}. %
The evaluation strategy is two-fold: intra-level inference focuses on a particular motif level whereas inter-level inference focuses on the fusion of motif level results.
Concerning the intra-level inference, probabilities are computed during training, which allow to associate motif vertices to certain shape categories, i.e., $P(y|v)$, where $y$ is a specific shape category that is associated to a given motif vertex $v$. 
If an activation of motif vertex $v$ is found, the stimulus is computed using %
the Jenson-Shannon divergence~($JS$) and an adapted Gaussian kernel combination -- see Eq.~\ref{eq:ch_js}, where $v$ denotes a motif vertex, $q$ the respective description of object segments in $g^o$ which match with the word motif of $v$,
$\mathcal{P}_y$ the prototype descriptions associated to $v$ of label $y$, and $\sigma$ (e.g. $\sigma\mathrm{=}0.025$) denotes the bandwidth.
\begin{equation} \label{eq:ch_js}
\alpha(v,g^o,y)\mathrm{=} 
\begin{cases}
    \frac{1}{|\mathcal{P}_y|} \cdot \sum^{|\mathcal{P}_y|}_{i=1}e^{\tfrac{JS(p_i^y \in \mathcal{P}_y,q)^2}{-2\sigma^2}},&\text{\footnotesize if $\mathds{1}_{v}(g^o)\mathrm{=}1$}\\
    0,              & \text{\footnotesize otherwise}
\end{cases}
\end{equation}
The stimulus is normalized by $|\mathcal{P}_y|$ in order to account for an unbalanced distribution of prototype descriptions from particular categories observed during training phase.
This procedure is applied on all activated motif vertices within a motif level as shown in Eq.~\ref{eq:ch_fusion_intra_level}. 
\begin{equation} 
\label{eq:ch_fusion_intra_level}
\begin{split}
 \beta(g^o,y,l) =  \frac{\sum^{|V^l|}_{i=1} \alpha(v^{l}_i,g^o,y) \cdot P(y|v^l_i)}{\sum^{|\mathcal{Y}|}_{j=1} \sum^{|V^l|}_{i=1} \alpha(v^{l}_i,g^o,\mathrm{\emph{y}}_j) \cdot P(\mathrm{\emph{y}}_j|v^l_i)}
 \end{split}
\end{equation}
Consequently, given object graph $g^o$, a normalized response w.r.t.~label $y$ at a motif level $l$ is returned by $\beta(g^o,y,l)$.
In a similar manner, the inter-level inference is drawn by the accumulated responses over all motif levels as shown in Eq.~\ref{eq:ch_fusion_inter_level}, where $m$ is the number of motif levels.
\begin{equation} 
\label{eq:ch_fusion_inter_level}
\begin{split}
 \gamma(g^o, y) = \frac{ \sum^{m}_{l=1} \beta(g^o,y,l) \cdot P(y|l)}{\sum^{|\mathcal{Y}|}_{j=1}\sum^{m}_{l=1} \beta(g^o,\mathrm{\emph{y}}_j,l) \cdot P(\mathrm{\emph{y}}_j|l)}
 \end{split}
\end{equation}
Furthermore, a probability $P(y|l)$ for a shape category $y$ is computed in this case for a given motif level $l$ that regards the shape complexity through the distribution of motif orders observed for objects instances of particular categories in the training.

\section{Shape Motif Hierarchy Ensemble}
Our main goal is to generate various perspectives on an object point cloud that lead to evidences which reveal distinctive patterns which can be used for shape inference purposes. 
 
\begin{figure}[tb]
	\centering
	\includegraphics[width=0.65\linewidth]{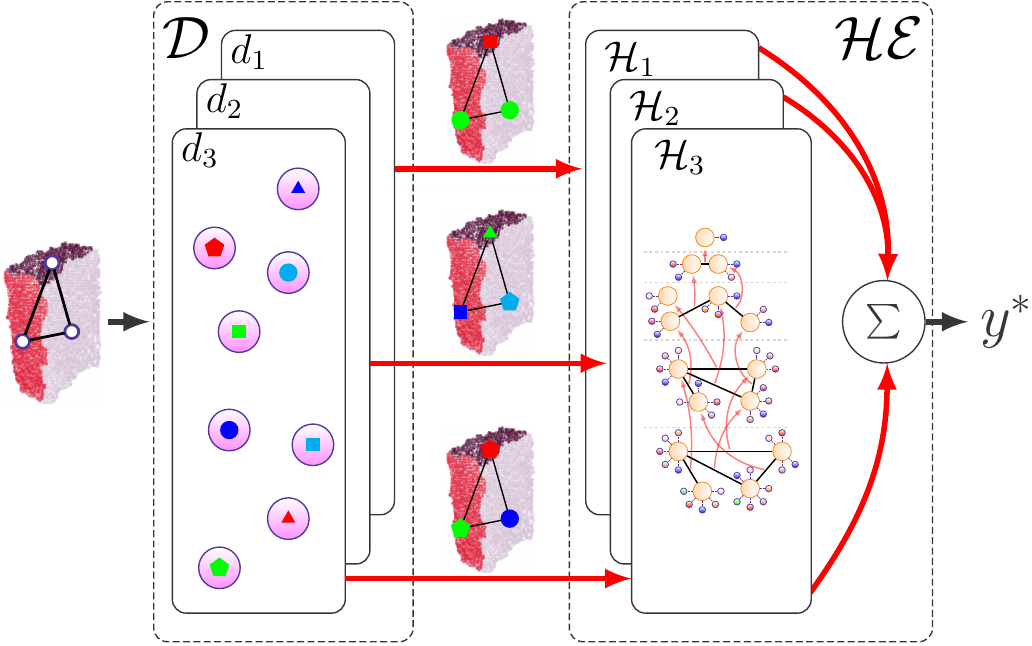}
	\caption{Illustration of an example shape motif hierarchy ensemble $\mathcal{HE}$ based on three shape motif hierarchies $\{\mathcal{H}_1,...,\mathcal{H}_3\}$ (see Fig.~\ref{fig:ch_illustration}) using respective description levels $\{d_1,...,d_3\}$ of $\mathcal{D}$ (see Fig.~\ref{fig:dict_illustration}).}
	\label{fig:hch_illustration}
\end{figure}

A hierarchical dictionary $\mathcal{D}\mathrm{=}\{d_1, d_2,...,d_n\}$ consisting of $n$ description levels was introduced in Sec.~\ref{sec:dict}. 
It allows to generate multiple evidences, e.g. in case of $n\mathrm{=}3$ description levels, the segment \inlineimage{box_super_part_c_single} of the box instance shown in Fig.~\ref{fig:dict_illustration} is represented in level $1$ with the word \sampleWordD, level $2$ with \sampleWordB\ and level $3$ with \sampleWordE. 
As a result for the decomposed shape of the box instance, shown in Fig.~\ref{fig:dict_illustration} and Fig.~\ref{fig:hch_illustration}, three instances graphs can be generated that are augmented with words of the respective description level.
Note that, by the increase of description level more words quantize the description space, i.e. words at higher levels describe more and more specific appearance variations of segments, whereas words at lower levels rather describe general appearances like flat or round segments.
One may interpret that words at lower levels under-fit the space (see Fig.~\ref{fig:hch_illustration}, two segments (\inlineimage{box_super_part_c_single} and \inlineimage{box_super_part_b_single}) at $d_1$ are assigned to the same word \sampleWordD) where as on higher levels over-fit, i.e. different words are assigned to minor appearance variations of segments.

Based on the symbolic representation of point cloud segments with a hierarchical dictionary, a shape motif hierarchy $\mathcal{H}_i$ encodes the shape decompositions and word assignments of object instances in the form of word motif occurrences described by the \emph{particular} set of words from description level $d_i\in\mathcal{D}$.
Subsequently, a Shape Motif Hierarchy Ensemble $\mathcal{HE}\mathrm{=}\{\mathcal{H}_1,...,\mathcal{H}_n\}$ given $n$ description levels is created: one shape motif hierarchy per description level, where object segments which are assigned to words of the respective $d_i$ description level are propagated through the respective shape motif hierarchy $\mathcal{H}_i$ as illustrated in Fig.~\ref{fig:hch_illustration}.

In the training phase, as described in Sec.~\ref{sec:ch:tr} each $\mathcal{H}_i$ evolves and generates evidences in form of word motifs according to observed word occurrences at description level $i$ of training samples.
Note that, in this data-driven fashion, shape motif hierarchies evolve differently due to the different variety of words in each description level, subsequently word motifs vary among shape motif hierarchies and can be exploited as evidences. 
As a result, an ensemble of multiple motif hierarchies $\mathcal{HE}\mathrm{=}\{\mathcal{H}_1,...,\mathcal{H}_n\}$ (see Table~\ref{tab:nocl:che}) allows to span the space of evidences while considering the quantization of the description space of $n$ description levels and the hierarchical shape decomposition within each motif hierarchy.
\begin{table}[]
	\centering
	\small
	\caption{Nomenclature -- Shape Motif Hierarchy Ensemble}
	\label{tab:nocl:che}
	\begin{tabular}{|ll|}
		\hline 
		$\mathcal{HE}\mathrm{=}\{\mathcal{H}_1,\mathcal{H}_2,\dots\}$ & shape motif hierarchy ensemble,  where $\mathcal{H}_i\mathrm{=}\{h^1,h^2, ...\}$ \\
		&using description level $d_i\in\mathcal{D}$\\
		\hline 
	\end{tabular}
\end{table}

In the inference phase, segments of object instance graph $g^o$ are assigned to words of the respective description level and accordingly propagated through the corresponding shape motif hierarchy as illustrated in Fig.~\ref{fig:hch_illustration}.
Based on Eq.~\ref{eq:ch_fusion_inter_level} each shape motif hierarchy $\mathcal{H}_i$ returns an individual shape category $y\in\mathcal{Y}$ response ($\gamma^{\mathcal{H}_i}(g^o, y)$) w.r.t. the observed object instance $g^o$.
As shown in Alg.~\ref{alg:hierarchical_clique_hiearchy_prediction}, these responses are accumulated and fused using majority voting (see Fig.~\ref{fig:hch_illustration}, \inlineimage{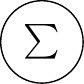}) to identify the final category label $y^*$ for $g^o$.
\begin{algorithm}
\small
\caption{\small Shape Motif Hierarchy Ensemble -- Inference Phase}
\label{alg:hierarchical_clique_hiearchy_prediction}   
\begin{algorithmic}[1]
\floatname{algorithm}{Procedure}
\renewcommand{\algorithmicrequire}{\textbf{Input:}}
\renewcommand{\algorithmicensure}{\textbf{Output:}}
\REQUIRE Object graph $g^o$, shape motif hierarchy ensemble $\mathcal{HE}$
\STATE Propagate $g^o$ through each shape motif hierarchy $\mathcal{H}_i\in\mathcal{HE} $
\STATE Find the final label $y^{*}$ for $g^o$ based on Eq.~\ref{eq:hierarchical_clique_hiearchy_prediction_mv} in which the responses  $\gamma^{\mathcal{H}_i}(g^o, y)$ are computed for the respective $\mathcal{H}_i$ for all $y \in \mathcal{Y}$ (Eq.~\ref{eq:ch_fusion_inter_level})
\ENSURE Corresponding label $y^{*}$ of instance $g^o$
\end{algorithmic}
\end{algorithm}
\begin{equation}
\label{eq:hierarchical_clique_hiearchy_prediction_mv} 
\begin{split}
 y^{*} =  \argmax_{y \in \mathcal{Y}} \frac{ \sum^{|\mathcal{D}|}_{i=1} \gamma^{\mathcal{H}_i}(g^o, y)}{ \sum^{|\mathcal{Y}|}_{j=1} \sum^{|\mathcal{D}|}_{i=1} \gamma^{\mathcal{H}_i}(g^o, \mathrm{\emph{y}}_j)}
\end{split}
\end{equation}
\section{Experimental Evaluation} \label{sec:experiment}
\begin{figure}[tb]
  \centering   
   \def \scaleImg{1.15} 
   \subfigure[]{\label{fig:trainsack}\scalebox{\scaleImg}{\includegraphics[width=0.16\textwidth]{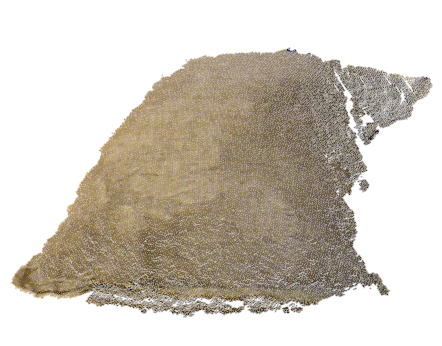}}}
   \subfigure[]{\label{fig:trainbarrel}\scalebox{\scaleImg}{\includegraphics[width=0.1\textwidth]{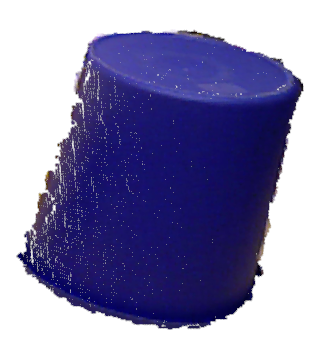}}}
   \subfigure[]{\label{fig:trainparcel}\scalebox{\scaleImg}{\includegraphics[width=0.16\textwidth]{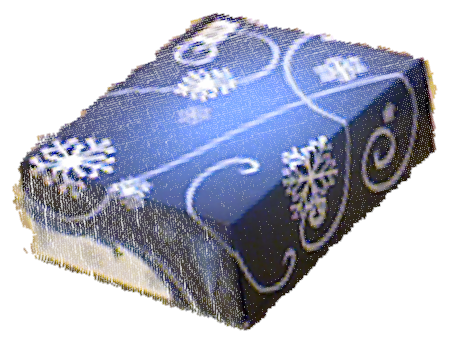}}}
   \subfigure[]{\label{fig:trainteddy}\scalebox{\scaleImg}{\includegraphics[width=0.09\textwidth]{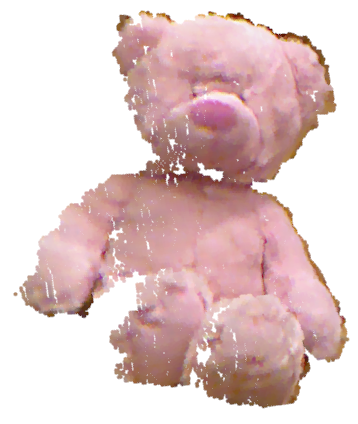}}}
   \subfigure[]{\label{fig:trainball}\scalebox{\scaleImg}{\includegraphics[width=0.1\textwidth]{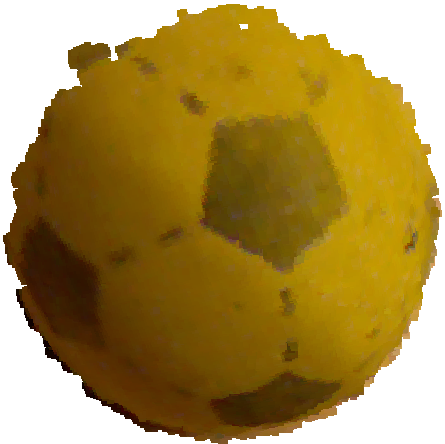}}}
   \subfigure[]{\label{fig:trainamphora}\scalebox{\scaleImg}{\includegraphics[width=0.08\textwidth]{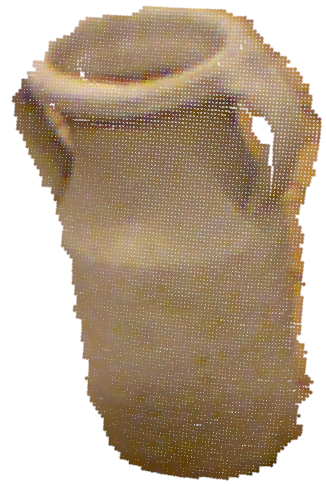}}}
   \subfigure[]{\label{fig:trainplate}\scalebox{\scaleImg}{\includegraphics[width=0.12\textwidth]{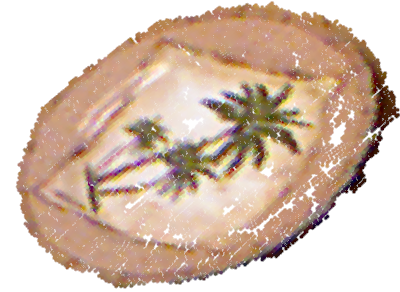}}}
   \caption{2.5D point cloud sample scan of one random object instance per category of the dataset: \emph{sack}~\usubref{fig:trainsack}, \emph{can}~\usubref{fig:trainbarrel}, \emph{box}~\usubref{fig:trainparcel}, \emph{teddy}~\usubref{fig:trainteddy}, \emph{ball}~\usubref{fig:trainball}, \emph{amphora}~\usubref{fig:trainamphora} and \emph{plate}~\usubref{fig:trainplate}.}
   \label{fig:trainexamples}
\end{figure}

\begin{figure}[t]
\centering
\includegraphics[width=0.95\linewidth]{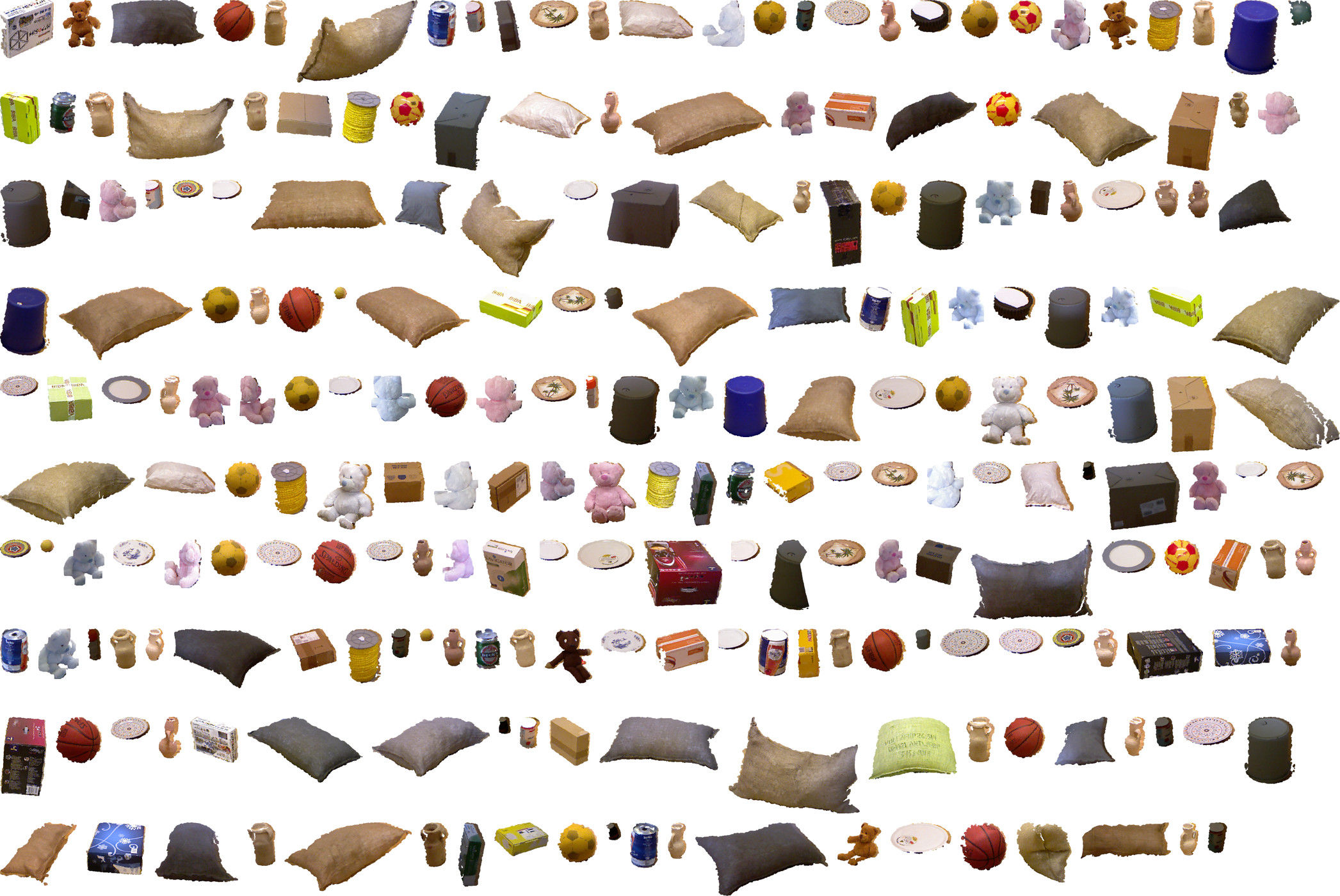}
\caption{A preview of a random subset of unique sample scans (50\% of all scans from each category) of the OSCD dataset.}
\label{fig:thumbnail_db_overview}
\end{figure}

For experimental evaluation, objects from seven different shape categories (\emph{sack}, \emph{can}, \emph{box}, \emph{teddy}, \emph{ball}, \emph{amphora}, \emph{plate}) are used, which show little, partial or strong similarity: e.g., \emph{plates}, \emph{cans}, \emph{boxes} contain flat parts, \emph{balls}, \emph{amphoras}, \emph{teddies}~(head) contain spherical parts, and \emph{cans}, \emph{sacks}, \emph{teddies}~(limbs) contain bulging surfaces.
For this purpose, we created a publicly available dataset, \emph{Object Shape Category Dataset}\footnote{\textbf{http://www.robotics.jacobs-university.de/datasets/2017-object-shape-category-dataset-v01/index.php}} (OSCD), that consists of about $66$ scans per category where each category contains multiple object instances. 
The scans are randomly split into a training/testing set with an average ratio of $75\%$/$25\%$ per category.
A scan is a 2.5D object point cloud (see Fig.~\ref{fig:trainexamples}) which is captured from a random viewpoint with an RGB-D (Kinect-style) camera. %
In Fig.~\ref{fig:thumbnail_db_overview} a preview is shown of a random set of scans captured from dataset objects illustrating the appearance variety regarding scale, deformability and spatial dimensionality.

Fig.~\ref{fig:recognition_eval} illustrates qualitative results of our approach based on two example scenes A and B.  
The object candidates are segmented based on our previous work~\cite{MuellerBirkIcra2016}.
As these are real-world unstructured scenes, the segmented objects tend to be noisy, occluded, and distinctive shape features can be hidden due to the viewpoint.
Candidates representing the background like the wall or the ground are successfully rejected due to low classification confidences ($<30\%$). 
In scene A on the left, one can observe classifications with high confidences, even for the noisily perceived point cloud of the purple \emph{can} or the partially visible blue \emph{teddy}.
In the right scene B, visible parts of the shelf are classified to \emph{box} due to strong shape similarity, but with a lower confidence ($40\%$ and $66\%$).
In the same scene, the occluded \emph{sack}, the non-visible top of the \emph{can} or the missing left handle of the detected \emph{amphora} candidate still lead to a correct classification of the category.

The computationally non-optimized, single-threaded implementation of our method on an \emph{Intel Core i7-3770} machine leads to a mean $2.3$\emph{s} ($\pm$ $2.2$\emph{s}, median $1.4$\emph{s}) runtime per object w.r.t.~the testing set. 
$44.2$\% of the runtime is dedicated to the generation of segment descriptions, whereas $55.8$\% is used for the classification~($\mathcal{HE}$).

\subsection{Hierarchical Description of Object Segments}
\label{sec:exp:dict}
\begin{figure}[tb]
  \centering
  \subfigure[Visual word assigmnent distribution regarding category labels of first three description levels  $\mathcal{D}\mathrm{=}\{d_1, d_2,d_3,...\}$. ]{\label{fig:vw_dict_distri}\includegraphics[width=0.45\linewidth]{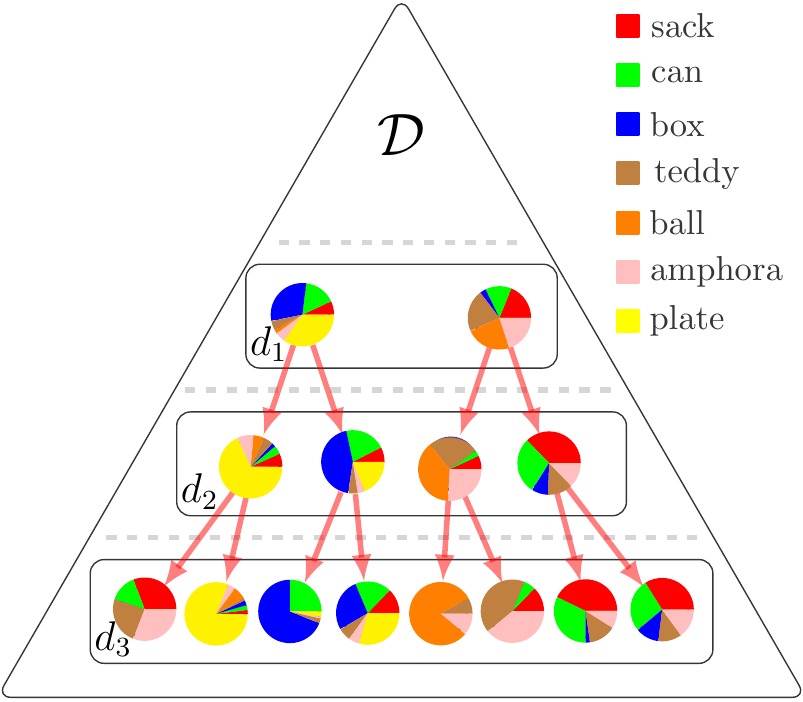}}
  \hspace{0.2cm}
  \subfigure[Motif level size evolution of shape motif hierarchies. ]{\label{fig:ch_vertices_vs_edges}\includegraphics[width=0.45\linewidth]{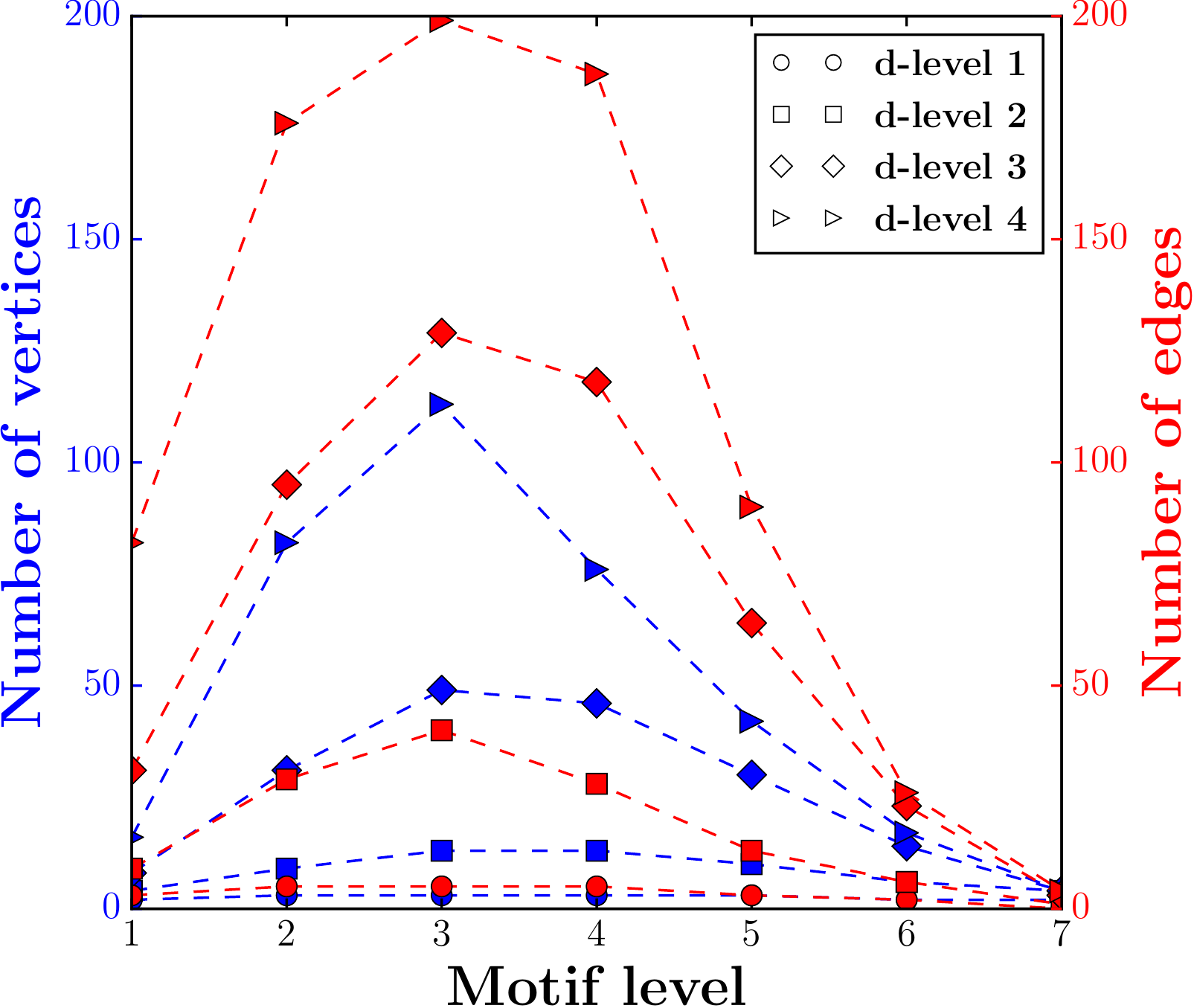}}
  \caption{The first three levels are shown in \protect\usubref{fig:vw_dict_distri}\ \ of a hierarchical dictionary which is trained with $50$ randomly selected segments per category (in total 350 segments). Each node represents a visual word showing the assignment distribution of the segments w.r.t.~the categories.
  In \usubref{fig:ch_vertices_vs_edges}\ \ for each description level (d-level 1 to 4 ) the respective motif hierarchy size (number of vertices and edges per motif level) of the ensemble is shown.
  }
  \label{fig:vw_dict_distri_and_ch_vertices_vs_edges}
\end{figure}

With increasing description level, improved discrimination is expected of the segments w.r.t.~their surface-structural appearance.
In Fig.~\ref{fig:vw_dict_distri}, the first three description levels are shown of a trained dictionary, which consists of seven levels in total.
Each node represents a visual word and shows the assignment distribution of object segments of the respective shape category.
A coarse distinction among segments can already be observed for $d_1$ description level: $65.1\%$ of the visual word representing the left branch is assigned to segments of objects, which are associated to shape categories that feature planar surfaces like \emph{box}~($30.1\%$) and \emph{plate}~($35\%$); whereas, the right branch is mostly assigned to curved instances ($62.8\%$): \emph{sack}~($19.1\%$), \emph{teddy}~($20.8\%$) or \emph{ball}~($22.9\%$).
In further levels a more fine-grained distinction is observable, i.e., certain words are mainly assigned to a particular category or to a small group of categories, e.g., in level $d_2$, \emph{plate}~($67.6\%$, first word from left), \emph{box}~($44\%$, second word) or \emph{sack}~($37.5\%$, fourth word).
In level $d_3$ a clear separation is observable for \emph{plate}~(second word from left), \emph{box}~(third word) and \emph{ball}~(fifth word).
We can conclude that shape characteristics are separable in a top-down and coarse-to-fine manner. 
This separation can be interpreted as an initial (unsupervised) classification step.
\subsection{Shape Motif Hierarchy}
\label{sec:exp:ch}
Given an object instance, the proposed shape motif hierarchy representation aims from the group of object instance segments to discover facets which serve as evidence to infer a category.
In the training phase, the training set is applied to augment $\mathcal{H}$ with shape category-related information.
This process is data-driven (see Sec.~\ref{sec:ch}), i.e. according to the appearances of training instances, $\mathcal{H}$ can continuously evolve and adapt to the appearances. %

In Fig.~\ref{fig:ch_vertices_vs_edges} the evolution of the motif hierarchies is illustrated according to the description levels.
It is observable that the more fine-grained the description space is (increasing number of words), the more motif vertices and edges are generated to encode the individual shape appearances. 
However, a decrease in the number of vertices and edges can be observed at higher motif levels, which can be attributed to the fact that on higher motif levels the variety of word motifs decreases.
Consequently, fewer vertices at higher motif levels sufficiently represent observed instances in the training phase (see Sec.~\ref{sec:ch:tr}). 

\begin{figure}[tb]
  \centering 
  \subfigure[$\mathcal{H}_1$]{\label{fig:bull_ch_assignement_d1}\includegraphics[width=0.425\linewidth]{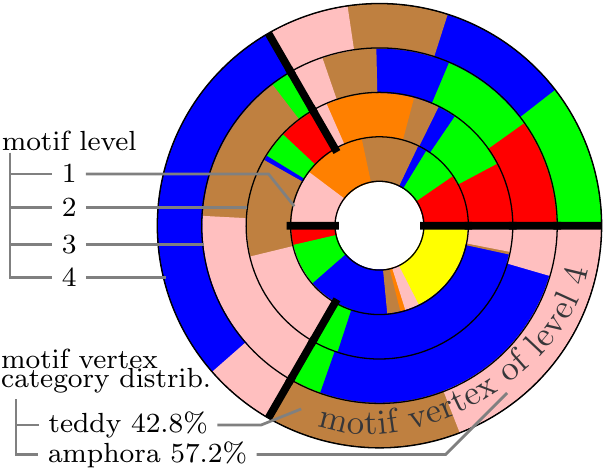}}\hspace{0.2cm}%
  \subfigure[$\mathcal{H}_2$]{\label{fig:bull_ch_assignement_d2}\includegraphics[trim=0 -15 0 0 ,width=0.32\linewidth]{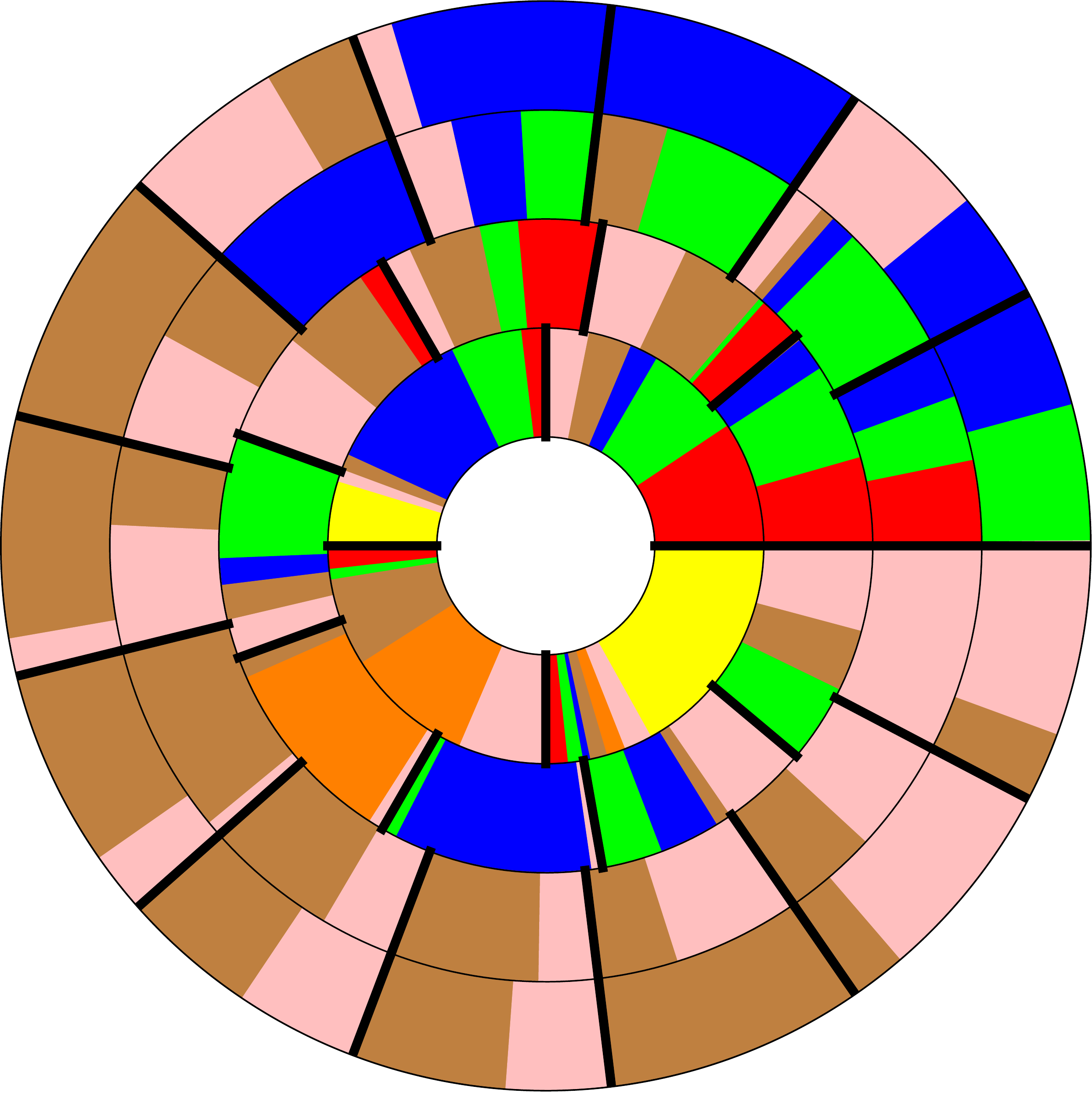}}\hspace{0.2cm}%
  
  \subfigure[$\mathcal{H}_3$]{\label{fig:bull_ch_assignement_d3}\includegraphics[width=0.14\linewidth]{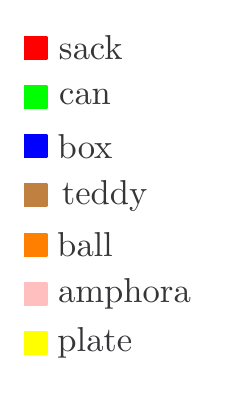}\includegraphics[width=0.32\linewidth]{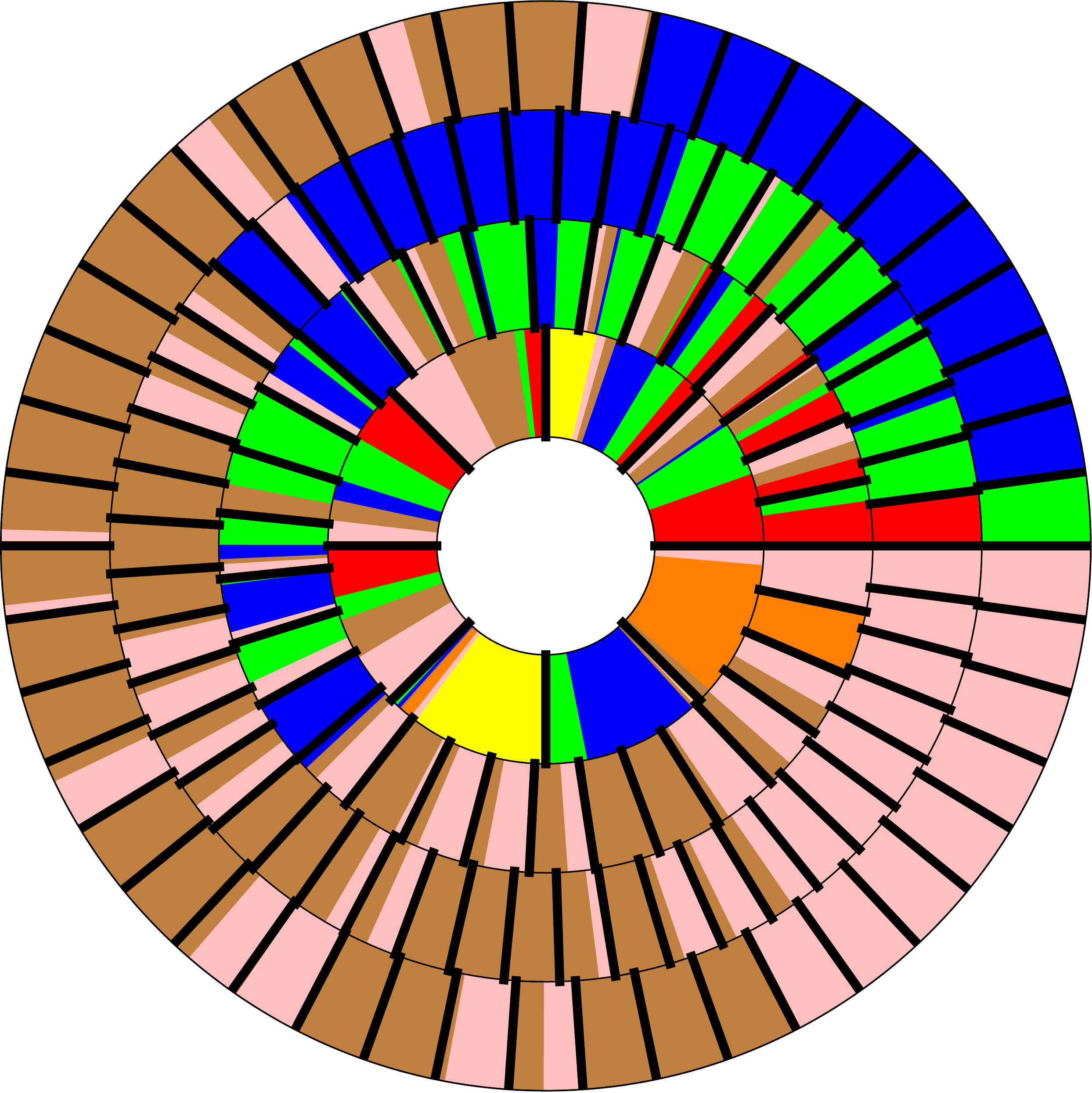}}\hspace{0.2cm}%
  \subfigure[$\mathcal{H}_4$]{\label{fig:bull_ch_assignement_d4}\includegraphics[width=0.32\linewidth]{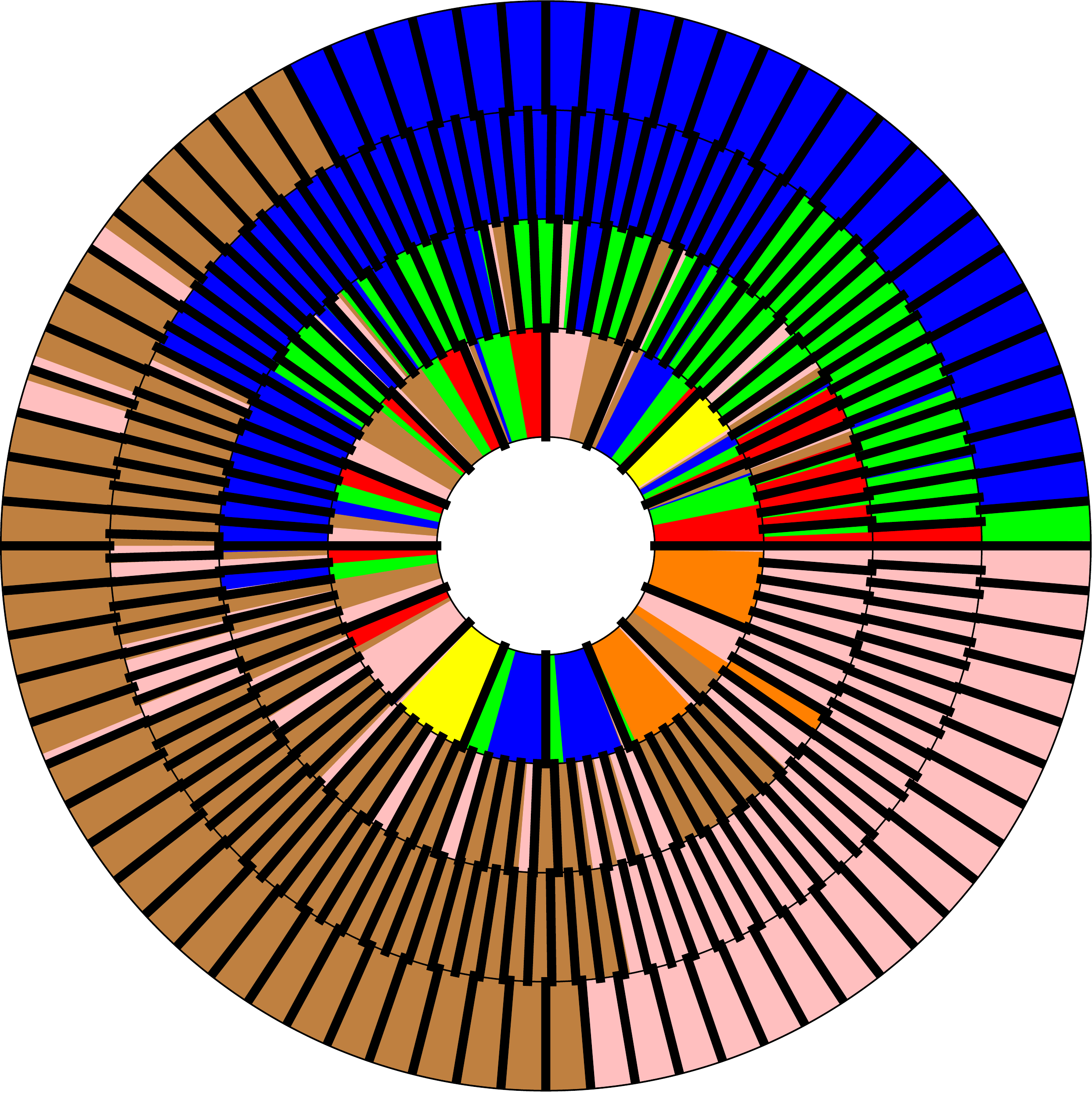}}\hspace{0.2cm}%
  \caption{Category assignment distribution of $\mathcal{H}_{\{1,2,3,4\}}$ of the first four motif levels. Each motif level is represented as a ring; beginning from the inside to the outside. The distribution of a motif vertex is represented as a partition within a ring separated by black bars.
  The size of the colored segments within a partition corresponds to the proportion of a category in the distribution.
  }
  \label{fig:bull_ch_assignement}
\end{figure}

Furthermore, a trained $\mathcal{H}$ is expected to encode distinctive mappings of motif vertices to certain shape categories, which facilitates a confident classification.
In Fig.~\ref{fig:bull_ch_assignement}, the assignment distribution is shown for segments of the shape categories w.r.t. to the motif hierarchies of the first four description levels.
The inner ring reflects the distributions of motif level $1$, that corresponds to single segment occurrences (see Fig.~\ref{fig:ch_illustration} motif level $1$).
It is observable that in the first motif level, curved and planar segments are separated by the motif vertices in $\mathcal{H}_{\{1,2,3,4\}}$.
For instance, at this level in $\mathcal{H}_{1}$ segments of \emph{cans}, \emph{boxes} and \emph{plates} share a high proportion (see motif vertex represented by the lower half of the inner ring) due to the similar appearance of segments -- e.g., top of \emph{can} and side of \emph{box}.
Considering motif vertices reflecting two segments (second ring from center) in $\mathcal{H}_{1}$, \emph{boxes} are clearly separated from instances of other categories, which feature non-planar segments such as \emph{teddies} and \emph{amphoras}.
Note that the higher the motif level, vertices feature a higher correlation with particular categories.
This can be especially observed in $\mathcal{H}_{4}$ for \emph{boxes}, \emph{teddies} and \emph{amphoras}.
The main observation is that word motifs as part of motif vertices show a repetitive activation/occurrence for similar shaped instances. 
Subsequently, motif vertices show a potential for \emph{reusability} and can be used as \emph{building blocks} for \emph{unknown} objects, which represents a key condition for a well performing classification.

\begin{figure}[tb]
  \centering
  \subfigure[Confidence result]{\label{fig:eval:hch_dict_vs_clique_confidence}\includegraphics[width=0.45\linewidth]{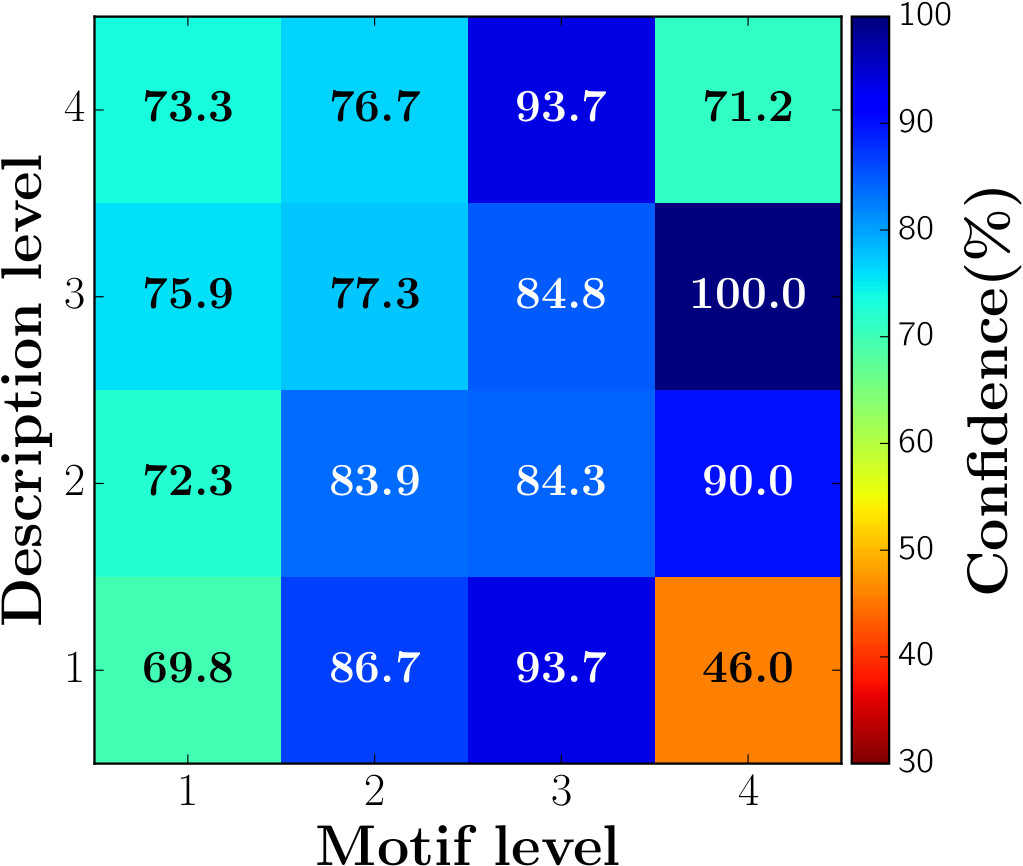}}  
  \subfigure[Classification result]{\label{fig:eval:hch_dict_vs_clique_error}\includegraphics[width=0.45\linewidth]{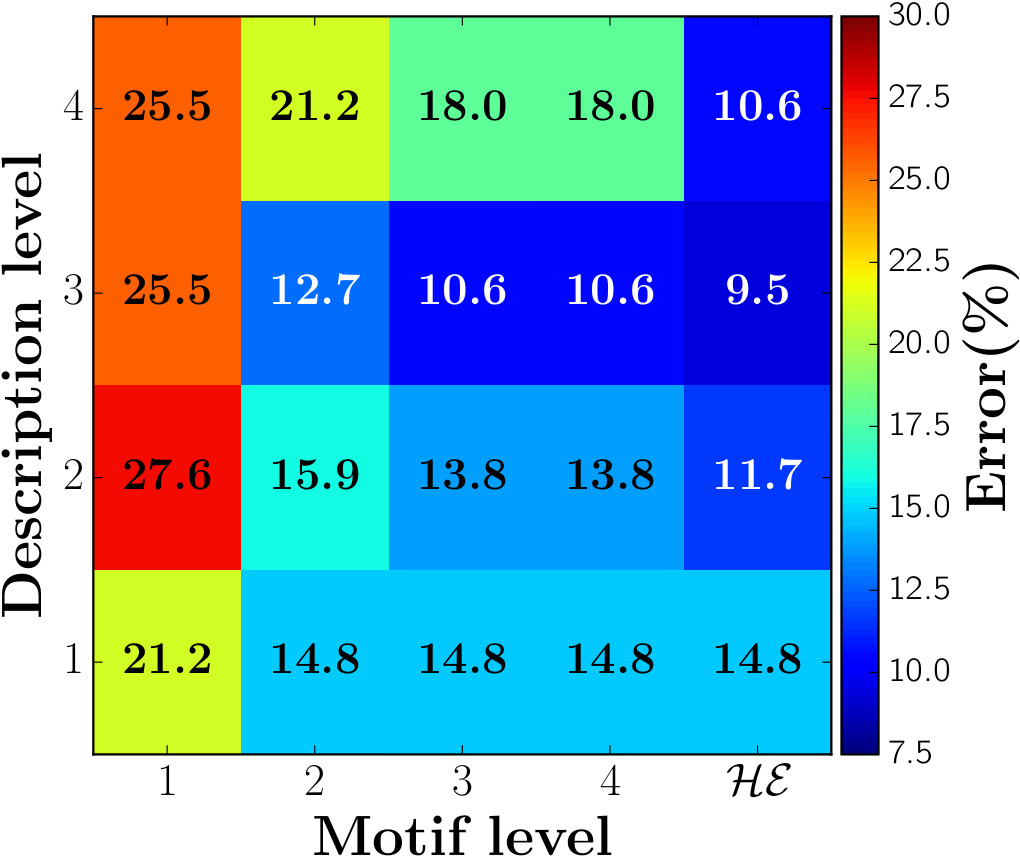}}
  
  \subfigure[Ranked classification result]{\label{fig:eval:hch:ranked_class}\includegraphics[width=0.49\linewidth]{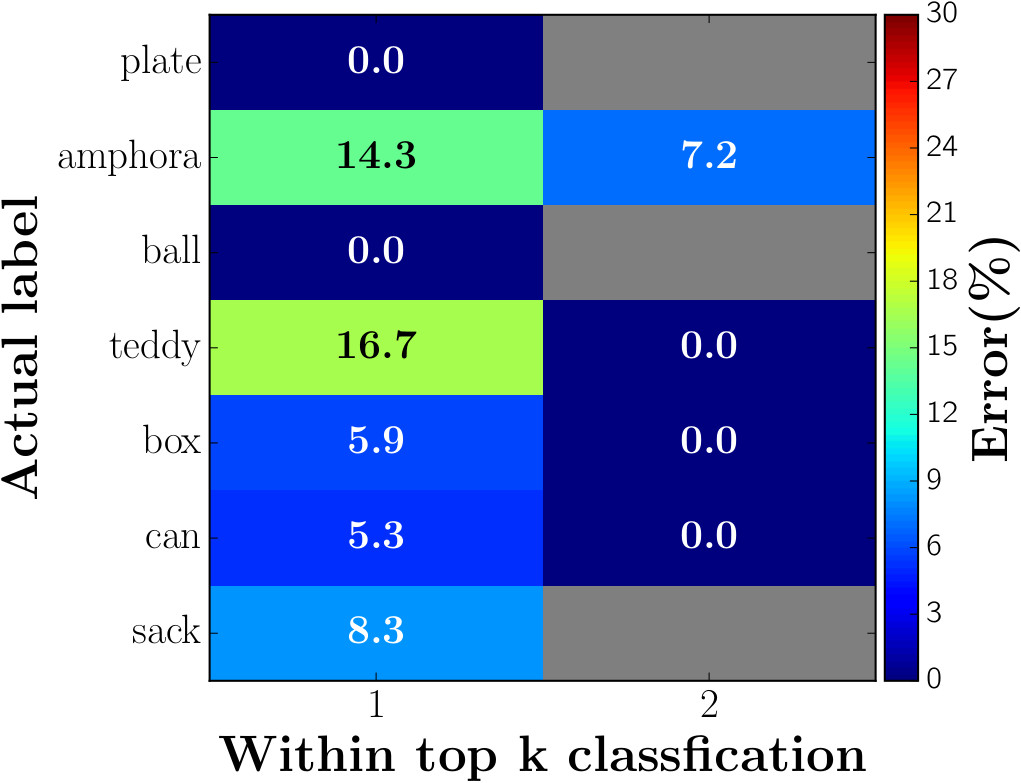}}
  \subfigure[Classification rate]{\label{fig:hch:classification}\includegraphics[width=0.49\linewidth]{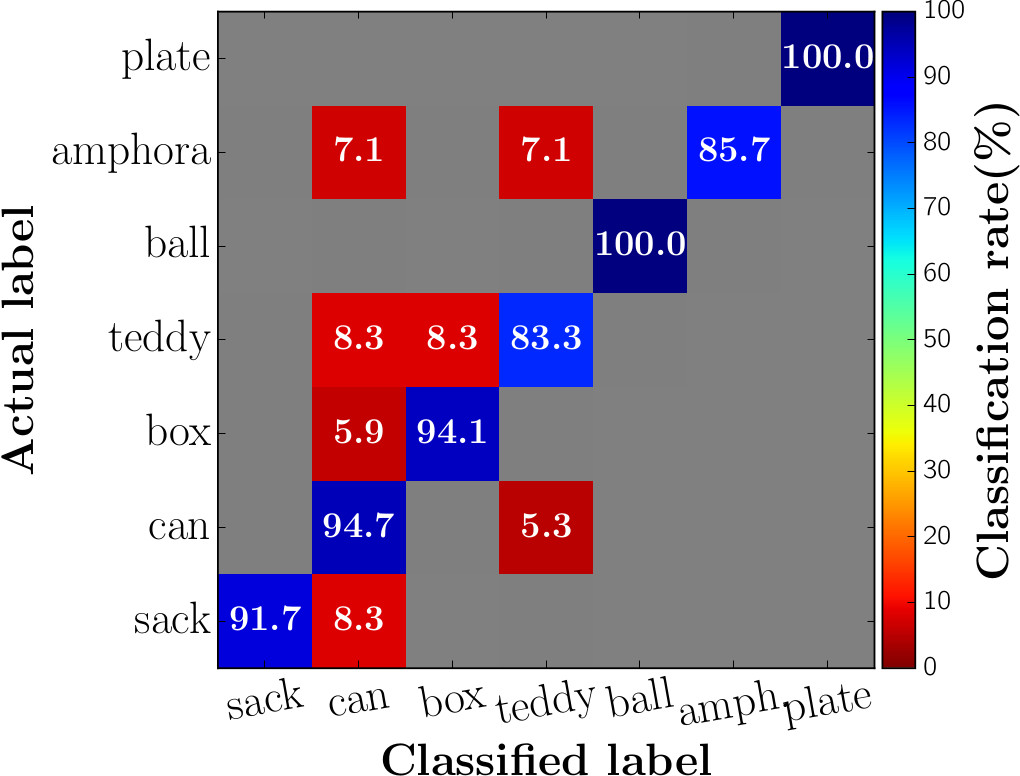}}
  \caption{Mean classification results (of 5 repetitions) of the description and motif levels w.r.t. the testing set (gray marked cell = no result is evident).} 
  \label{fig:recognition_error}
\end{figure}

In Fig.~\ref{fig:eval:hch_dict_vs_clique_confidence} the mean confidences of testing set classifications are shown of motif levels from each motif hierarchy  $\mathcal{H}_{\{1,2,3,4\}}$, respectively, description level.
The classification confidence of a motif level is the response of the direct comparison between a query object and the prototype descriptions of motif vertices of the respective motif hierarchy as shown in Eq.~\ref{eq:ch_fusion_intra_level} for the given label $y$ of a testing instance.
It can be interpreted that high confidences correspond to a high similarity to the model encoded in the respective motif hierarchy.
By increasing motif level, the confidence generally increases, i.e., the more evidences are observed in the form of larger word motifs, which cover an object by a greater extent, the more confident are the motif hierarchy responses. %
However, at high motif levels, for instance at level $4$, the confidence drops in case of description level $1$ and $4$, i.e. for $\mathcal{H}_1$ and $\mathcal{H}_4$. %
A low confidence can be mainly explained by the dissimilarity to the prototype descriptions encoded in motif hierarchy, which lead to low stimuli.
These observations are expected: a motif hierarchy encodes the description in a general-to-specific manner w.r.t.~the motif levels, i.e., lower levels can represent generic building blocks whereas higher levels reflect specific appearances of objects; if such specific appearances lead to a high confidence we can interpret the result as an instance recognition rather than as a categorization result.
\subsection{Shape Motif Hierarchy Ensemble}
\label{sec:exp:hch}
Given a set of $n$ description levels $\mathcal{D}\mathrm{=}\{d_1,...,d_n\}$, a motif hierarchy is trained for each description level and integrated to the ensemble  $\mathcal{HE}\mathrm{=}\{\mathcal{H}_1,...,\mathcal{H}_n\}$.
The dictionary and motif hierarchies are closely interrelated. 
In Fig.~\ref{fig:eval:hch_dict_vs_clique_error}, the effects are shown of motif level and description level on the classification accuracy.
Considering each $\mathcal{H}$ independently, the accuracy generally improves over the increase of motif levels.
For $\mathcal{H}_1$ a low increase of accuracy can be found due to under-fitting symptoms as the description level $1$ consists of only two words -- see Fig.~\ref{fig:vw_dict_distri}. 
Although, motifs in $\mathcal{H}_1$ consist of vertices that are represented by two words, a classification error of only $14.8\%$ is already achieved.
Similarly for $\mathcal{H}_4$, over-fitting symptoms are found on higher motif levels due to the larger word variety, which leads to larger motif vertex variety compared to hierarchies that base on lower description levels. %
The integrated results in the form of $\mathcal{HE}$ show that it outperforms the individual motif hierarchies. 
Furthermore, excluding the description level $4$, which is affected by over-fitting symptoms, the error gradually decreases from $14.8\%$ to $9.5\%$ over the first three description levels w.r.t.~the testing set.
This shows that
the hierarchical shape decomposition of object instances with multiple description levels allows to collect evidences that facilitate a confident category inference. %

Shape categories can share similar surface-structural properties that do not allow a clear distinction.
Considering also uncertainty caused by viewpoint variations, instances of different categories can appear similar, e.g., a \emph{sack} and a \emph{teddy}.
In Fig.~\ref{fig:eval:hch:ranked_class} the ranked classification results are shown.
While \emph{plates} and \emph{balls} are clearly distinctive, since they are classified as first choice (by rank), \emph{cans}, \emph{boxes} and \emph{teddies} are correctly classified within the first two ranks, whereas \emph{sacks} and \emph{amphoras} are not classified correctly -- $8.3\%$ of \emph{sacks} and $7.2\%$ of \emph{amphoras} are misclassified.
This observation is also reflected in the first-rank classification rates shown in Fig.~\ref{fig:hch:classification}.
It shows that instances of categories can be misclassified to categories, which share similar shape structures like planar, cylindrical, bulging surfaces etc.
It is expected that only using shape information, under certain conditions like viewpoint and self-occlusions, such misclassifications are encountered due to shape similarity.

\subsection{Comparison to Alternative Methods}
\label{sec:comparison}
\begin{figure}[tb]
  \centering %
  \subfigure[Vocab. tree~\cite{shen2016graph}]{\label{fig:eval:comat_voc}\includegraphics[height=0.3185\linewidth]{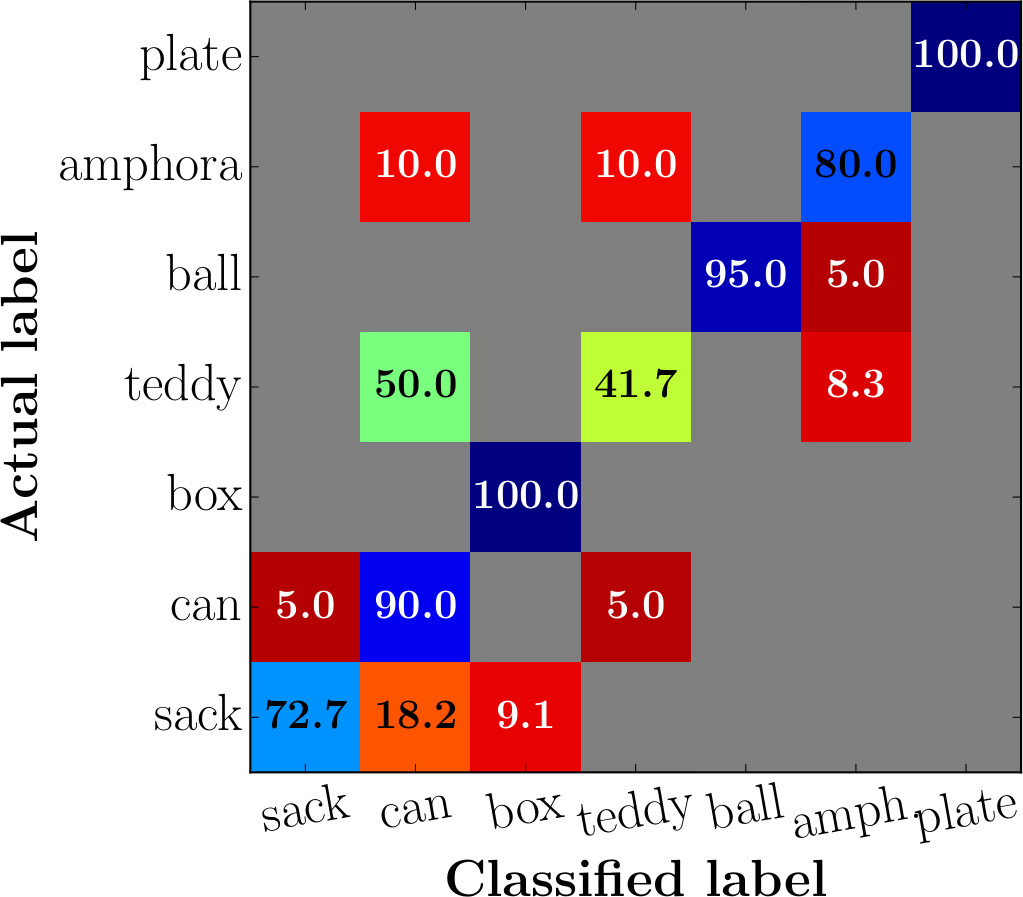}}
  \subfigure[FPFH~\cite{5152473}]{\label{fig:eval:comat_fpfh}\includegraphics[height=0.3185\linewidth]{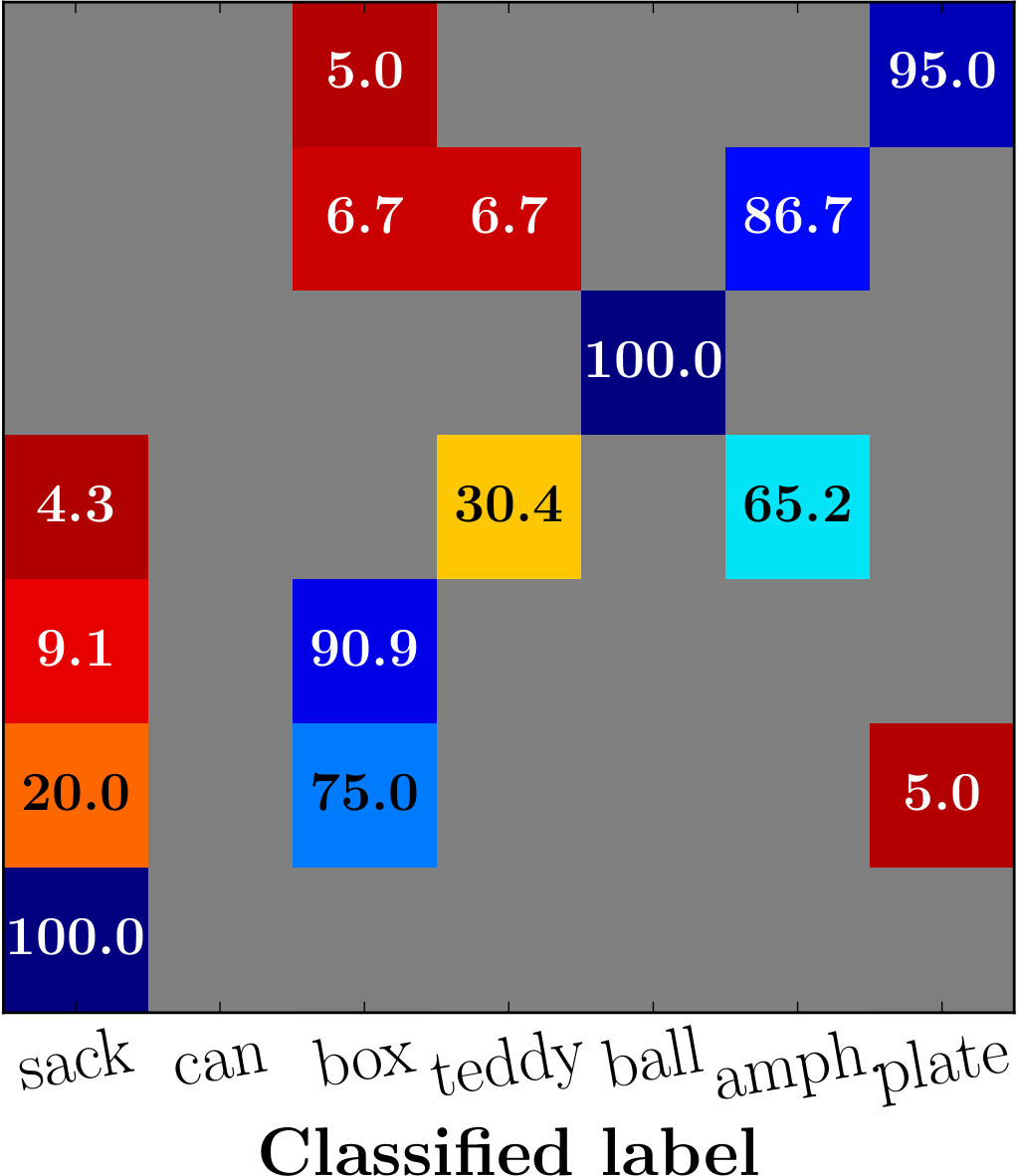}}
  \subfigure[Deep convolutional neural network~\cite{NIPS2012_4824}]{\label{fig:eval:comat_dl}\includegraphics[height=0.3225\linewidth]{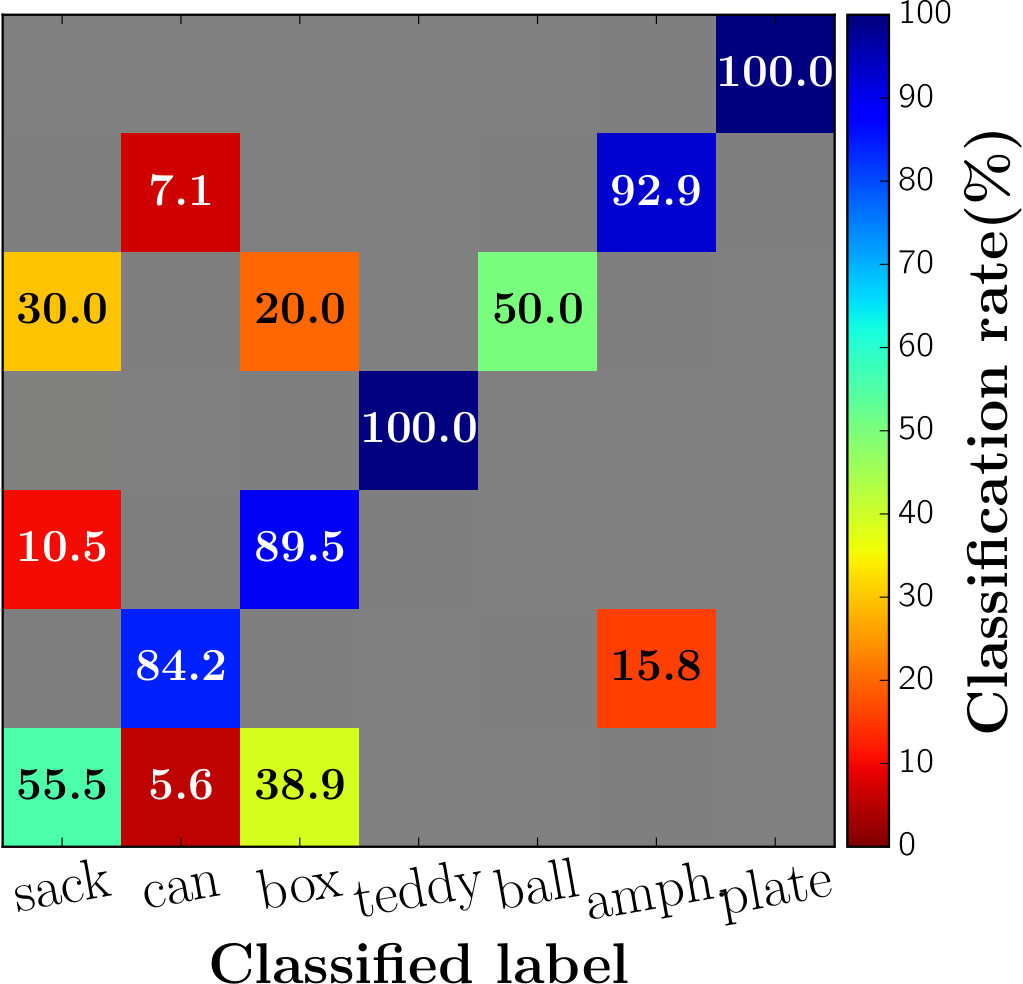}}
  \caption{Mean classification rates (of 5 repetitions) of other approaches w.r.t. the testing set (gray marked cell = no result is evident). A mean error of $13.8$\% for the vocabulary tree,  $29.1$\% for the FPFH and $16.7$\% for the deep convolutional neural network approach have been achieved.} 
  \label{fig:eval:comat}
\end{figure}

For comparison of the categorization results, three alternative methods (Vocabulary Tree, FPFH, and Deep Learning) are evaluated using our training and testing set.
The vocabulary tree~\cite{shen2016graph} is trained with extracted description vectors (see Sec.~\ref{sec:dict}) of the dataset using a tree depth of $6$ and a branch number of $8$.
A $13.8$\% testing set error is achieved (Fig.~\ref{fig:eval:comat_voc}).
In Fig.~\ref{fig:eval:comat_fpfh} a category discrimination is shown that is solely based on the FPFH descriptor: 
for each description of a query object, a nearest-neighbor search on the descriptions of the labeled training instances is performed. 
A mean distance for each category label is computed there based on the nearest-neighbor distances using $L^2$-\textit{norm}. 
The inferred label of the query is the category with the smallest mean distance. %
This simplistic comparison of descriptions does not lead to satisfying results ($29.1$\% testing set error) due to description ambiguities across locally similar shaped instances from different categories. 
In comparison (Fig.~\ref{fig:eval:comat_dl}), a deep convolutional neural network (CNN)~\cite{jia2014caffe}
with the point cloud instances as input in form of range images~\cite{7487310} reaches a $16.7$\% testing set error which saturated within maximum 1000 training iterations of the CNN. %
The AlexNet~\cite{NIPS2012_4824} architecture is chosen for the CNN and solely trained with our dataset.
It consists of $5$ convolutional layers and $3$ succeeding fully connected layers. 
The final softmax layer is reduced to a size of $7$ corresponding to our $7$ categories of the dataset.
The relatively low amount of training samples is likely to have contributed to the error (as discussed in \cite{7487310}), which shows that our approach can deal with such conditions and provide a lower $9.5$\% testing set error (see Sec.~\ref{sec:exp:hch}) in the experiment.
\subsection{Comparison to Alternative Datasets}
\label{sec:alter_db}
\begin{figure}[tb]
	\centering 
	\def \scaleImg{0.9} 
	\subfigure[]{\label{fig:eval:instance_variety0}\scalebox{\scaleImg}{\includegraphics[height=0.2\linewidth]{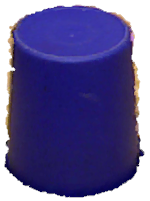}}}
	\subfigure[]{\label{fig:eval:instance_variety1}\scalebox{\scaleImg}{\includegraphics[height=0.17\linewidth]{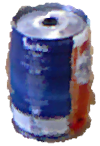}}}
	\subfigure[]{\label{fig:eval:instance_variety2}\scalebox{\scaleImg}{\includegraphics[height=0.06\linewidth]{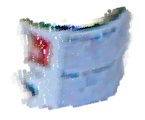}}}
	\subfigure[]{\label{fig:eval:instance_variety3}\scalebox{\scaleImg}{\includegraphics[height=0.09\linewidth]{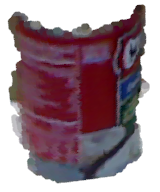}}}	
	\subfigure[]{\label{fig:eval:instance_variety4}\scalebox{\scaleImg}{\includegraphics[height=0.11\linewidth]{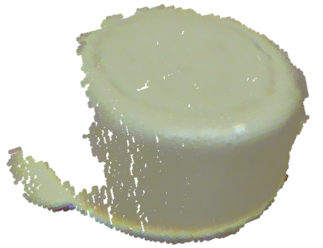}}}
	\subfigure[]{\label{fig:eval:instance_variety5}\scalebox{\scaleImg}{\includegraphics[height=0.12\linewidth]{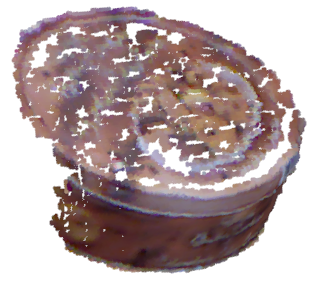}}}
	\caption{Illustration of appearance variations of sample point clouds from different distributions (datasets): \usubref{fig:eval:instance_variety0}, \usubref{fig:eval:instance_variety1} show \emph{can}~\emph{0} and \emph{56} of the OSCD-training set, whereas \usubref{fig:eval:instance_variety2} and \usubref{fig:eval:instance_variety3} show \emph{food\_can\_1\_1\_1} and \emph{food\_can\_14\_1\_1} of the \emph{Washington RGB-D Dataset}~\cite{5980382} and \usubref{fig:eval:instance_variety4}, \usubref{fig:eval:instance_variety5} show cylindrical instances from scenes \emph{learn 34} and \emph{test 42} of the \emph{Object Segmentation Database}~\cite{6385661}.} 
	\label{fig:eval:instance_variety}
\end{figure}

When dealing with supervisedly generated datasets several aspects have to be considered.
\textbf{1) Generalization}: conventional experimental evaluation, in particular in case of instance recognition, is generally performed with a dataset in which samples are drawn from a particular \emph{distribution}, i.e. the conditions of gathering samples are kept same regarding capturing procedure including used sensor, viewing perspective, type of clutter or occlusion, etc.
An illustration of dataset distribution variations is shown in Fig.~\ref{fig:eval:instance_variety}; these instances are labeled as \emph{can} or \emph{cylindrical} respectively.
Therefore the evaluation outcome of a method which is applied on a dataset, is mainly correlated to specific properties and conditions of the dataset.
Consequently, methods have to be tuned for specific datasets to comply and to achieve competitive results.
If not tuned, such evaluations will erroneously lead to disadvantages results even though the
design and concept of the untuned method is superior compared to tuned methods that provide better results for the specific dataset.
Note that, such tuning contradicts with the major objective of creating a method which is capable to cope with various conditions which have not been even observed during training phase. This generalization capability is  essential in many real-world scenarios~\cite{JonschkowskiEHM16,7553531}.
\textbf{2) Supervision}: samples of datasets are generally sampled and labeled in a supervised manner by humans. 
Human-based supervised labeling bears the risk of incorporating additional abstraction and model knowledge gained from experience.
Due to human individuality such labeling process does not guarantee an unique association of samples to a specific category~\cite{McCloskey1978}, particularly in case of ambiguous object observations caused by viewpoint, occlusions, etc.
Furthermore, humans perceive objects through the integration of a composition of various modalities~\cite{Seward97a,Palmeri2004a,ZimgrodHommel2013}: visual, auditory, tactual sensations, or semantic knowledge which fused together may allow to label a \emph{mug} not only by shape but also by material and functional knowledge like that a concave object represents a container for liquids which can be attributed to \emph{mugs}.
However this knowledge is not reflected in datasets with an unary classification label (e.g. instance or category label) of the sample, for instance Fig.~\ref{fig:eval:instance_variety3} shows only a half cylinder without a top or bottom part, nevertheless it is labeled as \emph{can} but could also be a part of a \emph{mug}, \emph{bottle} or limb of a \emph{teddy}.
As a consequence, one may suggest that an unary label does not allow to sufficiently describe a sample -- it is not expressive enough. 
Consequently, datasets whose purpose is the evaluation of a specific cue (e.g. visual reasoning of geometry or shape) are during the labeling process highly vulnerable to incorporate knowledge which is not inferable from the given data (e.g. point clouds).
Especially in case of 2.5D datasets~\cite{5980382,6385661} ambiguities in the classification occur where a \emph{hard-classification}, i.e. labeling a sample to a unique instance, is not feasible due to viewpoint, occlusions, etc.
\textbf{3) Data-driven learning}: supervised approaches are generally trained in a data-driven manner, i.e. pre-grouped samples w.r.t. specific labels are exposed to the method. 
Consequently, approaches only encode the observations made in the training phase (often in form of RGB images or point clouds) irrespective of semantic meaning or correlations among labels. 
Therefore the generated prediction model may only partially cover the variety of possible sample appearances regarding particular labels.
This aspect plays even a more important role in categorization tasks where an explicit object model-to-predict is absent.
As a result, an unique object-to-label assignment is often not feasible due to shape ambiguities compared to instance recognition task where a specific object model-to-predict is given beforehand.

\medskip \noindent
Under these circumstances, the following experiment is focused on the generalization ability of the proposed shape motif hierarchy ensemble method. 
We aim to decouple the semantical meaning of the dataset labels which our method is trained with and aim to investigate the ensemble response behavior with other datasets.

Initially the shape motif hierarchy ensemble $\mathcal{HE}$ is trained \textbf{once} with the training set of our OSCD dataset containing the seven categories.
Instances from other datasets, the \emph{Washington RGB-D Object Dataset}\footnote{This dataset contains of 51 categories which can be understood as \emph{semantic} categories; e.g., instances of the \emph{calculator} category, \emph{food box} category or the \emph{cereal box} category share all a very similar shape, namely the shape of a \emph{box}, i.e. a shape categorization approach should not differentiate those instances rather it is aimed to overcome these shape variations and to classify those instances to their common shape category \emph{box}.}~\cite{5980382} (WD) and \emph{Object Segmentation Database}~\cite{6385661} (SD), are propagated through $\mathcal{HE}$ without retraining; note that all three datasets are sampled from different distributions as above explained and illustrated in Fig.~\ref{fig:eval:instance_variety}. 
In order to account for ambiguous classifications due to sample appearance variations in different datasets and to allow to analyze the spectrum of responses for an unknown instance, 
an instance  $g^o$ is not only classified by the predicted unary label according to Eq.~\ref{eq:hierarchical_clique_hiearchy_prediction_mv}, in this experiment we represent $g^o$ as the set of category responses $g^o\mathrm{=}\{r_1, r_2,...\}$ (see Eq.~\ref{eq:stimuli_response_space}), where $r_i$ is the response of label $y_i$ ($y_i\in \mathcal{Y} \wedge |\mathcal{Y}|\mathrm{=}7$, given seven categories).

\begin{equation}
\label{eq:stimuli_response_space} 
\begin{split}
 r_i = \delta(g^o,y_i) = \frac{\sum^{|\mathcal{D}|}_{j=1} \gamma^{\mathcal{H}_j}(g^o, y_i)}{|\mathcal{D}|}
\end{split}
\end{equation}
To infer the actual degree of response w.r.t. $y$ we do not normalize over $\mathcal{Y}$.
Each label response $r_i$ can be interpreted as a \emph{stimulus} which is learned in a data-driven manner from the supervisedly grouped set of samples that is the training set of our OSCD dataset. 
As a result each instance is represented in a $|\mathcal{Y}|$-dimensional space encompassing stimuli.
This procedure can be denoted as \emph{soft-classification}.
In this way, the generalization capability can be investigated by the created  $|\mathcal{Y}|$-dimensional continuous space that allows to observe similarities among instances compared to a hard-classification where only an unary label is associated to an instance.

\begin{figure}[tb]
	\centering %
	\subfigure[OSCD dataset]{\label{fig:eval:interCatDistOSCD}\includegraphics[height=0.34\linewidth]{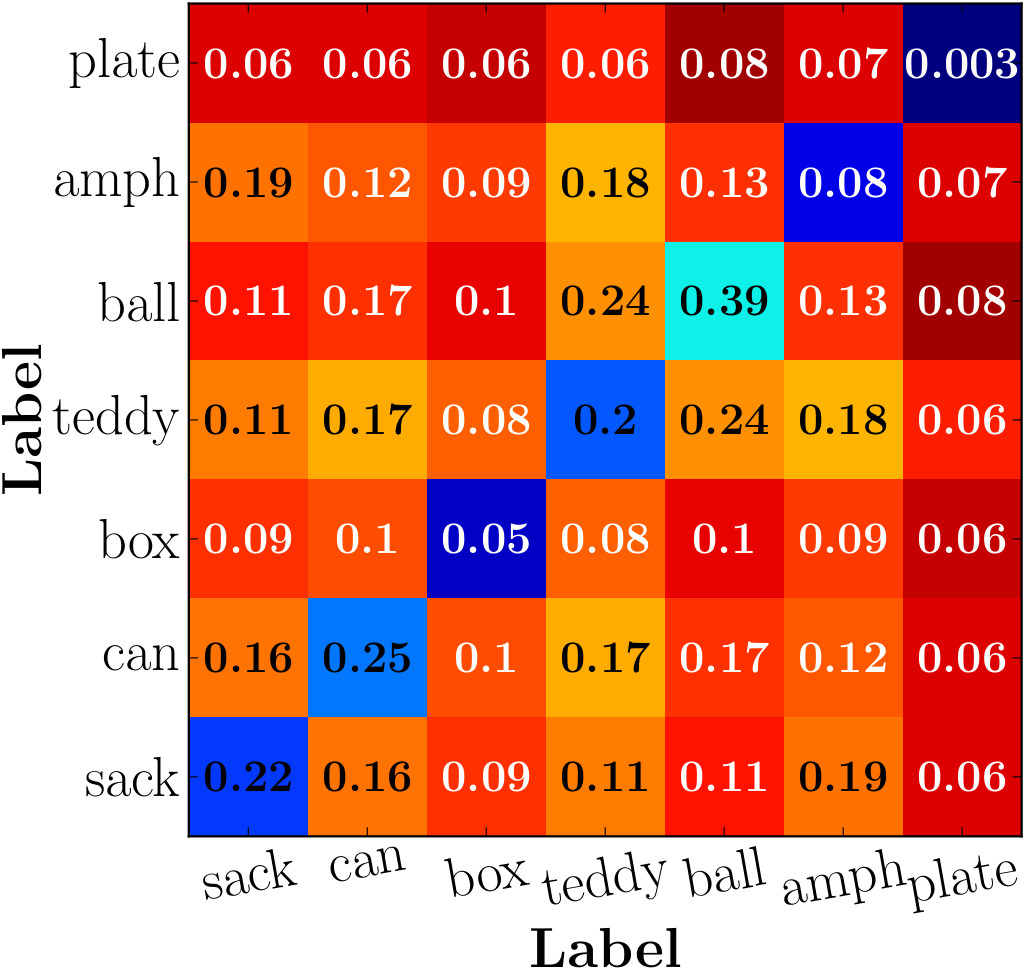}}
	\subfigure[WD dataset]{\label{fig:eval:interCatDistWD}\includegraphics[height=0.31\linewidth]{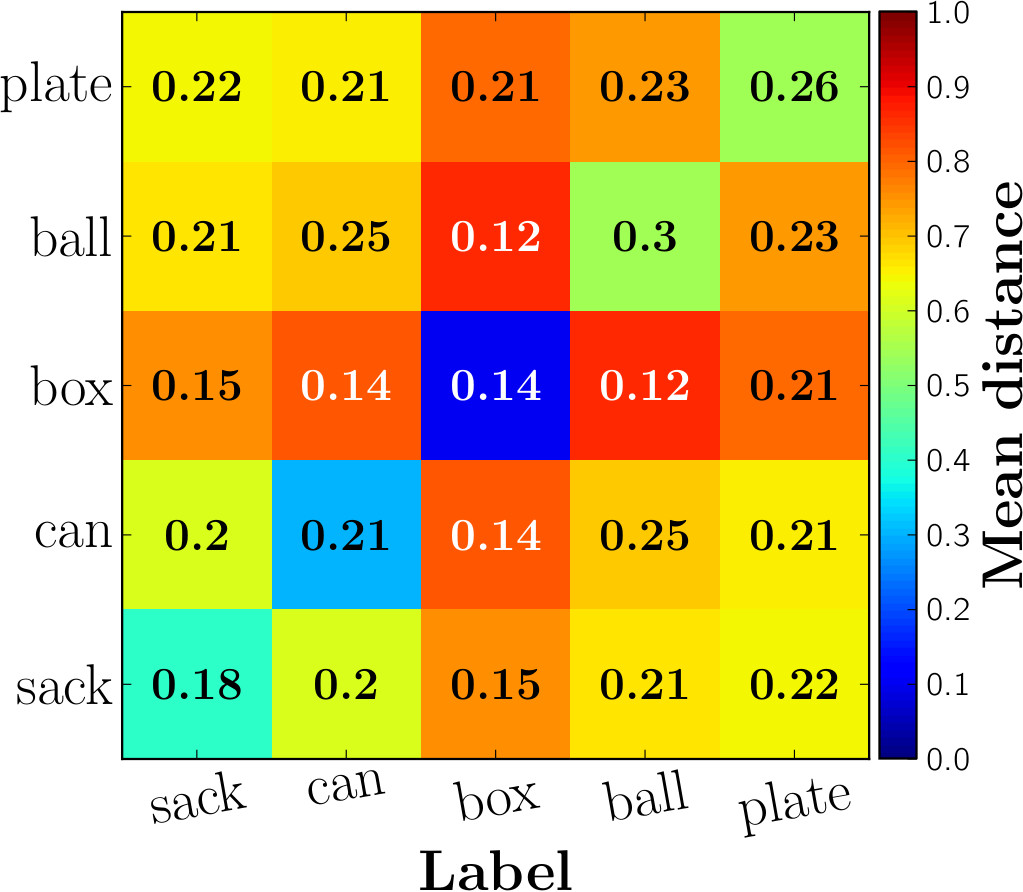}}	
	\subfigure[SD dataset]{\label{fig:eval:interCatDistSD}\includegraphics[height=0.32\linewidth]{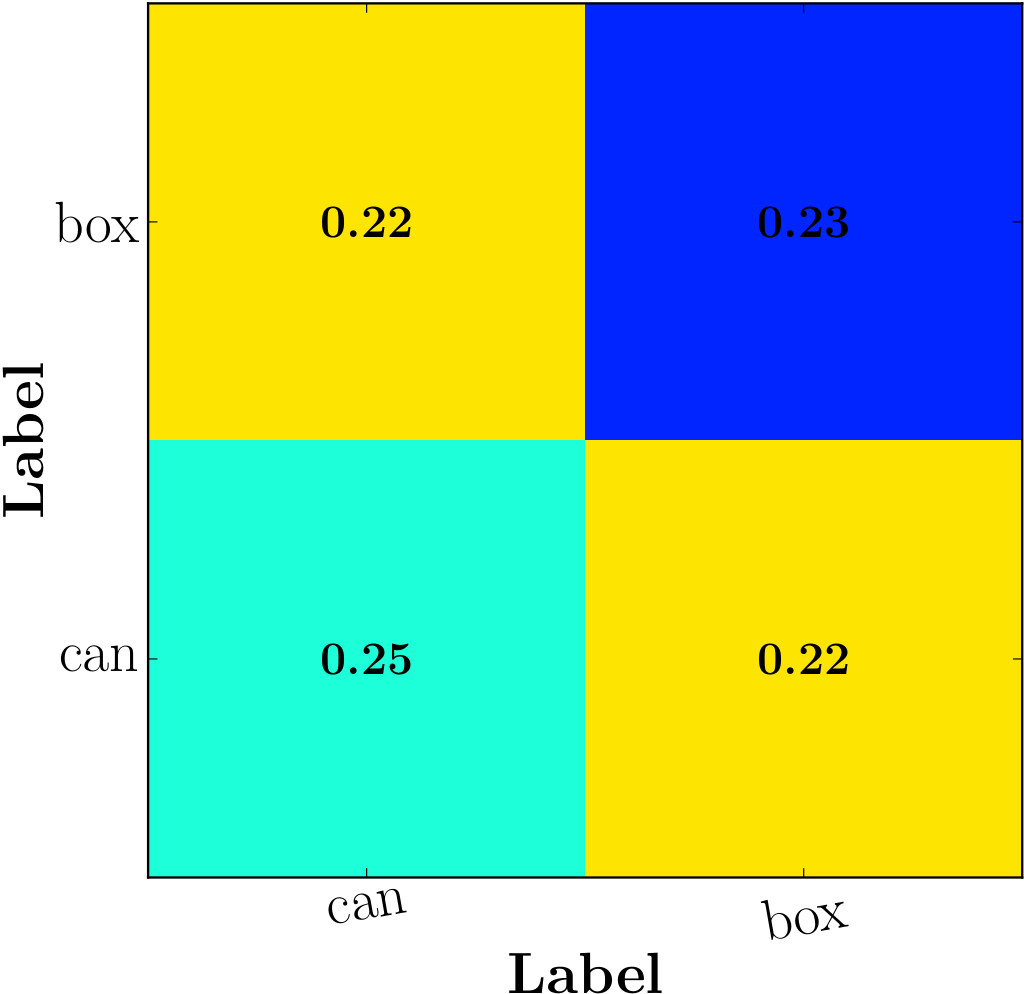}}
	\subfigure[OSCD $\cup$ WD $\cup$ SD datasets combined]{\label{fig:eval:interCatDistCombined}\includegraphics[height=0.34\linewidth]{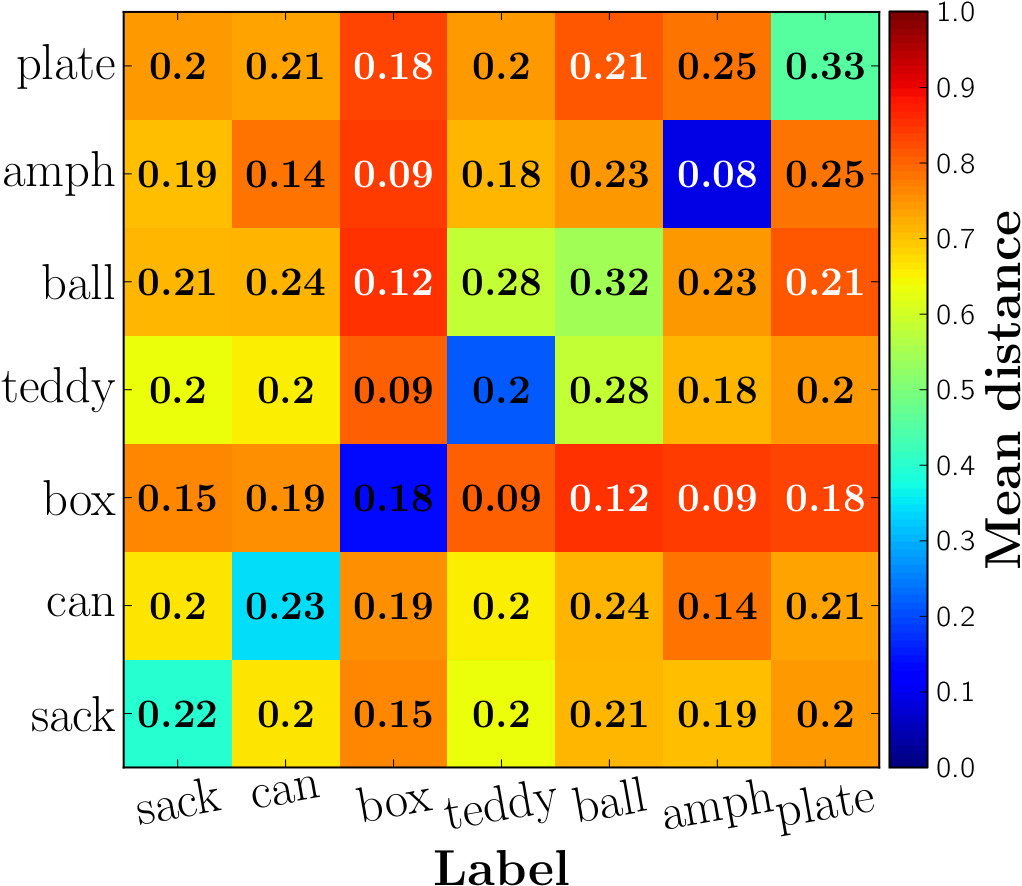}}
	\caption{Inter-category distances based on $\mathcal{HE}$ stimuli. $\mathcal{HE}$ is trained once with OSCD training set. For each category the mean Manhattan distances are computed among all labeled instances. Distances among instances of the respective dataset(s) are scaled ($[0,1]$). Each cell is colored according to the mean distance and annotated with the \emph{standard deviation} of the distances.}
	\label{fig:eval:interCatDist}
\end{figure}

In Fig.~\ref{fig:eval:interCatDist} the mean distances (cell color) and standard deviation (annotation within cell) are shown among the stimuli of dataset instances of the respective category.
It provides a general indication of the coherency among stimuli of different datasets.
Generally categories are distinctive: obviously categories of the OSCD testing set are most discriminative in comparison to WD and SD. 
Nevertheless, instances or parts of these may appear similar, for instance in Fig.~\ref{fig:eval:interCatDistCombined} it is observed that \emph{balls} and \emph{teddies} share similarity: this can be explained that the head or torso of a teddy (OSCD) and a ball including a (prolate spheroidal-shaped) football (WD) may appear similar.
Note that, complex-shaped instances (e.g. \emph{teddy}) may represent compositions of primitive-shaped instances (e.g. \emph{cylinder} or \emph{sphere}). 

\begin{figure}[t]
	\centering
	\includegraphics[width=0.6\linewidth]{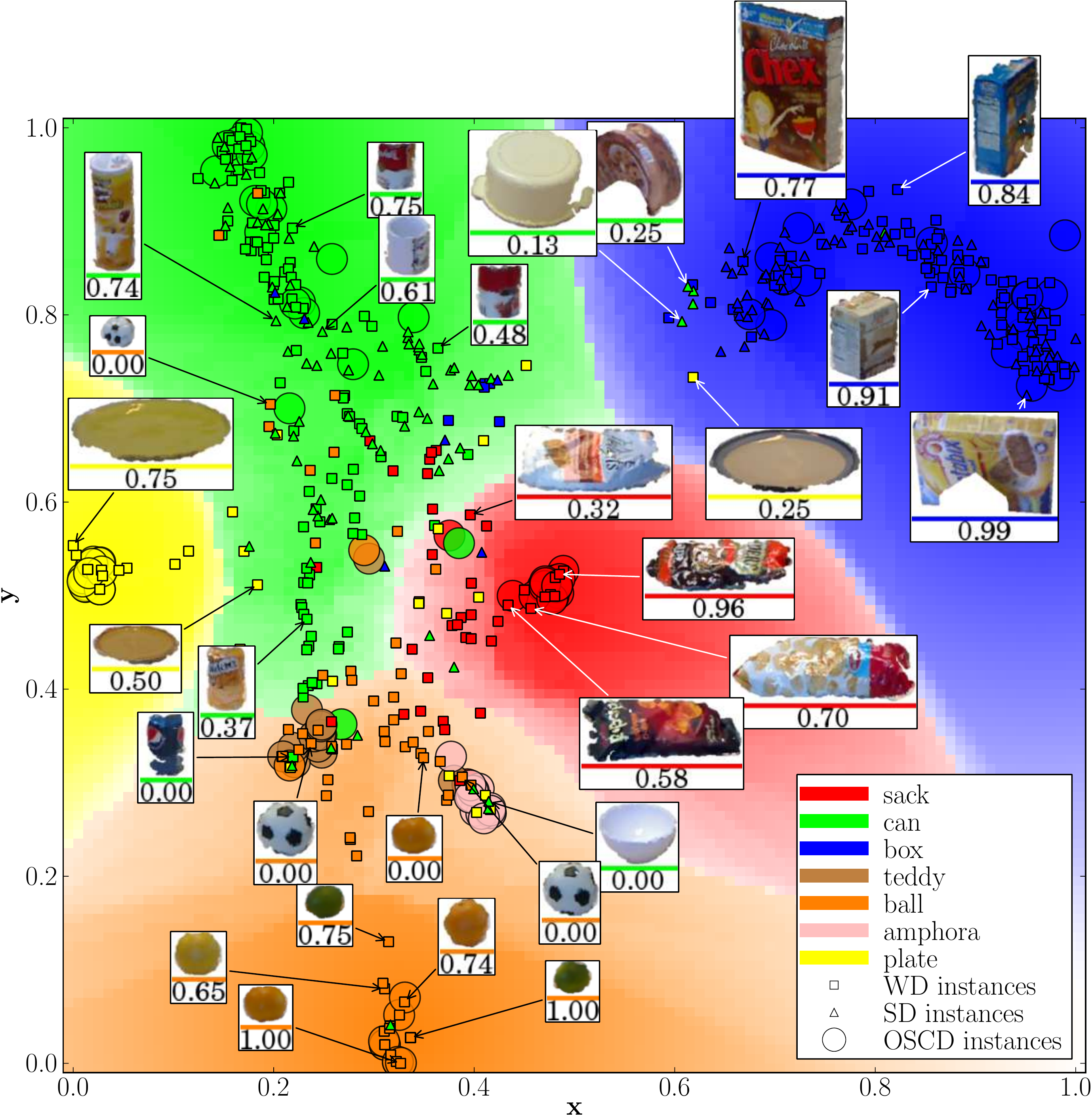}
	\caption{A two-dimensional embedded space $\mathcal{SR}$ showing instances from the OSCD testing set, \emph{Washington RGB-D Object Dataset} (WD) and \emph{Object Segmentation Database}~(SD). In total 603 instance scans are extracted (WD: 335, SD: 154, OSCD testing set: 114); further details about the compilation of instances from WD can be found in Fig.~\ref{fig:lai_tsne} and from SD in Fig.~\ref{fig:osd_tsne}, respectively. Exemplary instances are annotated with their respective scan and category response.}
	\label{fig:oscd_lai_osd_tsne}
\end{figure}

\begin{figure}[t]
	\centering 
	\begin{minipage}[b]{0.6\linewidth}
		\centering				
		\includegraphics[width=1.0\linewidth]{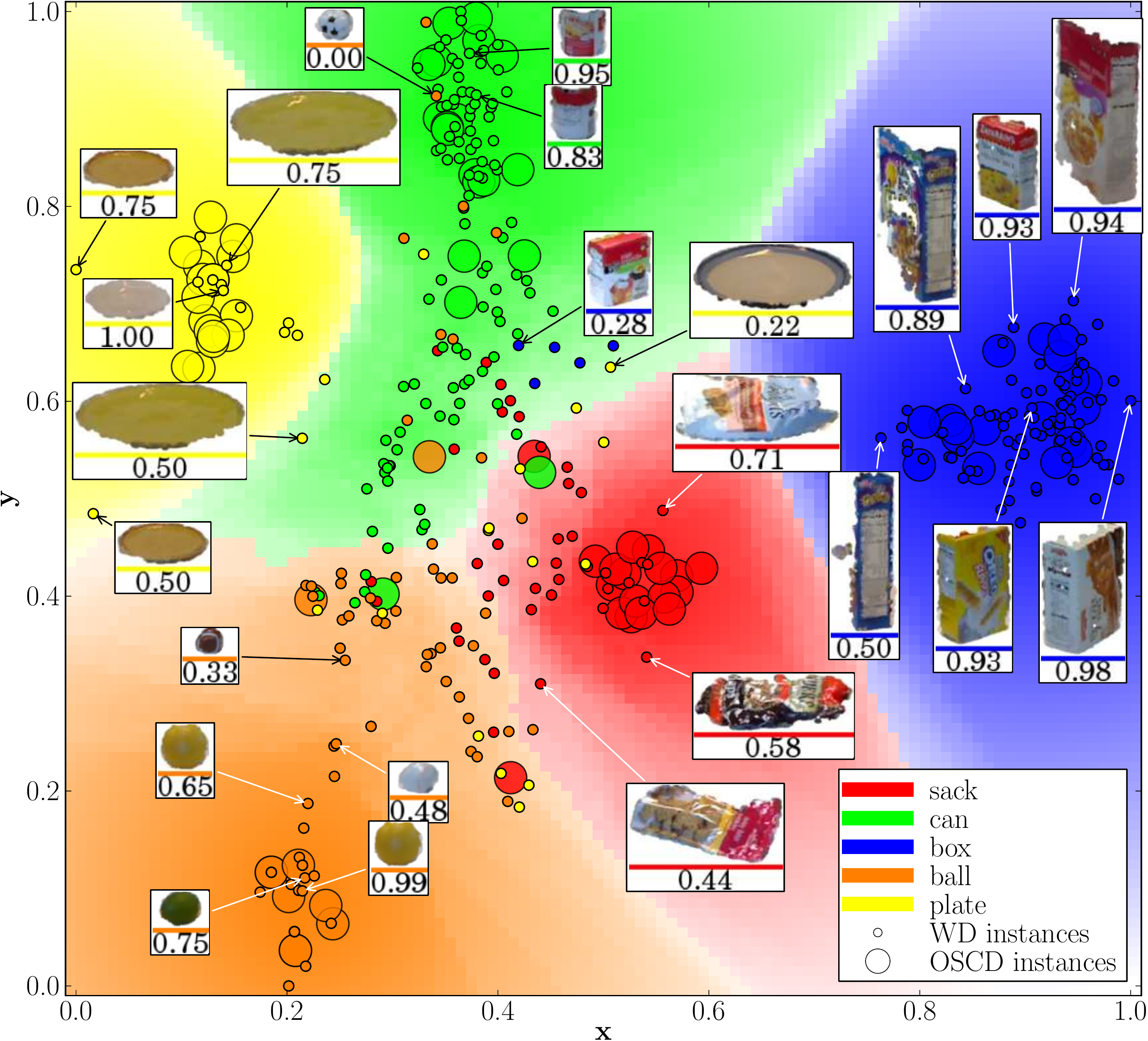}
	\end{minipage}	
	\begin{minipage}[b]{0.39\linewidth}
		\centering
		\scriptsize
		\setlength{\tabcolsep}{2pt}
		\renewcommand{\arraystretch}{1.0}
		\begin{tabular}[b]{ l|| l |l|l |l ||l }
			\multicolumn{6}{c}{} \\ %
			&\multicolumn{2}{c|}{WD instance} &\multicolumn{2}{c||}{OSCD testing}& \\
			&\multicolumn{2}{c|}{associations }&\multicolumn{2}{c||}{set instances}& \ $\Sigma$ \\
			Cat. & Scans &\ \#& Scans& \ \#\\ \hline
			\hline
			\emph{sack} &  \emph{food bag} 1-8 & 40&\emph{sack} 0-17&18&58\\ \hline
			\emph{can}  &  \emph{food can} 1-14 &70&\emph{can} 0-18&19&\multirow{2}{*}{119}\\
			&  \emph{soda can} 1-6 & 30&&&\\ \hline
			\emph{box}  & \emph{cereal box} 1-5 & 25&\emph{box} 0-18&19&\multirow{2}{*}{104}\\
			& \emph{food box} 1-12 & 60&&&\\ \hline
			\emph{ball} &  \emph{ball} 1-7 & 35&\emph{ball} 0-9&10&\multirow{3}{*}{85}\\
			&  \emph{lime} 1-4 & 20&&&\\
			&  \emph{orange} 1-4 & 20&&&\\ \hline
			\emph{plate}& \emph{plate} 1-7 &35&\emph{plate} 0-19&20&55 \\ \hline \hline
			$\Sigma$& - & 335 & - & 86 & 421
		\end{tabular} 
		\includegraphics[height=0.8\linewidth]{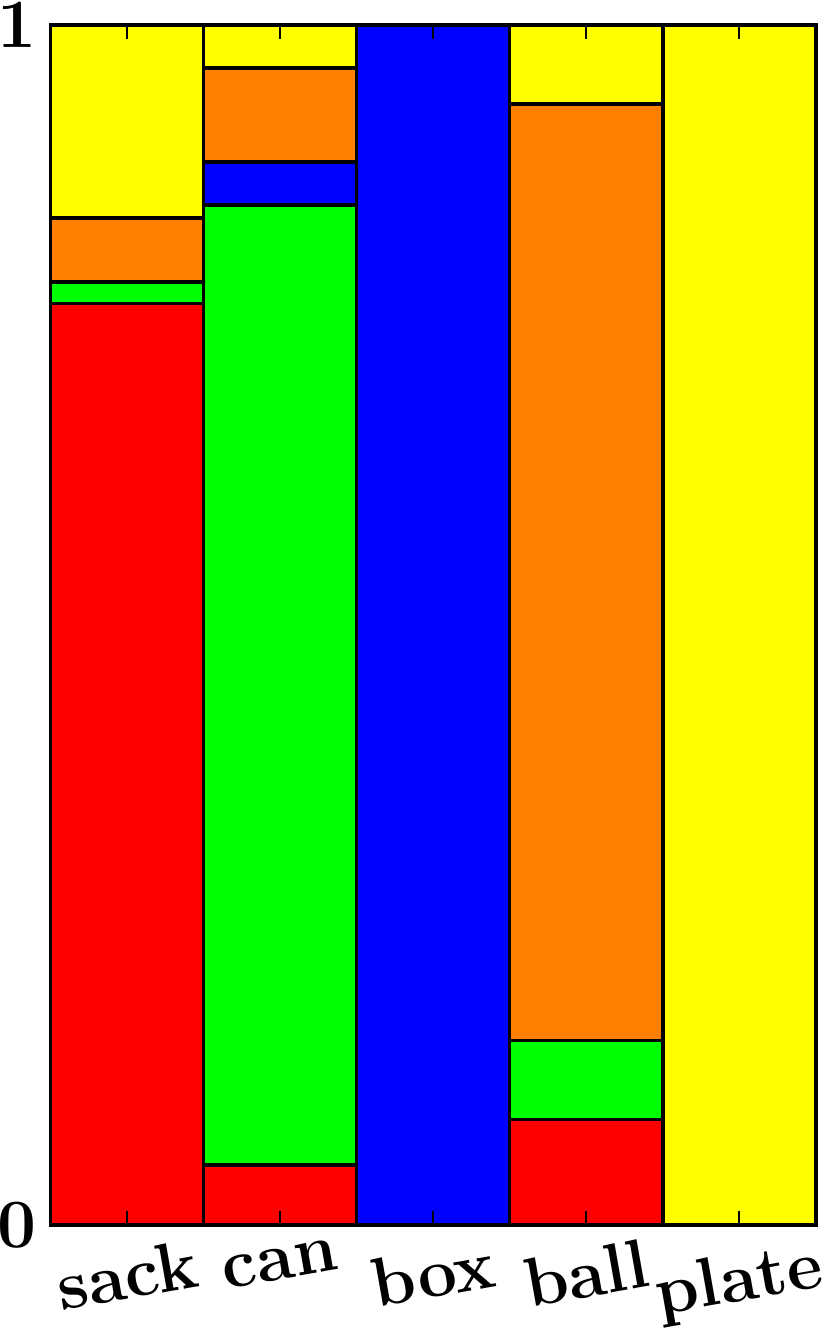}
	\end{minipage}
	\caption{A two-dimensional embedded space $\mathcal{SR}$ showing instances (large circles) from the OSCD testing set and instances (small circles) from the \emph{Washington RGB-D Object Dataset}~\cite{5980382} (WD). Remarks: teddies and amphoras are only excluded in this visualization due to the non-availability of corresponding instances in WD; for each instance the $1^{st}$ to $5^{th}$ point cloud scans are selected of the first video sequence; only for visualization purposes instances are colored according to the best-fit OSCD-category. In total 335 instance scans are extracted from WD. Exemplary instances are annotated with their respective scan and category response. The bar chart presents the sample distribution according to their supervisedly given label within each region.}
	\label{fig:lai_tsne}
\end{figure}

\begin{figure}[t]
	\centering 
	\begin{minipage}[b]{0.6\linewidth}
		\centering	\includegraphics[width=1.0\linewidth]{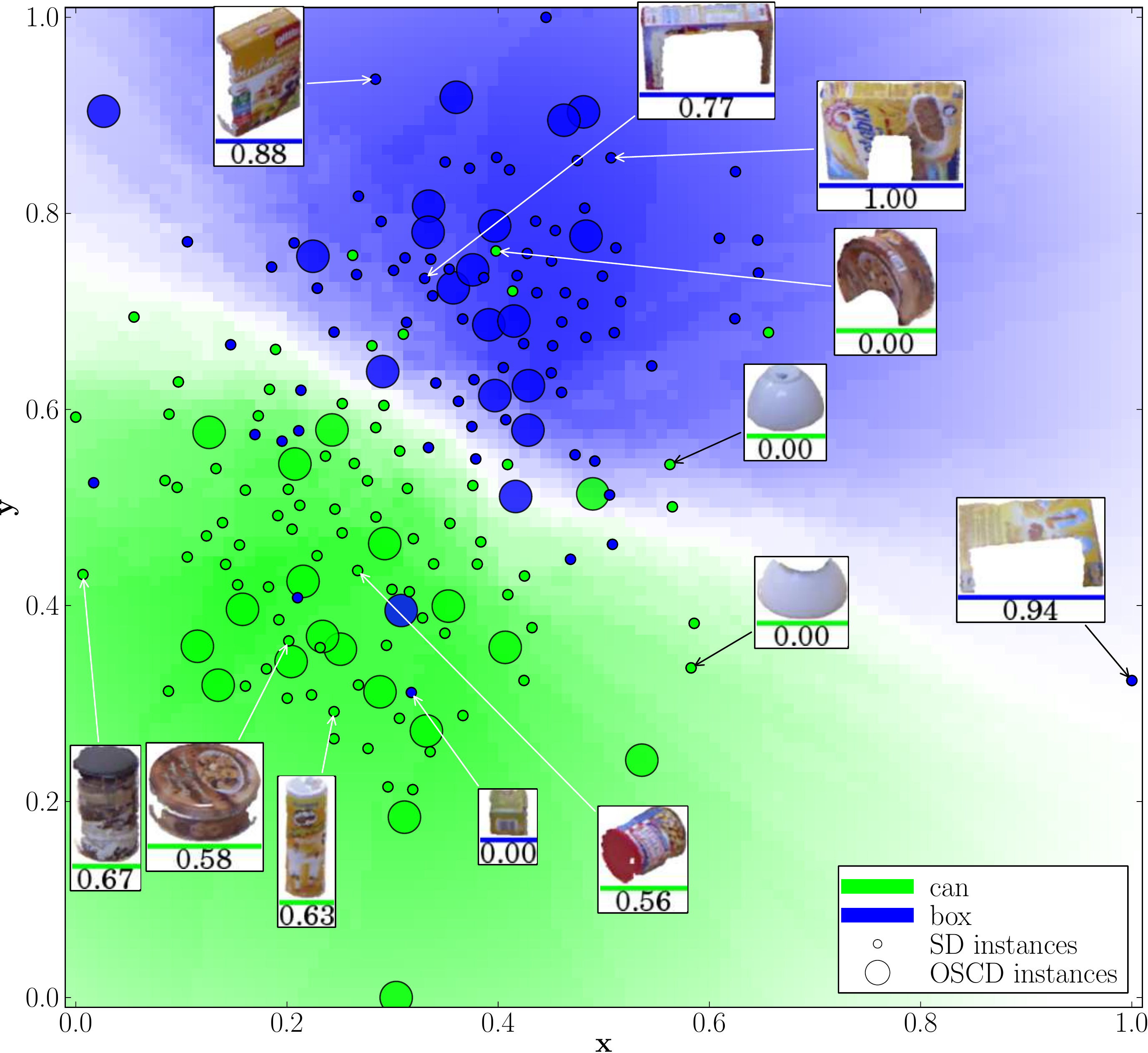}
	\end{minipage}	
	\begin{minipage}[b]{0.39\linewidth}
		\centering
		\scriptsize
			\setlength{\tabcolsep}{2pt}
			\renewcommand{\arraystretch}{1.0}
			\begin{tabular}[b]{ l|| l |l|l |l ||l }
				\multicolumn{6}{c}{} \\ %
				&\multicolumn{2}{c|}{SD instance} &\multicolumn{2}{c||}{OSCD testing}& \\
				&\multicolumn{2}{c|}{associations }&\multicolumn{2}{c||}{set instances}& \ $\Sigma$ \\
				Cat. & Scans &\ \#&Scans&\ \#\\ \hline
				\hline
				\emph{can}  &  \emph{learn} 33-44 &38&\emph{can} 0-18&19&\multirow{2}{*}{99}\\
				&  \emph{test} 31-42 & 42&&&\\  \hline
				\emph{box}  & \emph{learn} 0-16 & 38&\emph{box} 0-18&19&\multirow{2}{*}{93}\\
				& \emph{test} 0-15 & 36&&&\\  \hline \hline
				$\Sigma$& - & 154 & - & 38 & 192
			\end{tabular}  
		\includegraphics[height=0.8\linewidth]{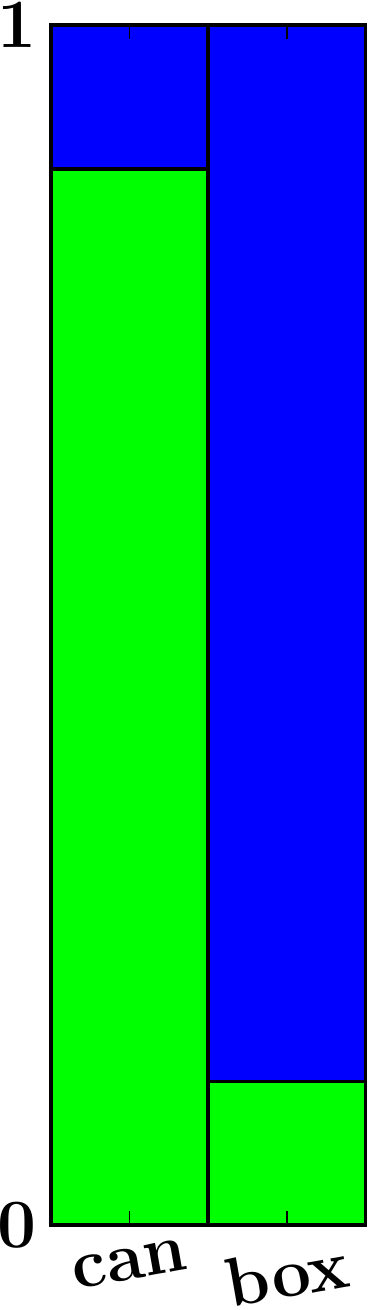}
	\end{minipage}
	\caption{A two-dimensional embedded space $\mathcal{SR}$ showing instances (large circles) from the OSCD testing set and instances (small circles) from the \emph{Object Segmentation Database}~\cite{6385661}~(SD). Remarks: for visualization purposes instances are colored according to the best-fit OSCD-category . In total 154 instance scans are extracted from scenes of SD tagged as ``Boxes'' and ``Cylindric Objects''. Exemplary instances are annotated with their respective scan and category response. The bar chart presents the sample distribution according to their supervisedly given label within each region.}
	\label{fig:osd_tsne}
\end{figure}

In order to visualize the $|\mathcal{Y}|$-dimensional space of stimuli we make use of the t-SNE~\cite{ictdbid:2777} embedding method to reduce the dimensionality to two. 
We denote this two-dimensional space as stimuli-response space $\mathcal{SR}$.
The embedding is performed in an unsupervised manner, i.e. label-agnostic.
Given $\mathcal{SR}$, instances are projected in this space from the WD, SD and the testing set of our OSCD dataset.
Fig.~\ref{fig:oscd_lai_osd_tsne} shows the space combining instances from OSCD, WD and SD dataset, whereas for the sake of readability and illustration purposes, Fig.~\ref{fig:lai_tsne} and Fig.~\ref{fig:osd_tsne} focus on the respective datasets.

Without considering the actual unary and supervisedly given instance label when creating $\mathcal{SR}$, $\mathcal{SR}$ allows to objectively analyze similarity among instances and also the characteristics of locations and regions in $\mathcal{SR}$.
Each projected instance in $\mathcal{SR}$ can be interpreted in this context as a \emph{prototype} which contributes to span the $\mathcal{SR}$ space.
For visual illustration, regions in $\mathcal{SR}$ are colored according to their dedication to a certain label by exploiting the projected instances as anchor points in space.
Therefore, an uniform fine grid is created within the 2D $\mathcal{SR}$ space and for each cell in the grid the $k$-nearest-instances are determined (e.g. $k\mathrm{=}$5\% of total number of instances used in the respective experiment); the label of the majority of the $k$ determined instances leads to the label of the cell in the grid; each cell is weighted ($[0,1]$) and visually depicted in form of cell opacity. The weight represents the observed proportion of instances associated to the majority-label compared to the other labels; the opacity is depicted in an interval of low to high proportion $[$transparent (white)$\mathrm{=}0$, opaque (solid majority label color)$\mathrm{=}1$$]$.

In the context of \emph{Cognitive Science}, specifically in the field of representation architectures, $\mathcal{SR}$ can be interpreted as a \emph{Conceptual Space}~\cite{2000:CSG:518647,zenker2015} where points (prototypes) in space represent multidimensional vectors of \emph{stimuli} and regions in space \emph{concepts}.
These stimuli are often denoted as \emph{Quality Dimensions} and can be interpreted as the stimuli responses of $\mathcal{HE}$. 
Further on another attribute can be observed that $\mathcal{HE}$ stimuli of similar instances appear close in $\mathcal{SR}$ in comparison to dissimilar ones: majority of instances of the respective label are closest (see Fig.~\ref{fig:eval:interCatDist}) or within the same region and form groups, see Fig.~\ref{fig:oscd_lai_osd_tsne}.

\begin{figure}[t]
	\centering
	\subfigure[]{\label{fig:eval:lai_cfmat:dist}\includegraphics[width=0.39\linewidth]{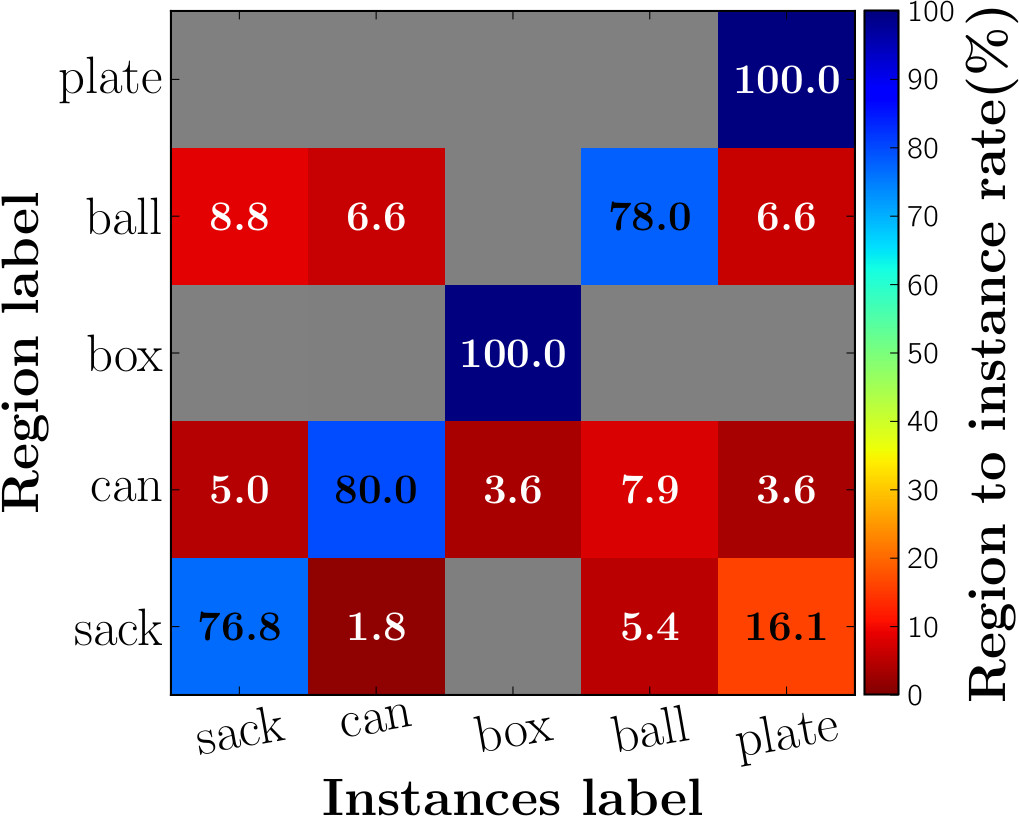}}
	\subfigure[]{\label{fig:eval:lai_cfmat:assigm}\includegraphics[width=0.38\linewidth]{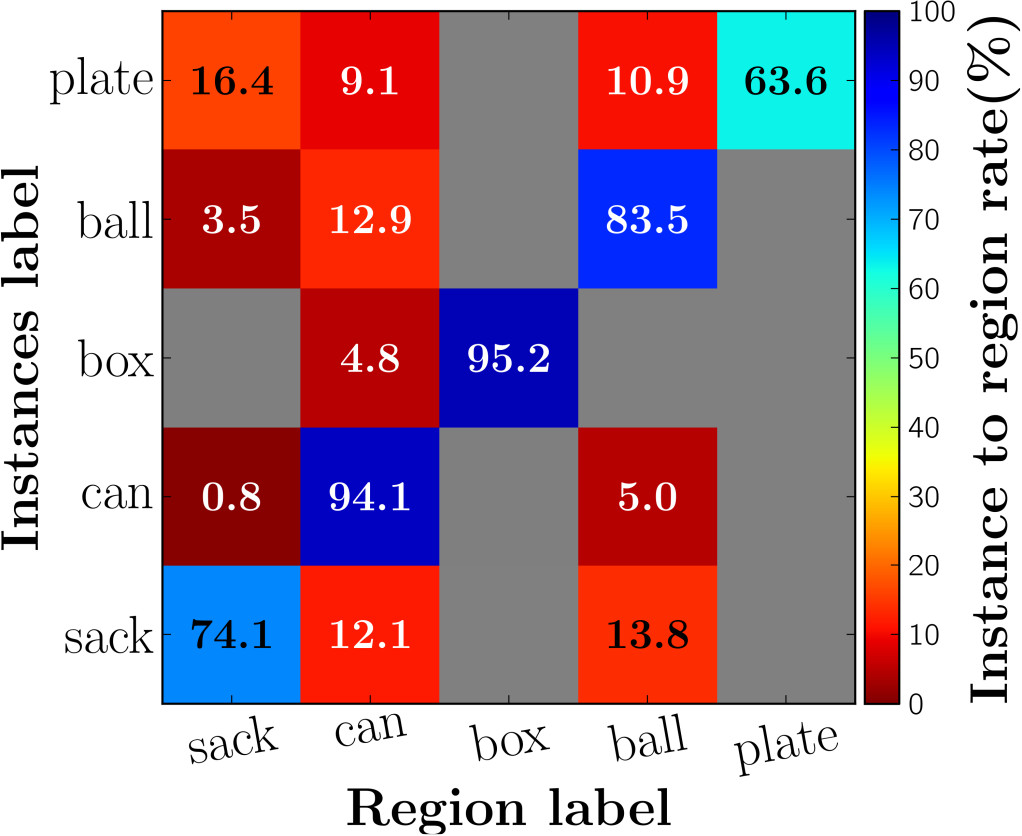}}	
	\caption{Confusion matrices regarding WD dataset showing the distribution of instances within a region~\usubref{fig:eval:lai_cfmat:dist} and the assignment of instances to particular regions~\usubref{fig:eval:lai_cfmat:assigm} (gray marked cell = no result is evident).} 
	\label{fig:eval:lai_cfmat}
\end{figure}

\begin{figure}[t]
	\centering %
	\subfigure[]{\label{fig:eval:osd_cfmat:dist}\includegraphics[width=0.39\linewidth]{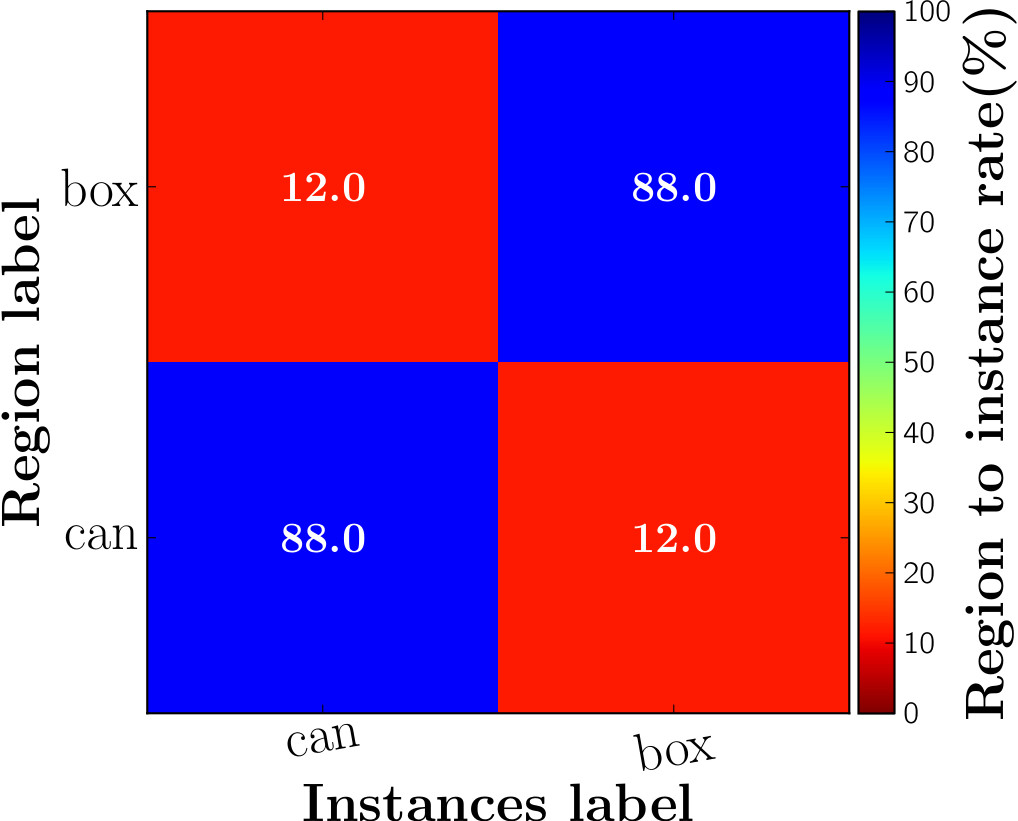}}
	\subfigure[]{\label{fig:eval:osd_cfmat:assigm}\includegraphics[width=0.38\linewidth]{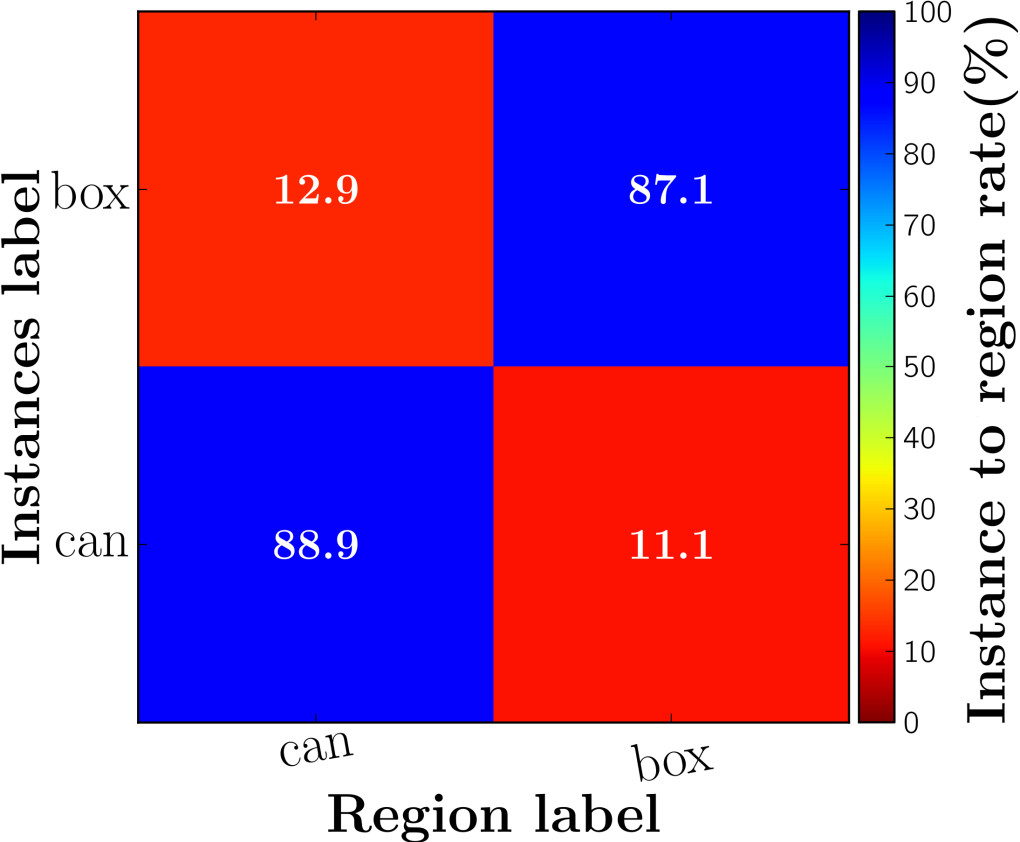}}
	\caption{Confusion matrices regarding SD dataset showing the distribution of instances within a region~\protect\usubref{fig:eval:osd_cfmat:dist} and the assignment of instances to particular regions~\protect\usubref{fig:eval:osd_cfmat:assigm}.} 
	\label{fig:eval:osd_cfmat}
\end{figure}

In contrary to a conventional evaluation analyzing discrete and unary classifications to specific labels (\emph{hard-classification}), the continuous space $\mathcal{SR}$ shown in Fig.~\ref{fig:lai_tsne} and Fig.~\ref{fig:osd_tsne} allows to learn about the regional characteristics and relations among locations in $\mathcal{SR}$ and instances of the three datasets.
A main observation is that instances from different datasets are propagated through  $\mathcal{HE}$ and resulting $\mathcal{HE}$ responses show \emph{coherency}. 
It can be shown that instances of all evaluated datasets together form interrelated and coherent groups -- see uniformly colored regions in Fig.~\ref{fig:lai_tsne} and Fig.~\ref{fig:osd_tsne} as well as the variety of instances within the region in form of the bar chart.
In case of the WD dataset, the bar chart in Fig.~\ref{fig:lai_tsne} and the confusion matrix in Fig.~\ref{fig:eval:lai_cfmat:dist} show that $76\%$ of instances within the \emph{sack}-region are \emph{sacks} ($80\%$ in case of \emph{cans}, $100\%$ for \emph{boxes}, $78\%$ for \emph{balls} and $100\%$ for \emph{plate}). 
In Fig.~\ref{fig:eval:lai_cfmat:assigm} the distribution of instances according to their supervisedly given labels are shown: \emph{ball} instances are found at regions of  \emph{cans} and \emph{sacks} which can be explained that \emph{balls} share similar properties like \emph{cans} or \emph{sacks} in 2.5D -- all three shape types may contain bulging and roundish surfaces.
A similar observation is made for the SD dataset in Fig.~\ref{fig:osd_tsne} and  Fig.~\ref{fig:eval:osd_cfmat}.

OSCD instances and instances from other datasets are interrelated and form groups in $\mathcal{SR}$ but they show deviations in their positions: an extreme case can be found for \emph{ball}, see Fig.~\ref{fig:lai_tsne}: OSCD instances are predominately accumulated at region ($x \mathrm{\approx} 0.2|y \mathrm{\approx} 0.15$) whereas WD instances are partially scattered in vicinity of the accumulation.
This can be explained that OSCD instances do not necessarily cover the entire region of the respective shape given the $\mathcal{HE}$ response -- note, $\mathcal{HE}$ is trained in a data-driven manner with the training set that represents a subset of possible shape appearances of a particular type or label in space.
Thus it provides a specific perspective on shape appearances in space.
Therefore instances of other datasets may cover or extend related regions dedicated to the respective shape. %

\begin{figure}[!tb]
	\centering
	\subfigure[Cylindrical instance scans (total: 80) from \emph{learn} 33-44 and \emph{test} 31-42 scenes of SD dataset with $\mathcal{HE}$ responses (bar chart) and below scalar \emph{can} response.]{\label{fig:eval:osd_can_gallery}
		\includegraphics[width=0.47\linewidth]{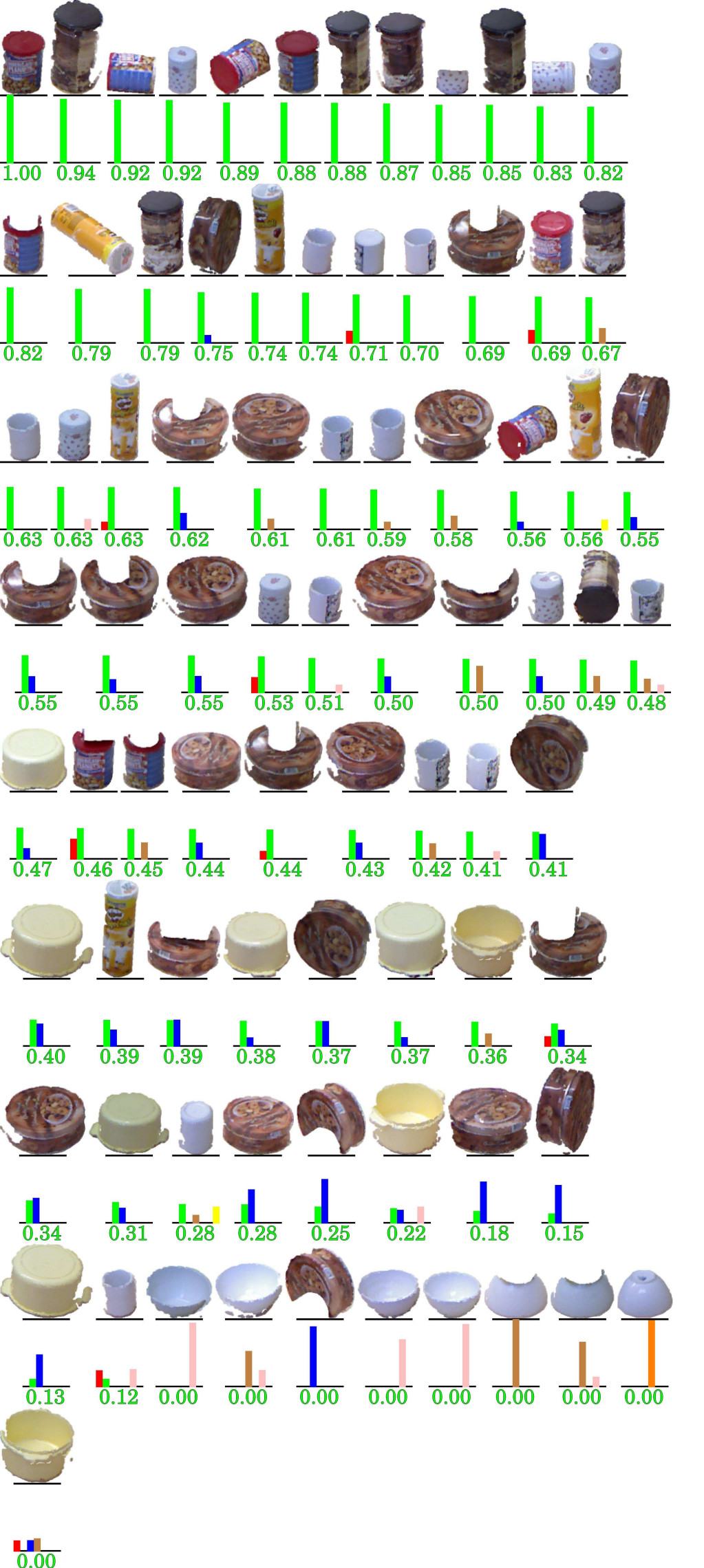}}
	\hspace{0.1cm}
	\subfigure[Box scan instances (total: 74) from \emph{learn} 0-16 and \emph{test} 0-15 scenes of SD dataset with $\mathcal{HE}$ responses (bar chart) and below scalar \emph{box} response.]{\label{fig:eval:osd_box_gallery}\includegraphics[width=0.47\linewidth]{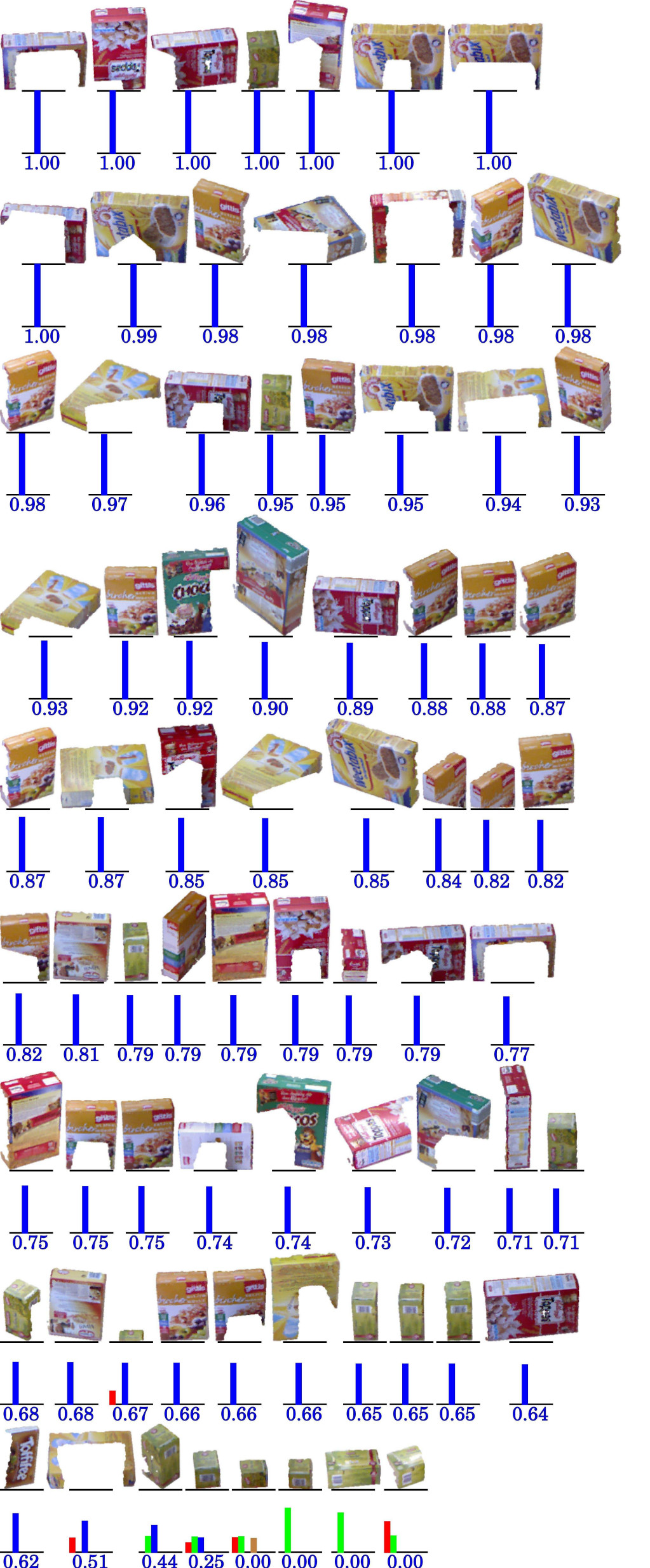}}
	\caption{In descending order, ranked instances from the SD dataset regarding the $\mathcal{HE}$ response of the respective labels \emph{can} and \emph{box}. These instances are extracted from scenes of the SD dataset. Therefore instances can appear partial due to occlusions.} 
	\label{fig:eval:osd_gallery}
\end{figure}

Boundaries among (labeled) regions are continuous as shapes undergo deformations in space and a supervisedly given label may change at border regions; keeping in mind that objects may not be uniquely assignable to the given set of labels due to shape ambiguities, especially at border regions.  
To illustrate the shape variation within $\mathcal{SR}$, sample locations are annotated and depicted in Fig.~\ref{fig:lai_tsne} and Fig.~\ref{fig:osd_tsne} with the respective point cloud and response.
Samples at border regions obviously are not uniquely assignable to a particular label due to the encountered higher shape ambiguity which is reflected by a lower response of the supervisedly given label or by responses of multiple labels as depicted for SD dataset instances in Fig.~\ref{fig:eval:osd_gallery}.

In Fig.~\ref{fig:eval:osd_gallery} instances of the SD dataset are ranked in descending order according to their response of the supervisedly given label.
This ranking reveals the increasing shape degradation, for instance the first 15 higher ranked cylindric objects ($18.75\%$ of all cylindric objects in Fig.~\ref{fig:eval:osd_can_gallery}) are represented with a single \emph{can} response (avg. $0.87$ response). At lower ranked instances, higher ambiguities can be observed: among others \emph{bowls} and \emph{pots} are found, where a bowl can be interpreted as a mixture of cylinder and cone but was labeled as a cylindric object in SD dataset (see in Fig.~\ref{fig:osd_tsne} bowls are located at border regions of \emph{can} group). 
The box ranking in Fig.~\ref{fig:eval:osd_box_gallery} show a similar pattern. 
Noteworthy, boxes experience substantial occlusions but still show a high \emph{box} response, which can be explained that box related features (partial but distinctive box constellation of segments
 like flat surfaces in $\approx90\ensuremath{^\circ}$ alignment) are mostly unaffected by the occlusion; $93.2\%$ of boxes responded with box as strongest stimulus.
At the lower end \emph{small} boxes are observed with \emph{curved edges} which can lead to ambiguity; it is also worthy of mention that sensor noise has a higher impact on smaller objects (signal-to-noise ratio) and may lead to non-reliable responses.

\begin{figure}[!tb]
	\centering
	\subfigure[Embedded $\mathcal{SR}$ space]{\label{fig:eval:complete_tsne_cf}
		\includegraphics[width=0.45\linewidth]{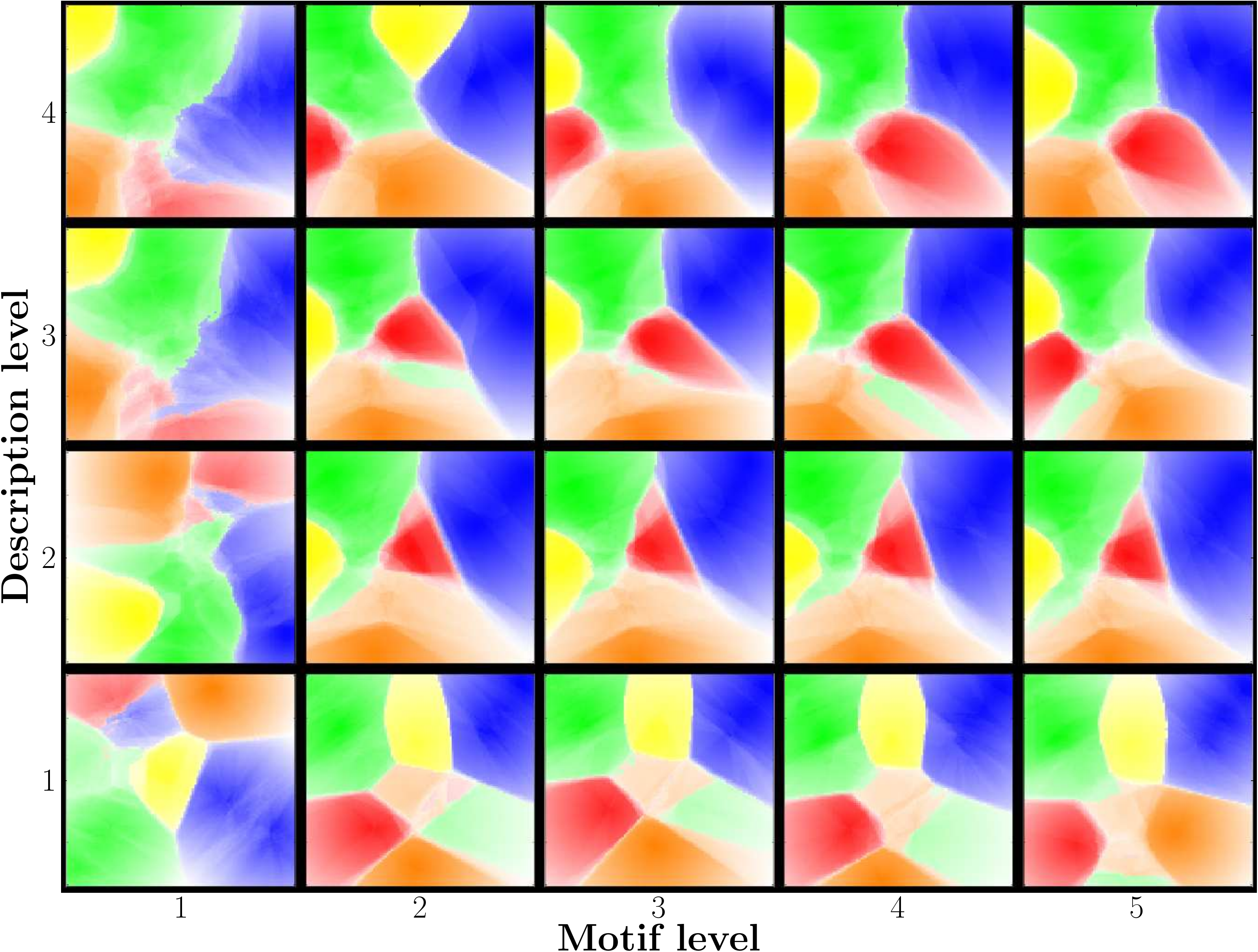}
		\includegraphics[trim=0 -10 0 0,width=0.1\linewidth]{labels_legend}
	}	
	\subfigure[Distribution of instances within a region.]{\label{fig:eval:complete_tsne_cf_reg_in}
		\includegraphics[width=0.46\linewidth]{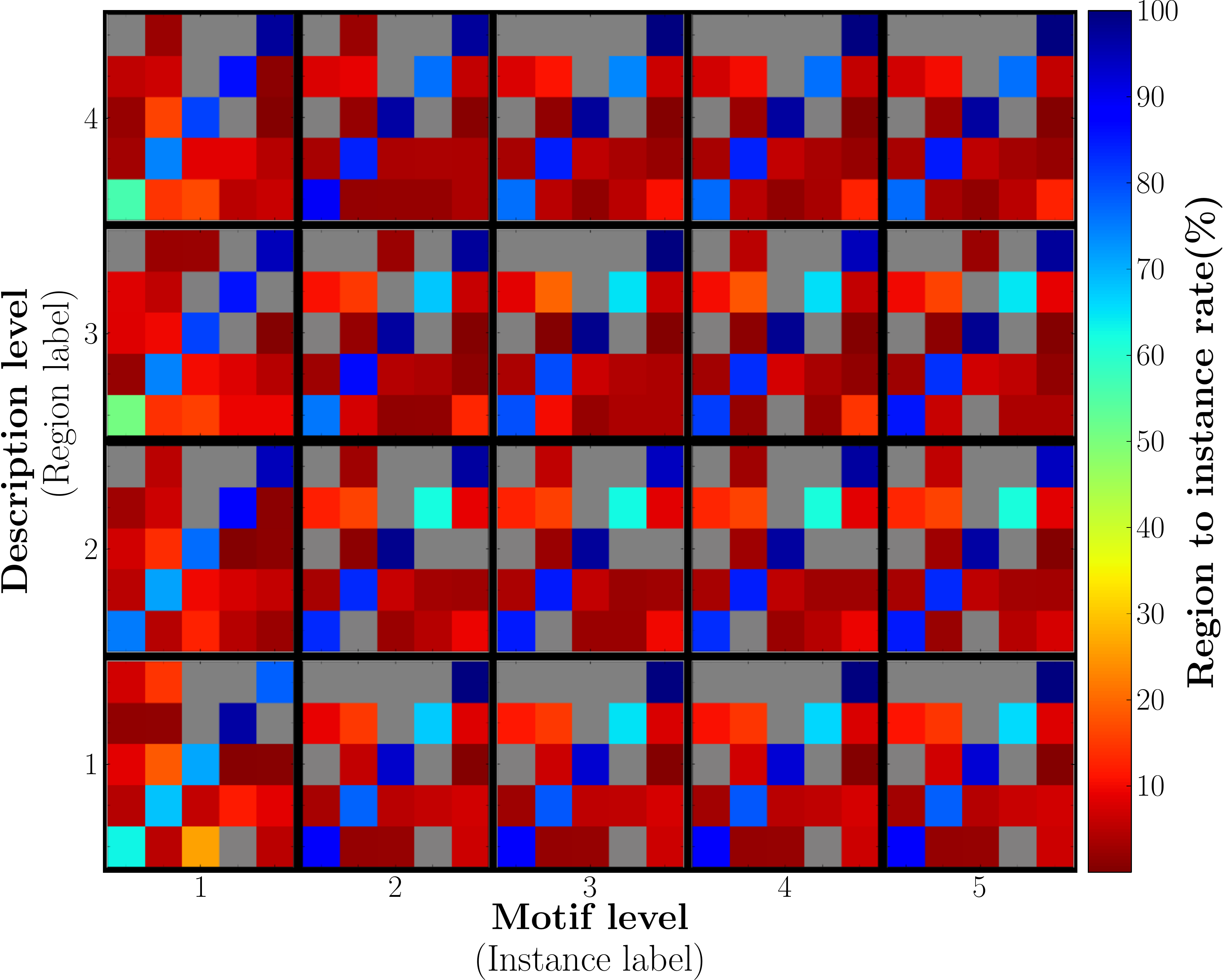}
	}
	\subfigure[Assignment of instances to particular regions.]{\label{fig:eval:complete_tsne_cf_in_reg}
		\includegraphics[width=0.46\linewidth]{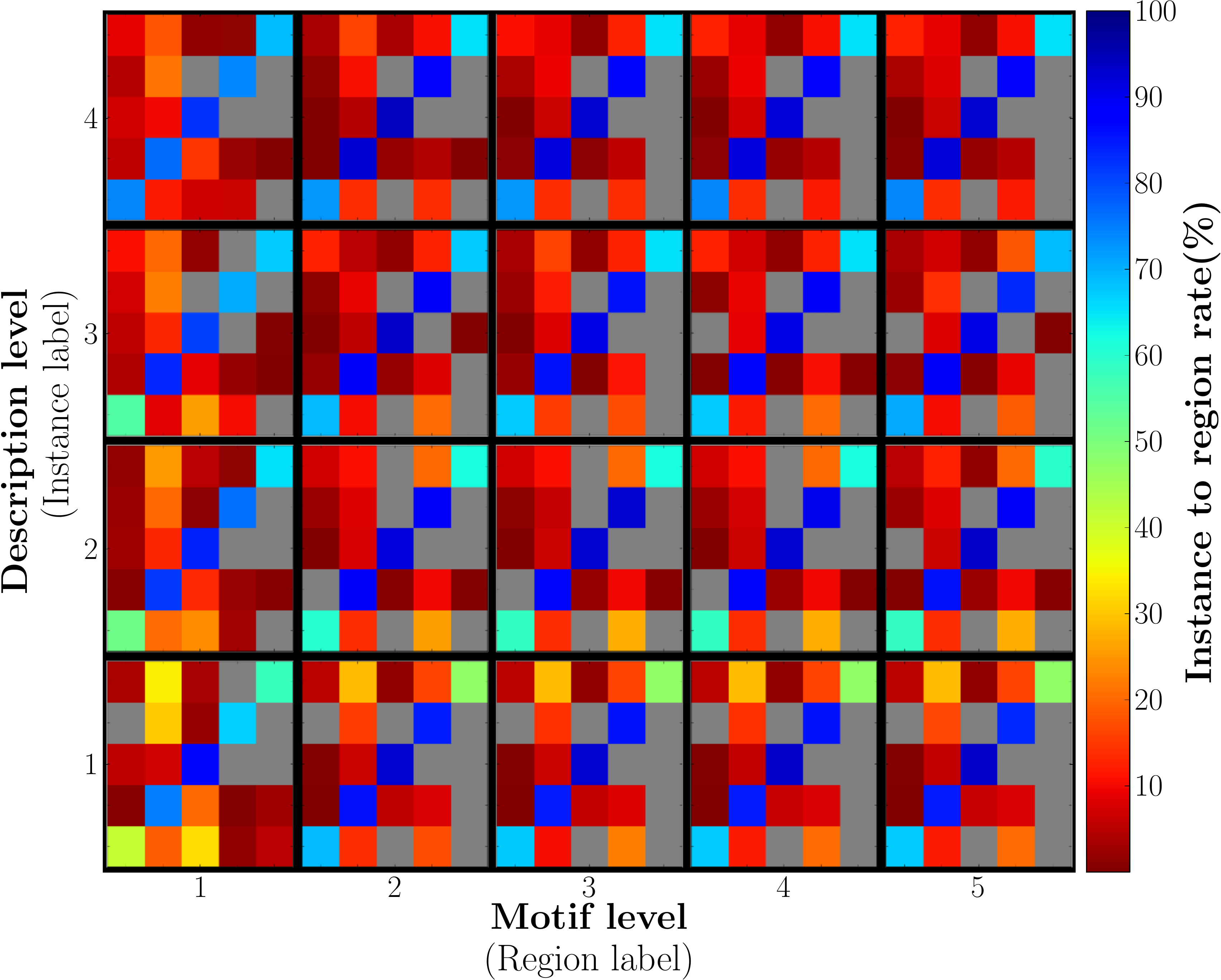}
	}	
	\caption{
		A two-dimensional embedded space $\mathcal{SR}$ \usubref{fig:eval:complete_tsne_cf} generated from the OSCD testing set, \emph{Washington RGB-D Object Dataset} (WD) and \emph{Object Segmentation Database}~(SD) instances for particular description level and motif level $\mathcal{HE}$ configurations. Given $\mathcal{SR}$, the respective distribution of instance within a region \usubref{fig:eval:complete_tsne_cf_reg_in} and the assignment of instances to a region \usubref{fig:eval:complete_tsne_cf_in_reg} are shown -- specifically focusing on alternative dataset (WD, SD) labels. Each cell (description vs. motif level) within \usubref{fig:eval:complete_tsne_cf}, \usubref{fig:eval:complete_tsne_cf_reg_in} and \usubref{fig:eval:complete_tsne_cf_in_reg} is analogously generated according to the description and motif level configuration: for \usubref{fig:eval:complete_tsne_cf} as Fig.~\ref{fig:oscd_lai_osd_tsne}, for \usubref{fig:eval:complete_tsne_cf_reg_in} as Fig.~\ref{fig:eval:lai_cfmat:dist} (including order of labels on axes) and for \usubref{fig:eval:complete_tsne_cf_in_reg} as Fig.~\ref{fig:eval:lai_cfmat:assigm} (including order of labels on axes).
		For \usubref{fig:eval:complete_tsne_cf_reg_in} and \usubref{fig:eval:complete_tsne_cf_in_reg}, the axis label within a cell is shown below the main axis label in parentheses.
	} 
	\label{fig:eval:combined_db_tsne}
\end{figure}

Previous experimental results in Fig.~\ref{fig:oscd_lai_osd_tsne}, \ref{fig:lai_tsne} and \ref{fig:osd_tsne} have shown coherent responses in the respectively generated $\mathcal{SR}$ space from different datasets.
Given the combined instances from the OSCD testing set, \emph{Washington RGB-D Object Dataset} (WD) and \emph{Object Segmentation Database}~(SD) in Fig.~\ref{fig:eval:combined_db_tsne} results are shown to illustrate the effect in $\mathcal{SR}$ space regarding the number of \emph{description} and \emph{motif levels} applied in $\mathcal{HE}$.
Confusion matrices are generated for description and motif level configurations.
It is observable that the discrimination which can be interpreted as distinguishable borders among homogeneous regions, increases with the number of description and motif level until a saturation is reached.
This tendency is similarly observed in previous results shown in Fig.~\ref{fig:eval:hch_dict_vs_clique_error}; however Fig.\ref{fig:eval:complete_tsne_cf} provides a further insight into the distribution of the dataset instances.
For instance few description or motif levels lead to a distorted definition of regions which may has been caused by under-fitting whereas higher levels provide more distinctive separations.

Results in Fig.~\ref{fig:eval:complete_tsne_cf_reg_in} and Fig.~\ref{fig:eval:complete_tsne_cf_in_reg} are based on the generated $\mathcal{SR}$ spaces shown in Fig.~\ref{fig:eval:complete_tsne_cf}.
A blue (100\%) diagonal in each cell of a particular description and motif level configuration represents a perfect discrimination -- note, that $\mathcal{SR}$ is unsupervisly (label-agnostic) generated and purely based on object shape appearances.
Following the diagonal from lower levels to higher levels, one can observe an increasing discrimination expressed by polarization of rates to 100\%~(blue) at diagonals and 0\%~(red) at non-diagonals or even no distortion marked as gray cells. 
For instance, at description level 1 and motif level 1 within the \emph{box} region $70.8\%$ \emph{boxes} are present whereas at description 4 and motif level 5, $97.1\%$ \emph{boxes}, $2.4\%$ \emph{cans} and $0.6\%$ \emph{plates} are present (see Fig.~\ref{fig:eval:complete_tsne_cf_reg_in}).
On the other hand, at description level 1 and motif level 1,  $67.1\%$ of \emph{ball} instances are assigned to \emph{ball} region whereas at description 4 and motif level 5, $88.2\%$ are assigned to ball region (see Fig.~\ref{fig:eval:complete_tsne_cf_in_reg}).

\subsection{Evaluation Summary}
Our conducted evaluation can be divided into three parts.
In the first part, Sec.~\ref{sec:exp:dict} to Sec.~\ref{sec:exp:hch}, experiments were conducted to provided insights into process of the propagation of objects through the hierarchical levels of $\mathcal{HE}$.
Thereby intermediate steps were analyzed of the internal $\mathcal{HE}$ components. %

In the second part, Sec.~\ref{sec:comparison}, a comparison of other approaches was conducted. Each approach was \emph{separately} trained and tested with our OSCD dataset. 
In this experiment an unary classification task was performed which allows to compare the final classification capability of the respective approach under same conditions. 
Such experiment performs a classification accuracy comparison under supervision but neglects the analysis of the approach's generalization capability. 

\begin{figure}[!tb]
	\centering  
	\subfigure[Powerball-Husky platform]{\label{fig:robot1Anno}\includegraphics[height=0.28\textwidth]{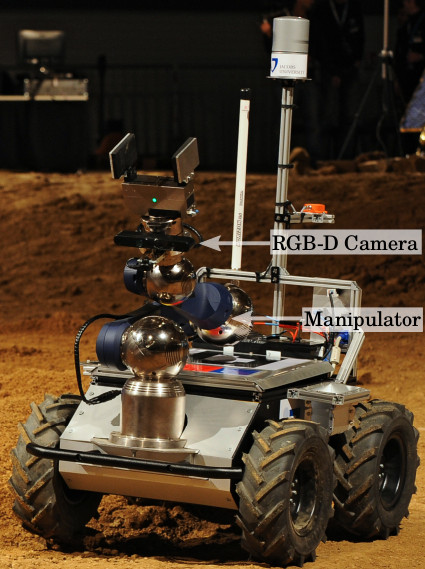}}
	\subfigure[Platform located in front of scene.]{\label{fig:robot1}\includegraphics[height=0.28\textwidth]{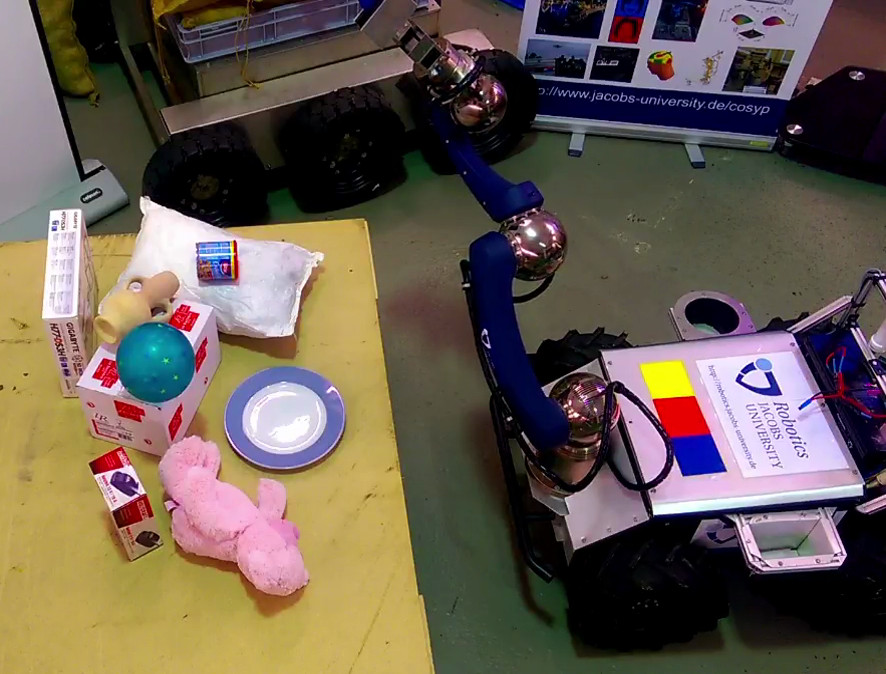}}
	\subfigure[RGB scene observed from platform's perspective.]{\label{fig:robot2}\includegraphics[height=0.28\textwidth]{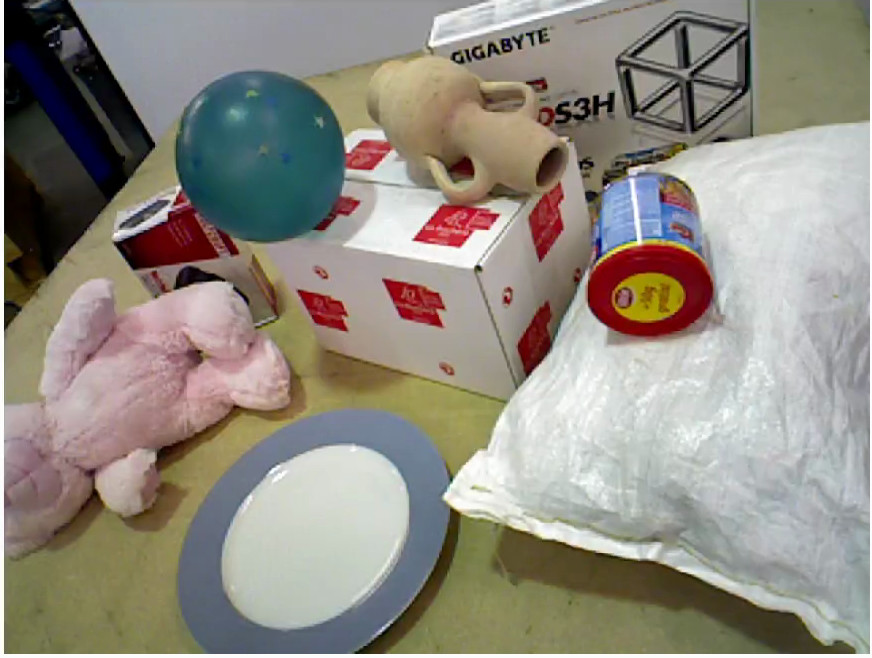}}
	\subfigure[Point cloud scene observed from platform's perspective.]{\label{fig:robot3}\includegraphics[height=0.38\textwidth]{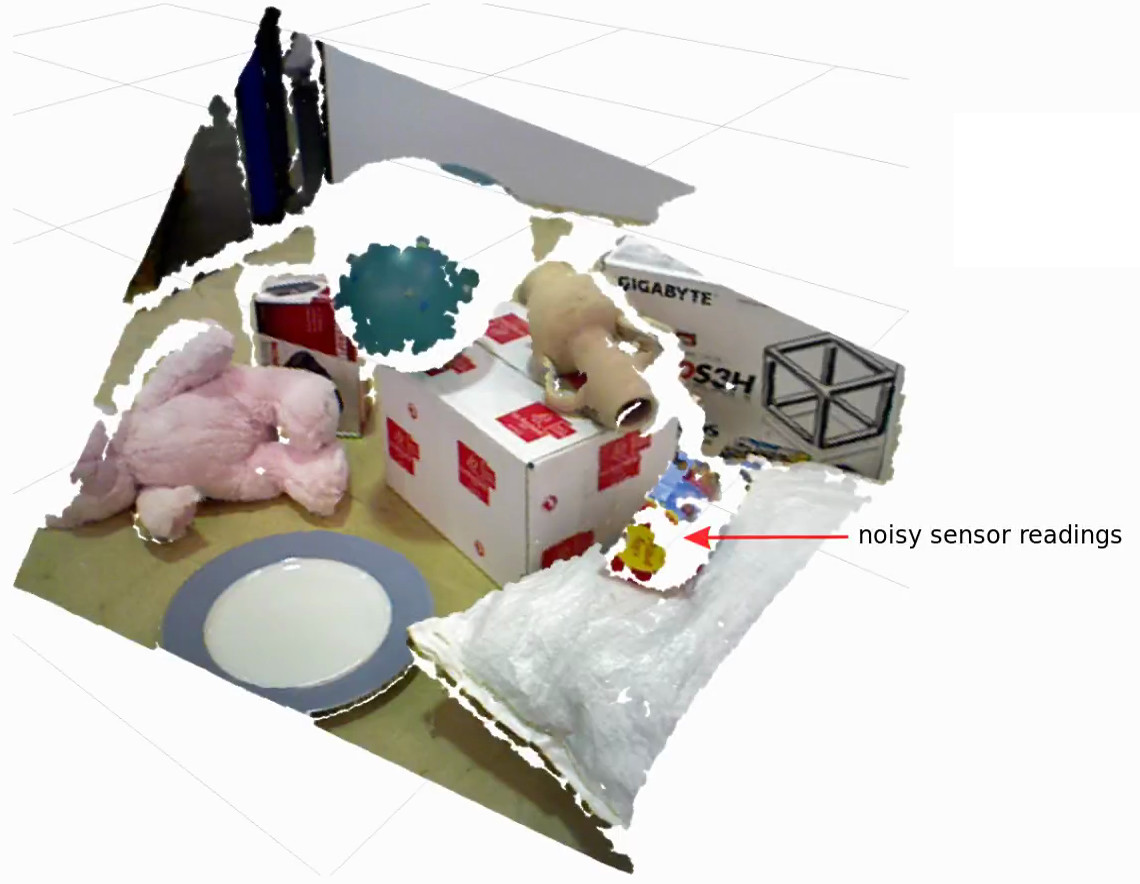}}
	\subfigure[Object candidate segmentation~\cite{MuellerBirkIcra2016} and classification results ($\mathcal{HE}$)]{\label{fig:robot4}\includegraphics[height=0.3\textwidth]{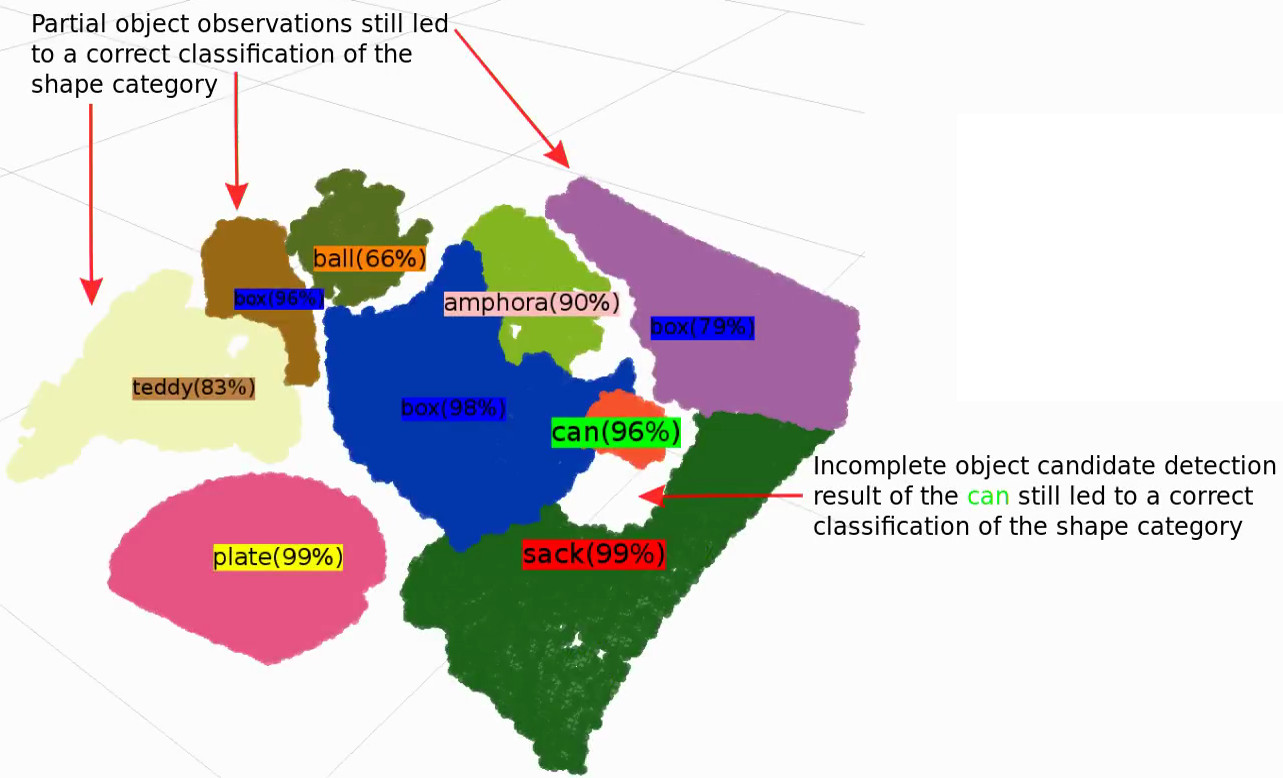}}
	\caption{Illustration of a visual perception task with a Powerball-Husky platform of the Robotics Group (Jacobs University Bremen) in an unstructured scene where instances are noisy and partially perceived due to viewpoint and maneuverability limitations caused by platform's kinematic constraints and setup conditions. Object candidates (randomly colored) are segmented using our previous work~\cite{MuellerBirkIcra2016}. The shape category labels of the classification results of the work presented here ($\mathcal{HE}$) are colored as in Fig.~\ref{fig:vw_dict_distri}.}
	\label{fig:robot_scene}
\end{figure}

In the last evaluation part, Sec.~\ref{sec:alter_db}, other datasets were applied to our approach. 
In this experiment we have not fine-tuned $\mathcal{HE}$ for each dataset and we have not separately conducted experiments for each dataset and retrained $\mathcal{HE}$. %
The proposed approach was confronted with instances from all evaluated datasets (OSCD, WD and SD) in the same experiment and $\mathcal{HE}$ was initially trained once with the training set of our OSCD dataset.
It provides a valuable feedback about the actual generalization capability of the proposed approach which is a crucial property for applications in real world scenarios.
For instance, in the field of robotic logistics \cite{JonschkowskiEHM16,7553531} instance appearances may encounter strong variations (as they were drawn from different distributions) due to unstructured and confined spaces, clutter or even limited maneuverability or kinematic constraints of the robot platform, see Fig.~\ref{fig:robot_scene}.  
Therefore the performed experiment of our approach with instances which are drawn from different distributions in form of different datasets (see Fig.~\ref{fig:eval:instance_variety}) provides insights into the generalization capability compared to single and independent dataset evaluations.

\section{Conclusion}
\label{sec:conclusion}

We presented a part-based object shape categorization approach that focuses on two aspects.

\textbf{i)} A quantization of the \emph{description space} with a \emph{Hierarchical Dictionary} allows to classify point cloud segments in an unsupervised way to symbols in a coarse-to-fine manner, i.e., from basic surface primitives to fine-grained facets of individual object instances. 
Consequently this allows to abstract and analyze unstructured point cloud data in a symbolic manner.

\textbf{ii)}  Our hierarchical representation of object shape decompositions with a \emph{Shape Motif Hierarchy} allows w.r.t.~the \emph{shape space} to reveal topological shape patterns in form of motifs by gradually encoding the decompositions in a local to global manner, i.e. from single segments over composition of segments to a single composition that represents an entire object.
Furthermore, part-based shape reasoning approaches can only perform as good as
coherency and stability of the extracted parts permits.
Parts have to be coherently and stably extractable from unstructured and noisy scenes considering various objects of different shape complexity.
The shape motif hierarchy can alleviate this segmentation problem since the propagation of fine-granular object segments through the hierarchy allows to observe segment compositions at different granularity levels. These segments composed at different levels can be interpreted as object segmentation results at different topological levels.
Subsequently, the shape motif hierarchy encompasses a beneficial representation within the context of segmentation and shape reasoning.

The combination of both, the abstraction from point cloud data to symbols on multiple levels and the hierarchical decomposition of shapes on multiple levels, leads to a representation in form of a \emph{Shape Motif Hierarchy Ensemble}, which discriminatively encodes shape properties of object categories. %
The effectiveness of the proposed representation of shape information is reflected in our experiments in which a classification error of $9.5\%$ was achieved.

We can interpret the classification of objects as a discretization process from a continuous point cloud space in which objects are represented in, to a supervisely determined set of labels which are supposedly uniquely assigned to objects.
Keeping in mind that the categorization of shapes bears uncertainty by the absence of an explicit object model, an unique object-to-label assignment is often not feasible due to shape ambiguities in comparison to instance recognition tasks where an explicit model-to-predict is given beforehand. %
Thus the generalization in shape categorization is even more challenging. 
We believe, a meaning of an unknown object appearance can be drawn by projecting the unknown object into a similarity space where known shape prototype appearances serve as anchors in space which subsequently provide a perspective on the unknown object.
The generated stimuli of a trained $\mathcal{HE}$ from various object observations, can be exploited to create this space ($\mathcal{SR}$).
Consequently, $\mathcal{HE}$ can function as a \emph{descriptor} in form of retrieved stimuli from unknown object instances that can be projected into $\mathcal{SR}$ for reasoning purposes.
Going beyond the shape categorization task, reasoning about shapes where \emph{commonalities} in form of similar $\mathcal{HE}$-\emph{descriptions} lead to similar behavior finds its application in many robotic areas ranging from household to industry such as in generation of grasping primitives for similar object appearances in manipulation~\cite{6696928}, finding substitutes for currently absent objects~\cite{DBLP:conf/icra/AbelhaGS16}, etc.

Further on in the evaluation, an a-priori trained $\mathcal{HE}$ was confronted with object instances from alternative datasets which were not known at the training phase.
As a result the effectiveness of the proposed approach was revealed as a
distinction of instances w.r.t. shape categories was observed which supports the generalization capability of the proposed shape motif hierarchy ensemble under heterogeneous conditions present in different datasets.

\section*{Acknowledgement}
The research leading to the results presented here has received funding from the European Community's Horizon 2020 Framework Programme (H2020-EU.3.2.) within the project (ref.: 635491) ``Effective dexterous ROV operations in presence of communication latencies (DexROV)''.

\bibliographystyle{spmpsci}      %

\end{document}